\documentclass[sigconf]{acmart}

\usepackage{booktabs} 
\usepackage{algorithmic,tikz,rotating}
\usetikzlibrary{arrows,shapes,snakes,matrix}
\usepackage{adjustbox}
\usepackage{subfigure}
\usepackage{comment}
\usepackage{latexsym}
\usepackage{balance}
\usepackage{graphicx}
\usepackage{wrapfig}
\usepackage{multirow}

\definecolor{ao}{rgb}{0.0, 0.5, 0.0}
\definecolor{burntorange}{rgb}{0.8, 0.4, 0.0}

\acmVolume{9}
\acmNumber{4}
\acmArticle{39}
\acmYear{2019}
\acmMonth{3}

\acmDOI{0000001.0000001}

\begin{document}
\title{Image Privacy Prediction 
Using Deep Neural Networks} 

\author{Ashwini Tonge}
\affiliation{%
  \institution{Kansas State University}
  \city{}
  \state{KS}
  \country{USA}}
\email{atonge@ksu.edu}
\author{Cornelia Caragea}
\affiliation{%
  \institution{University of Illinois at Chicago}
  \city{IL}
  \country{USA}
}
\email{cornelia@uic.edu}

\begin{abstract}
Images today are increasingly shared online on social networking sites such as Facebook, Flickr, Foursquare, and Instagram. Image sharing occurs not only within a group of friends but also more and more outside a user's social circles for purposes of social discovery. Despite that current social networking sites allow users to change their privacy preferences, this is often a cumbersome task for the vast majority of users on the Web, who face difficulties in assigning and managing privacy settings. When these privacy settings are used inappropriately, online image sharing can potentially lead to unwanted disclosures and privacy violations. Thus, automatically predicting images' privacy to warn users about private or sensitive content before uploading these images on social networking sites 
has become a necessity in our current interconnected world.

In this paper, we explore learning models to automatically predict appropriate images' privacy as {\em private} or {\em public} using carefully identified image-specific features. We study deep visual semantic features that are derived from various layers of Convolutional Neural Networks (CNNs) as well as textual features such as user tags and deep tags generated from deep CNNs. Particularly, we extract deep (visual and tag) features from four pre-trained CNN architectures for object recognition, i.e., AlexNet, GoogLeNet, VGG-16, and ResNet, and compare their performance for image privacy prediction. Among all four networks, we observe that ResNet produces the best feature representations for this task. We also fine-tune the pre-trained CNN architectures on our privacy dataset and compare their performance with the models trained on pre-trained features. The results show that even though the overall performance obtained using the fine-tuned networks is comparable to that of 
pre-trained networks, the fine-tuned networks provide an improved performance for the private class as compared to models trained on the pre-trained features. Results of our experiments on a Flickr dataset of over thirty thousand images show that the learning models trained on features extracted from ResNet outperform the state-of-the-art models for image privacy prediction. We further investigate the combination of user tags and deep tags derived from CNN architectures using two settings: (1) SVM on the bag-of-tags features; and (2) text-based CNN. We compare these models with the models trained on ResNet visual features obtained for privacy prediction. Our results show that even though the models trained on the visual features perform better than those trained on the tag features, the combination of deep visual features with image tags shows improvements in performance over the individual feature sets. Our  code, features, and the dataset used in experiments are available at \url{https://github.com/ashwinitonge/deepprivate.git}.

\end{abstract}

%
%


%
%


\maketitle

\section{Introduction}

Online image sharing through social networking sites such as Facebook, Flickr, 
and Instagram is on the rise, and so is the sharing of private or sensitive images, which can lead to potential threats to users' privacy when inappropriate privacy settings are used in these platforms. Many users quickly share private images of themselves and their family and friends, without carefully thinking about the consequences of unwanted disclosure and privacy violations \cite{Ahern:2007:OPP:1240624.1240683,Zerr:2012}. For example, it is common now to take photos at cocktail parties and share them on social networking sites without much hesitation. The smartphones facilitate the sharing of photos virtually at any time with people all around the world. These photos can potentially reveal a user's personal and social habits and may be used in the detriment of the photos' owner. 

\begin{figure*}[t]
\setlength{\tabcolsep}{10pt}
\centering
\begin{adjustbox}{max width=14.0cm}
\begin{tabular}{@{}cc@{}}
\setlength{\tabcolsep}{0pt}
\centering
	\begin{tabular}{@{}ccc@{}}
		\includegraphics[scale=0.2]{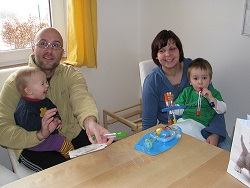} &
		\includegraphics[scale=0.3]{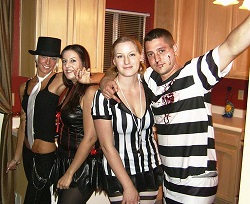} & 
		\includegraphics[scale=0.25]{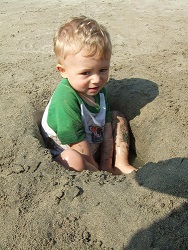} \\ 
		\includegraphics[scale=1.0]{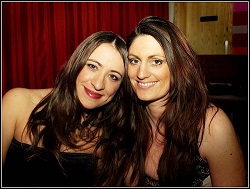} &
		\includegraphics[scale=0.16]{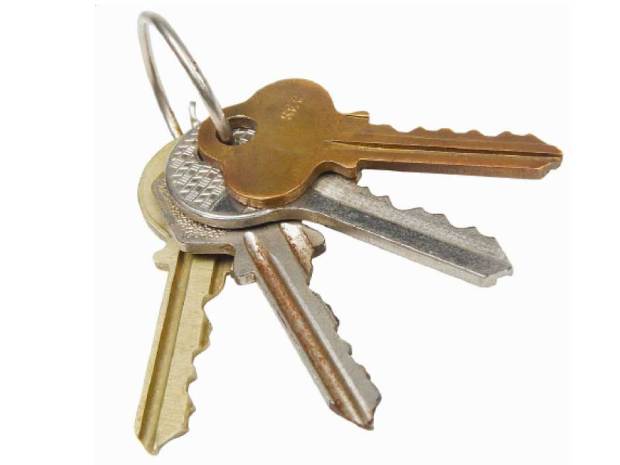} & 
		\includegraphics[scale=0.07]{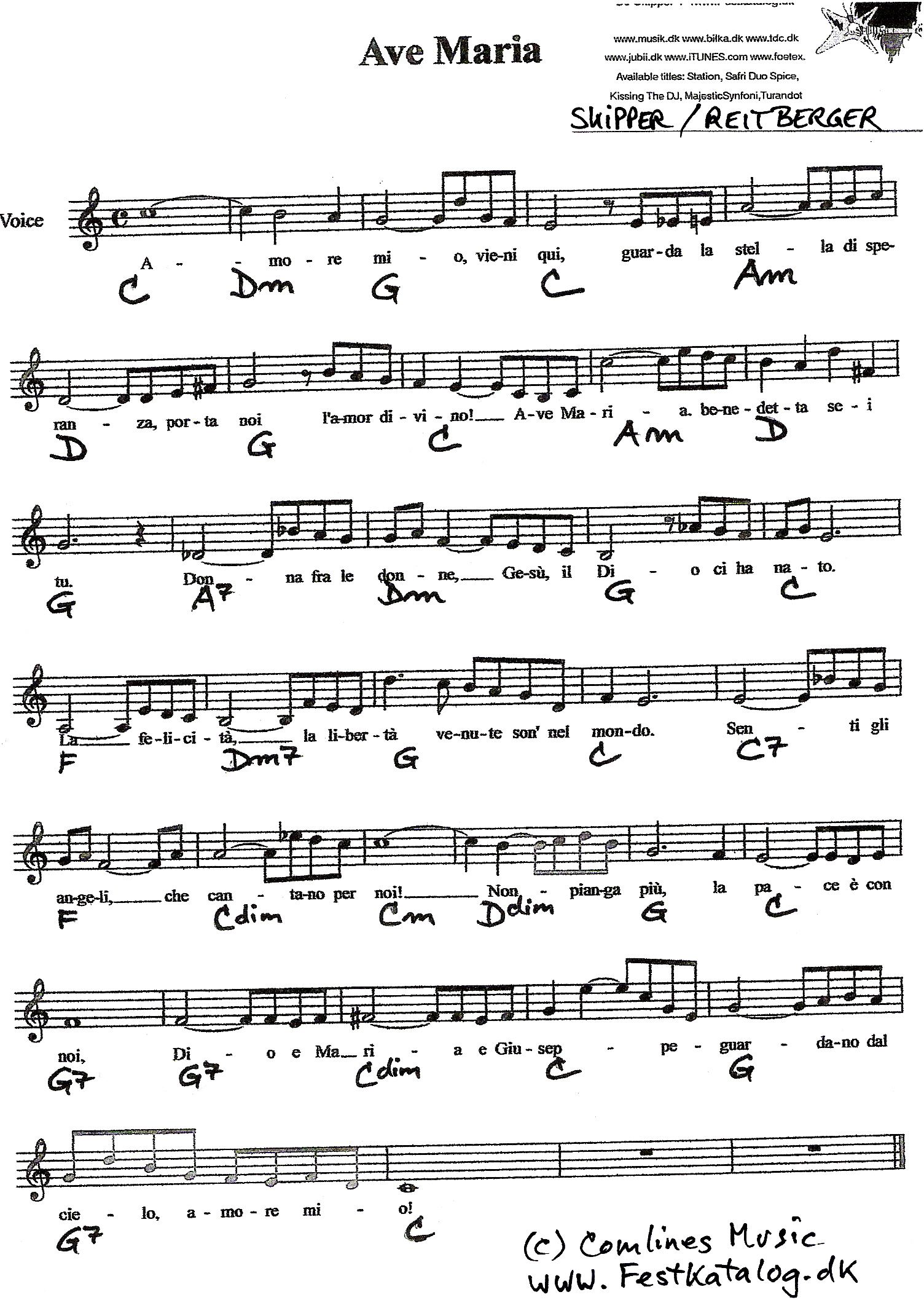} \\         
		& private images& \\
	 \end{tabular} &
\setlength{\tabcolsep}{0pt} 
\centering
	\begin{tabular}{@{}ccc@{}}
		\includegraphics[scale=0.2]{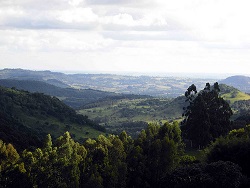} &
		\includegraphics[scale=0.075]{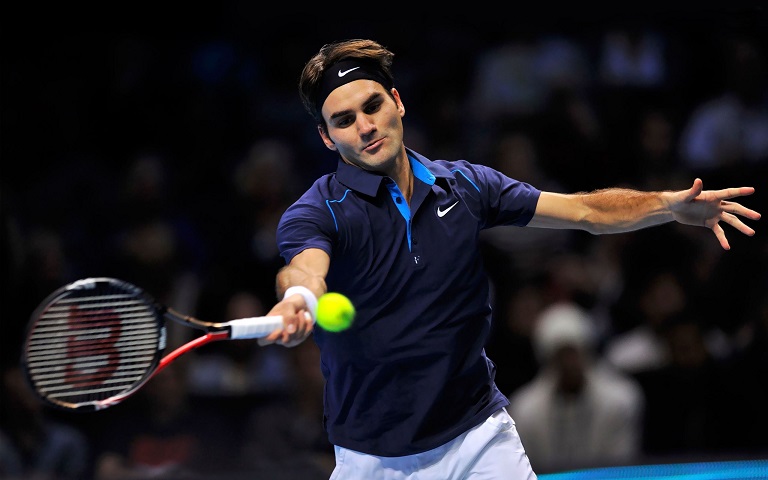} & 
		\includegraphics[scale=1.0]{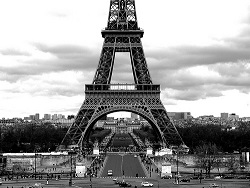} \\
		\includegraphics[scale=1.0]{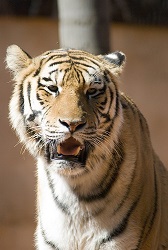} &
		\includegraphics[scale=1.0]{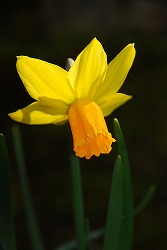} & 
		\includegraphics[scale=0.8]{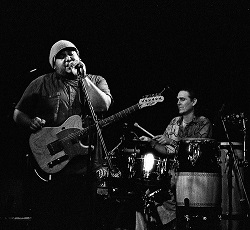} \\        
		& public images& \\
	 \end{tabular}\
\end{tabular}	 
\end{adjustbox}
\caption{Examples of images manually identified as private (left) and public (right).} 
\label{fig:imagesexps}
\end{figure*}

\citet{Gross:2005:IRP:1102199.1102214} analyzed more than 4,000 Carnegie Mellon University students' Facebook profiles and outlined potential threats to privacy. The authors found that users often provide personal information generously on social networking sites, but they rarely change default privacy settings, which could jeopardize their privacy. 
In a parallel study, \citet{Lipford:2008} showed that, although current social networking sites allow users to change their privacy preferences, 
the vast majority of users on the Web 
face difficulties in assigning and managing privacy settings.
Interestingly, \citet{OrekondySF17iccv} showed that, even when users change their privacy settings to comply with their personal privacy preference, they often misjudge the private information in images, which fails to enforce their own privacy preferences. Not surprising, employers these days often perform background checks for their future employees using social networks and about $8\%$ of companies have already fired employees due to their inappropriate social media content \cite{citeulike:10040269}. A study carried out by the Pew Research center reported that $11\%$ of users of social networks regret the posted content \cite{PewResearchPriavcy}.
The Director of the AI Research at Facebook, Yann LeCun \shortcite{yannlecun} urges the development of a digital assistant to warn people about private or sensitive content before embarrassing photos are shared with everyone on social networks.

Identifying private or sensitive content from images is inherently difficult because images' privacy 
is dependent on the owners' personality traits and their level of awareness towards privacy. Still, images' privacy is not purely subjective, but generic patterns of privacy exist. Consider, for example, the images shown in Figure \ref{fig:imagesexps}, which are manually annotated and consistently rated as {\em private} and {\em public} by multiple annotators in a study conducted by \citet{conf/cikm/ZerrSH12,Zerr:2012}.
Notice that the presence of people generally pinpoints to private images, although this is not always true. For example, an image of a musical band in concert is considered to be public. Similarly, images with no people in them could be private, e.g., images with door keys, music notes, legal documents, or someone's art are considered to be private. Indeed, \citet{Laxton:2008:RPK:1455770.1455830} described a ``tele-duplication attack'' that allows an adversary to create a physical key duplicate simply from an image.

Researchers showed that generic patterns of images' privacy can be automatically identified when a large set of images are considered for analysis 
and investigated binary prediction models based on user tags and image content features such as SIFT (Scale Invariant Feature Transform) and RGB (Red Green Blue) \cite{Zerr:2012,Squicciarini2014,Squicciarini:2017:TAO:3062397.2983644}. More recently, several studies \cite{Tran:2016:PFD:3015812.3016006, DBLP:conf/aaai/TongeC16,tongemsm18,DBLP:journals/corr/TongeC15} started to explore privacy frameworks that leverage the benefits of Convolutional Neural Networks (CNNs) for object recognition since, intuitively, the objects present in images 
significantly impact images' privacy (as can be seen from Figure \ref{fig:imagesexps}). 
%
However, these studies used only the AlexNet architecture of CNNs on small dataset sizes.
To date, many deep CNN architectures have been developed and achieve state-of-the-art performance on object recognition. These CNNs include GoogLeNet \cite{DBLP:journals/corr/SzegedyLJSRAEVR14}, VGG-16 \cite{DBLP:journals/corr/SimonyanZ14a}, and ResNet \cite{DBLP:conf/cvpr/HeZRS16} (in addition to AlexNet \cite{NIPS2012_4824}). 
Towards this end, in this paper, we present an extensive study to carefully identify the CNN architectures 
and features derived from these CNNs that can 
adequately predict the class of an image as {\em private} or {\em public}. Our research is motivated by the fact that increasingly, online users' privacy is routinely compromised by using social and content sharing applications \cite{Zheleva:2009}. 
Our models can help users to better manage their participation in online image sharing sites by identifying the sensitive content from the images so that it becomes easier for regular users to control the amount of personal information that they share through these images.

Our contributions are as follows:

\begin{itemize}
\item We study deep visual semantic features and deep image tags derived from CNN architectures pre-trained on the ImageNet dataset and use them in conjunction with Support Vector Machine (SVM) classifiers for image privacy prediction. Specifically, we extract deep features from four successful (pre-trained) CNN architectures for object recognition, AlexNet, GoogLeNet, VGG-16, and ResNet and compare their performance on the task of privacy prediction. 
Through carefully designed experiments, we find that ResNet produces the best feature representations for privacy prediction compared with the other CNNs.

\item We fine-tune the pre-trained CNN architectures on our privacy dataset and use the softmax function to predict the images' privacy as {\em public} or {\em private}. We compare the fine-tuned CNNs with the SVM models obtained on the features derived from the pre-trained CNNs 
%
and show that, although the overall performance obtained by the fine-tuned CNNs is comparable to that of SVM models, 
the fine-tuned networks provide improved recall for the private class as compared to the SVM models trained on the pre-trained features. 


\item We show that the best feature representation produced by ResNet outperforms several baselines for image privacy prediction that consider CNN-based models and SVM models trained on traditional visual features such as SIFT and global GIST descriptor.

\item Next, we investigate the combination of user tags and deep tags derived from CNNs in two settings: (1) using SVM on the bag-of-tags features; and (2) applying the text CNN \cite{DBLP:conf/emnlp/Kim14} on the combination of user tags and deep tags for privacy prediction using the softmax function. We compare these models with the models trained on the most promising visual features extracted from ResNet (obtained from our study) for privacy prediction. Our results show that the models trained on the visual features perform better than those trained on the tag features. 

\item Finally, we explore the combination of deep visual features with image tags and show further improvement in performance over the individual sets of features. 





\end{itemize}




The rest of the paper is organized as follows. We summarize prior work in Section \ref{sec:relwork}. In Section \ref{sec:problem}, we describe the problem statement in details. Section \ref{sec:features} describes the image features obtained from various CNNs for privacy prediction, whereas in Section \ref{sec:data}, we provide details about the dataset that we use to evaluate our models. In Section \ref{sec:experiments}, we present the experiments and describe the experimental setting and results. We finish our analysis in Section \ref{sec:conclusion}, where we provide a brief discussion of our main findings, interesting applications of our work, future directions, and conclude the paper.



\section{Related work} \label{sec:relwork}
Emerging privacy violations 
in social networks have started to attract various researchers to this field \cite{Zheleva:2009}. Researchers also provided public awareness of privacy risks associated with images shared online \cite{conf/raid/XuWS15,Henne:2013:SYP:2462096.2462113}. Along this line, several works are carried out to study users' privacy concerns in social networks, privacy decisions about sharing resources, and the risk associated with them \cite{Krishnamurthy:2008:CPO:1397735.1397744,Simpson:2008:NUF:1461469.1461470,Ghazinour:2013:MRP:2457317.2457344,DBLP:journals/dke/Parra-ArnauRFME12,6399467,Ilia:2015:FPP:2810103.2813603,Gross:2005:IRP:1102199.1102214}. 

Moreover, several works on privacy analysis examined privacy decisions and considerations in mobile and online photo sharing \cite{Ahern:2007:OPP:1240624.1240683,Gross:2005:IRP:1102199.1102214,Jones:2011,1520704}. For example, \citet{Ahern:2007:OPP:1240624.1240683}  explored critical aspects of privacy such as users' consideration for privacy decisions, content and context based patterns of privacy decisions, and how different users adjust their privacy decisions and behavior towards personal information disclosure. The authors concluded that applications that could support and influence users' privacy decision-making process should be developed. 
\citet{Jones:2011}  reinforced the role of privacy-relevant image concepts. For instance, the authors determined that people are more reluctant to share photos capturing social relationships than photos taken for functional purposes; certain settings such as work, bars, concerts cause users to share less. 
\citet{1520704} mentioned that users want to regain control over their shared content, but meanwhile, they feel that configuring proper privacy settings for each image is a burden.

More recent and related to our line of work are the automated image privacy approaches that have been explored along 
four lines of research: {\em social group based approaches}, in which users' profiles are used to partition the friends' lists into multiple groups or circles, and the friends from the same circle are assumed to share similar privacy preferences; {\em location-based approaches}, in which location contexts are used to control the location-based privacy disclosures; {\em tag-based approaches}, in which tags are used for privacy setting recommendations;
and {\em visual-based approaches}, in which the visual content of images is  leveraged for privacy prediction.

\subsubsection*{Social group based  approaches.} Several works emerged to provide the automated privacy decisions for images shared online based on the social groups or circles \cite{Pesce:2012:PAS:2185354.2185358,Christin:2013:SSP:2445646.2446120,Mannan:2008:PSP:1367497.1367564,DBLP:conf/colcom/2009,Squicciarini:2009:CPM:1526709.1526780,Bonneau:2009:PSS:1572532.1572569,Bonneau:2009:PDO:1602240.1602695,Fang:2010:PWS:1772690.1772727,6450896,Danezis:2009:IPP:1654988.1654991,Watson:2015:MUP:2830543.2811257,Kepez:2016:LPR:2970030.2970036,Petkos:2015:SCD:2808797.2809303,DBLP:journals/tkde/SquicciariniLSW15,tags12,Zerr:2012,Yuan:226368}. For example, \citet{Christin:2013:SSP:2445646.2446120} proposed an approach to share content with the users within privacy bubbles. Privacy bubbles represent the private sphere of the users and the access to the content is provided by the bubble creator to people within the bubble. \citet{Bonneau:2009:PDO:1602240.1602695}  introduced the notion of privacy suites which recommend users a  set of  privacy settings that  ``expert'' users or the trusted friends have already established so that ordinary users can either directly accept a  setting or perform minor modifications only. \citet{Fang:2010:PWS:1772690.1772727} developed a privacy assistant to help users grant privileges to their friends. The approach takes as input the privacy preferences for the selected friends and then, using these labels, constructs a  classifier to assign privacy labels to the rest of the (unlabeled) friends based on their profiles. \citet{Danezis:2009:IPP:1654988.1654991} generated privacy settings based on the policy that the information produced within the social circle should remain in that circle itself. Along these lines, \citet{Kapadia_socialcircles} obtained privacy settings by forming clusters of friends by partitioning a user's friends' list. \citet{Yuan:226368} proposed an approach for context-dependent and privacy-aware photo sharing. This approach uses the semantics of the photo and the requester's contextual information in order to define whether an access to the photo will be granted or not at a certain context.
These social group based approaches mostly considered the user  trustworthiness, but ignored the image content sensitiveness, and thus, they may not necessarily provide appropriate privacy settings for online images as the privacy preferences might change according to sensitiveness of the image content.    

\subsubsection*{Location-based  approaches.} These approaches  \cite{Yuan:226368,olejnik:hal-01489684,Bilogrevic2016AMB,10.1007/978-3-642-27576-0_3,Arjun-Baokar,Fisher:2012:SPL:2381934.2381945,DBLP:conf/uss/FriedlandS10,Ravichandran2009,DBLP:conf/sp/ShokriTBH11,Toch:2014:CPP:2581638.2581674,Zhao:2014:PLP:2692983.2693000,Choi:2017:GBP:3078971.3080543} leverage  geo-tags, visual  landmarks  and  other  location contexts  to  control  the  location-based  privacy  disclosures. The geo-tags can be provided manually via social tagging or by adding  location information automatically through the digital cameras or smart-phones having GPS.  The location can also be inferred by identifying places from the shared images through the computer vision techniques. 

\subsubsection*{Tag-based  approaches.} Previous work in the context of tag-based access control policies and privacy prediction for images \cite{yeung2009,tags12,N09-2011,vyas,6450896,DBLP:journals/tkde/SquicciariniLSW15,DeChoudhury:2009:CCC:1698924.1699229,Pesce:2012:PAS:2185354.2185358,Mannan:2008:PSP:1367497.1367564,Ra:2013:PTP:2482626.2482675,Kurtan:2018:PPE:3237383.3238047,tagtoprotect,Zerr:2012} showed initial success in tying user tags with access control rules. For example, \citet{tagtoprotect,6450896}, \citet{Zerr:2012}, and \citet{vyas} explored learning models for image privacy prediction using user tags and found that user tags are informative for predicting images' privacy. Moreover, 
\citet{DBLP:journals/tkde/SquicciariniLSW15} proposed an Adaptive Privacy Policy Prediction framework to help users control access for their shared images. The authors investigated social context, image content, and metadata as potential indicators of privacy preferences. \citet{tags12} studied whether the user annotated tags help to create  and  maintain  access-control  policies  more  intuitively. 
However, the  scarcity of tags for many online images \cite{mum12} and the dimensions of user tags precluded an accurate analysis of images' privacy. 
Hence, in our previous work, \cite{Tonge:2018:PTR:3209542.3209574,tongemsm18,DBLP:conf/aaai/TongeC16,DBLP:conf/aaai/TongeC18,DBLP:journals/corr/TongeC15}, we explored automatic image tagging and showed that the predicted tags combined with user tags can improve 
the overall privacy prediction performance. 

\subsubsection*{Visual-based  approaches.} Several works used visual  features derived from the images' content and showed that they are informative for predicting images' privacy settings \cite{Zerr:2012,Squicciarini2014,Squicciarini:2017:TAO:3062397.2983644,Tran:2016:PFD:3015812.3016006,DBLP:conf/aaai/TongeC16,tongemsm18,DBLP:journals/corr/TongeC15,Buschek:2015,DBLP:conf/bigmm/KuangLL017,DBLP:conf/icmla/YuKYL017,DBLP:journals/tifs/YuKZZLF18,1530530,DBLP:journals/ieicet/NakashimaBF16,DBLP:journals/mms/NakashimaBF12,DBLP:conf/icmcs/NakashimaBF11,Dufaux:2008:SPP:2322613.2324344,orekondy18connecting,7866133,7026919,7568994,vonZezschwitz:2016:YCW:2858036.2858120,DBLP:conf/eccv/WuWWJ18}. For example, \citet{Buschek:2015} presented an approach to assigning privacy to shared images using metadata (location, time, shot details) and visual features (faces, colors, edges). \citet{Zerr:2012} proposed privacy-aware image classification and learned classifiers on Flickr images. The authors considered image tags and visual features such as color histograms, faces, edge-direction coherence, and Scale Invariant Feature Transform (SIFT) for the privacy classification task. SIFT as well as GIST are among the most widely used traditional features for image analysis in computer vision. SIFT \cite{Lowe:2004:DIF:993451.996342} detects scale, rotation, and translation invariant key-points of objects in images and extracts a pool of visual features, which are represented as a ``bag-of-visual-words.'' GIST 
\cite{Oliva:2001:MSS:598425.598462} encodes global descriptors for images and extracts a set of perceptual dimensions (naturalness, openness, roughness, expansion, and ruggedness) that represent the dominant spatial structure of the scene. 
\citet{Squicciarini2014,Squicciarini:2017:TAO:3062397.2983644} performed an in-depth analysis of image privacy classification using Flickr images and found that SIFT and image tags work best for predicting privacy of users' images. 

Recently, the computer vision community has shifted towards convolutional neural networks (CNNs) for tasks such as object detection \cite{sermanet:iclr:14,Sermanet:2013:PDU:2514950.2516194} and semantic segmentation \cite{farabet-pami-13}. CNNs have acquired state-of-the-art results on ImageNet for object recognition  
\cite{ILSVRC15} using supervised learning \cite{NIPS2012_4824}. 
Given the recent success of CNNs, several researchers \cite{Tran:2016:PFD:3015812.3016006,DBLP:conf/aaai/TongeC16,tongemsm18,DBLP:conf/bigmm/KuangLL017,DBLP:conf/icmla/YuKYL017,DBLP:journals/tifs/YuKZZLF18,DBLP:journals/corr/TongeC15} showed promising privacy prediction results compared with visual features such as SIFT and GIST. \citet{DBLP:journals/tifs/YuZKLF17} adopted CNNs to achieve semantic image segmentation and also learned object-privacy relatedness to identify privacy-sensitive objects. 

Using CNNs, some works started to explore personalized privacy prediction models \cite{Spyromitros-Xioufis:2016:PPI:2911996.2912018,zhong2017ijcai,OrekondySF17iccv}. For example, \citet{Spyromitros-Xioufis:2016:PPI:2911996.2912018} used features extracted from CNNs to provide personalized image privacy classification. \citet{zhong2017ijcai} proposed a Group-Based Personalized Model for image privacy classification in online social media sites that learns a set of archetypical privacy models (groups) and associates a given user with one of these groups. \citet{OrekondySF17iccv} defined a set of privacy attributes, which were 
first predicted 
from the image content and then used these attributes in combination with users' preferences to estimate personalized privacy risk. 
Although there is evidence that individuals' sharing behavior is unique, \citet{zhong2017ijcai} argued that personalized models generally require large amounts of user data to learn reliable models, and are time and space consuming to train and store models for each user, while taking into account possible 
sudden changes of users' sharing activities and privacy preferences. 
\citet{OrekondySF17iccv} tried to resolve some of these limitations by clustering users' privacy profiles and training a single classifier that maps the target user into one of these clusters to estimate the personalized privacy score. However, the users' privacy profiles are obtained using a set of attributes. which are defined based on the Personally Identifiable Information
\cite{McCallister:2010:SGP:2206206}, the US Privacy Act of $1974$ and official online social network rules, 
instead of collecting opinions about sensitive content from the actual users of social networking sites. Hence, the definition of sensitive content may not meet a user's actual needs, which limits their applicability in a real-world usage scenario \cite{Li_2018_CVPR_Workshops}. 
In this context, it is worth mentioning that CNNs were also used in another body of privacy related work such as multi-party privacy conflict detection \cite{Zhong:2018:TAM:3269206.3269329} and automatic redaction  of sensitive image content \cite{orekondy18connecting}.

The image representations using visual features and tags are pivotal in above privacy prediction works. In this paper, we aim to study ``deep'' features derived from CNNs, by abstracting out users' privacy preferences and sharing behavior. Precisely, our goal is to identify a set of 
``deep'' features that have the highest discriminative power for image privacy prediction and to flag images that contain private or sensitive content before they are shared on social networking sites. 
%
To our knowledge, this is the first study to provide a detailed analysis of various CNN architectures for privacy prediction. 
Our comprehensive set of experiments can provide the community with evidence about the best CNN architecture and features for the image privacy prediction task, especially since the results obtained outperformed other complex approaches, on a large dataset of more than $30,000$ images.



 




\section{Problem Statement} \label{sec:problem}

Our goal is to accurately identify private or sensitive content from images before they are shared on social networking sites. 
Precisely, given an image, we aim to learn models to classify the image into one of the two classes: {\em private} or {\em public}, based on generic patterns of privacy. Private images belong to the private sphere (e.g., self-portraits, family, friends, someone's home) or contain information that one would not share with everyone else (e.g., private documents). Public images capture content that can be seen by everyone without incurring privacy violations. To achieve our goal, we extract a variety of features from several CNNs and identify those features that have the highest discriminative power for image privacy prediction. 

As the privacy of an image can be determined by the presence of one or more objects described by the visual content and the description associated with it in the form of tags, we consider both visual features and image tags for our analysis.  
For the purpose of this study, we did not consider other contextual information about images (e.g., personal information about the image owner or the owner social network activities, which may or may not be available or easily accessible) since our goal is to predict the privacy of an image solely from the image's content itself. 
We rely on the assumption that, although privacy is a subjective matter, generic patterns of images' privacy exist that can be extracted from the images' visual content and textual tags.

We describe the feature representations considered for our analysis in the next section.


\section{Image encodings}\label{sec:features}

In this section, we provide details on visual content encodings and tag content encodings derived from various CNNs (pre-trained and fine-tuned) to carefully identify the most informative feature representations for image privacy prediction. Particularly, we explore four CNN architectures, AlexNet \cite{NIPS2012_4824}, GoogLeNet \cite{DBLP:journals/corr/SzegedyLJSRAEVR14}, VGG-16 \cite{DBLP:journals/corr/SimonyanZ14a}, and ResNet \cite{DBLP:conf/cvpr/HeZRS16} to derive features for all images in our dataset, which are labeled as private or public. The choice of these architectures is motivated by their good performance on the large scale ImageNet object recognition challenge \cite{ILSVRC15}. 
We also leverage a text-based CNN architecture used for sentence classification \cite{DBLP:conf/emnlp/Kim14} and apply it to images' textual tags for privacy prediction.

\subsection{Preliminary: Convolutional Neural Networks}\label{sec:CNNprelim}
CNN is a type of feed-forward artificial neural network which is inspired by the organization of the animal visual cortex. The learning units in the network are called neurons. These neurons learn to convert input data, i.e., a picture of a dog into its corresponding label, i.e., ``dog'' through automated image recognition. The bottom layers of a CNN consist of interleaved convolution and pooling layers, and the top layers consist of fully-connected (fc) layers, and a probability (prob) layer obtained by applying the softmax function to the input from the previous fc layer, which represents the probability distribution over the available categories for an input image. 
As we ascend through an architecture, the network acquires: (1) lower layers features (color blobs, lines, corners); (2) middle layer features (textures resulted from a combination of lower layers); and (3) higher (deeper) layer features (high-level image content like objects obtained by combining middle layers). Since online images may contain multiple objects, we consider features extracted from deeper layers as they help to encode the objects precisely.

A CNN exploits the 2D topology of image data, in particular, {\em local connectivity} through convolution layers, performs {\em weight sharing} to handle very high-dimensional input data, and can deal with more {\em abstract or global information} through pooling layers. Each unit within a convolution layer receives a small region of its input at location $l$, denoted ${\bf r}_l({\bf x})$ (a.k.a. {\em receptive field}), and applies a non-linear function to it. More precisely, given an input image ${\bf x}$, a unit that is responsible for region $l$ computes $\sigma({\bf W} \cdot {\bf r}_l({\bf x}) + {\bf b})$, where ${\bf W}$ and ${\bf b}$ represent the matrix of weights and the vector of biases, respectively, and $\sigma$ is a non-linear function such as the sigmoid activation or rectified linear activation function. ${\bf W}$ and ${\bf b}$ are learned during training and are shared by all units in a convolution layer. 
Each unit within a pooling layer receives a small region from the previous convolution layer and performs average or max-pooling to obtain more abstract features. 
During training, layers in CNNs are responsible for a forward pass and a backward pass. The forward pass takes inputs and generates the outputs. The backward pass takes gradients with respect to the output and computes the gradient with respect to the parameters and to the inputs, which are consecutively back-propagated to the previous layers \cite{Jia:2014:CCA:2647868.2654889}.

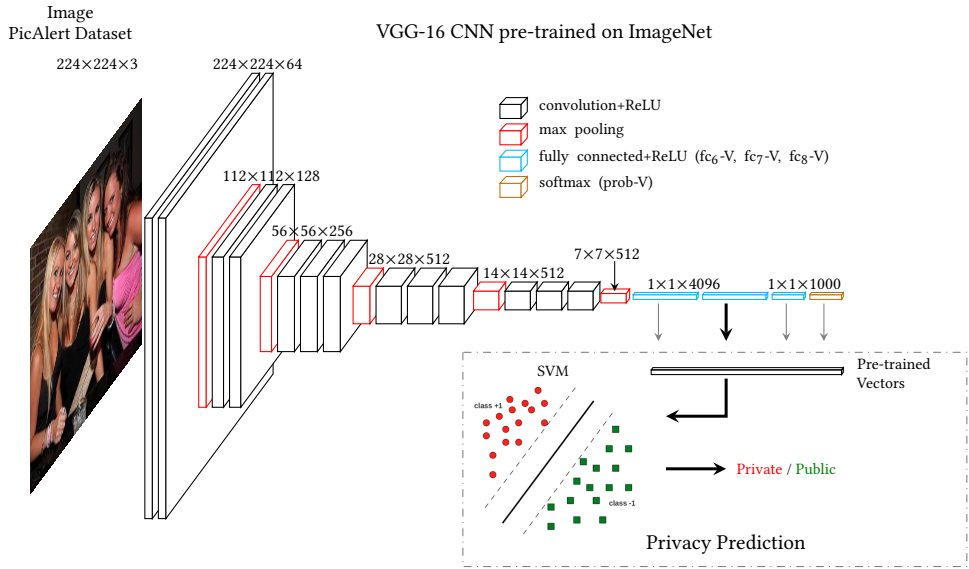
\begin{figure*}[t]
\centering
\begin{tikzpicture}

\pgfmathsetmacro{\cubex}{0.1}
\pgfmathsetmacro{\cubey}{4}
\pgfmathsetmacro{\cubez}{5}

\pgfmathsetmacro{\multx}{1.5}
\pgfmathsetmacro{\divyz}{2}

\definecolor{convfill}{RGB}{255,255,255}
\definecolor{convline}{RGB}{0,0,0}

\definecolor{maxline}{RGB}{255,0,0}
\definecolor{maxfill}{RGB}{255,255,255}

\definecolor{fcfill}{RGB}{255,255,255}
\definecolor{fcline}{RGB}{6,185,251}

\definecolor{softmaxfill}{RGB}{255,255,255}
\definecolor{softmaxline}{RGB}{191,115,16}

\pgfmathsetmacro{\x}{0}
\pgfmathsetmacro{\offset}{0.08}

\pgfmathsetmacro{\xx}{0.5}

\node[inner sep=0pt,cm={0.37 ,0.5 ,0 ,1  ,(0 cm,0 cm)}] (image) at (-1.6,0)
    {\includegraphics[width=4 cm]{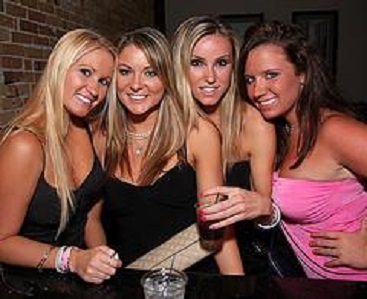}};
    
\node at (-1.8,3.5) {\footnotesize{PicAlert Dataset}};
\node at (-1.8,3.75) {\footnotesize{Image}};

\node at (4.5,3.5) {\small{VGG-16 CNN pre-trained on ImageNet}};
  
\node[text width=3cm] at (\x+\xx/2-\xx-1.2,\cubey/2+0.12,\cubez/2-\cubez) {\scriptsize $224 \! \times \! 224 \! \times \! 3$};

\draw[convline,fill=convfill] (\x+\xx/2,\cubey/2,\cubez/2) -- ++(-\cubex,0,0) -- ++(0,-\cubey,0) -- ++(\cubex,0,0) -- cycle;
\draw[convline,fill=convfill] (\x+\xx/2,\cubey/2,\cubez/2) -- ++(0-\xx,0,-\cubez) -- ++(0,-\cubey,0) -- ++(0+\xx,0,\cubez) -- cycle;
\draw[convline,fill=convfill] (\x+\xx/2,\cubey/2,\cubez/2) -- ++(-\cubex,0,0) -- ++(0-\xx,0,-\cubez) -- ++(\cubex,0,0) -- cycle;

\pgfmathsetmacro{\x}{\x + \cubex + \offset}
\draw[convline,fill=convfill] (\x+\xx/2,\cubey/2,\cubez/2) -- ++(-\cubex,0,0) -- ++(0,-\cubey,0) -- ++(\cubex,0,0) -- cycle;
\draw[convline,fill=convfill] (\x+\xx/2,\cubey/2,\cubez/2) -- ++(0-\xx,0,-\cubez) -- ++(0,-\cubey,0) -- ++(0+\xx,0,\cubez) -- cycle;
\draw[convline,fill=convfill] (\x+\xx/2,\cubey/2,\cubez/2) -- ++(-\cubex,0,0) -- ++(0-\xx,0,-\cubez) -- ++(\cubex,0,0) -- cycle;

\node[text width=3cm] at (\x+\xx/2-\xx+0.7,\cubey/2+0.12,\cubez/2-\cubez) {\scriptsize $224 \! \times \! 224 \! \times \! 64$};

\pgfmathsetmacro{\cubey}{\cubey/\divyz}
\pgfmathsetmacro{\cubez}{\cubez/\divyz}
\pgfmathsetmacro{\x}{\x + \cubex + \offset}
\pgfmathsetmacro{\xx}{\xx/\divyz}
\draw[maxline,fill=maxfill] (\x+\xx/2,\cubey/2,\cubez/2) -- ++(-\cubex,0,0) -- ++(0,-\cubey,0) -- ++(\cubex,0,0) -- cycle;
\draw[maxline,fill=maxfill] (\x+\xx/2,\cubey/2,\cubez/2) -- ++(0-\xx,0,-\cubez) -- ++(0,-\cubey,0) -- ++(0+\xx,0,\cubez) -- cycle;
\draw[maxline,fill=maxfill] (\x+\xx/2,\cubey/2,\cubez/2) -- ++(-\cubex,0,0) -- ++(0-\xx,0,-\cubez) -- ++(\cubex,0,0) -- cycle;

\pgfmathsetmacro{\cubex}{\cubex*\multx}
\pgfmathsetmacro{\x}{\x + \cubex + \offset}
\draw[convline,fill=convfill] (\x+\xx/2,\cubey/2,\cubez/2) -- ++(-\cubex,0,0) -- ++(0,-\cubey,0) -- ++(\cubex,0,0) -- cycle;
\draw[convline,fill=convfill] (\x+\xx/2,\cubey/2,\cubez/2) -- ++(0-\xx,0,-\cubez) -- ++(0,-\cubey,0) -- ++(0+\xx,0,\cubez) -- cycle;
\draw[convline,fill=convfill] (\x+\xx/2,\cubey/2,\cubez/2) -- ++(-\cubex,0,0) -- ++(0-\xx,0,-\cubez) -- ++(\cubex,0,0) -- cycle;

\pgfmathsetmacro{\x}{\x + \cubex + \offset}
\draw[convline,fill=convfill] (\x+\xx/2,\cubey/2,\cubez/2) -- ++(-\cubex,0,0) -- ++(0,-\cubey,0) -- ++(\cubex,0,0) -- cycle;
\draw[convline,fill=convfill] (\x+\xx/2,\cubey/2,\cubez/2) -- ++(0-\xx,0,-\cubez) -- ++(0,-\cubey,0) -- ++(0+\xx,0,\cubez) -- cycle;
\draw[convline,fill=convfill] (\x+\xx/2,\cubey/2,\cubez/2) -- ++(-\cubex,0,0) -- ++(0-\xx,0,-\cubez) -- ++(\cubex,0,0) -- cycle;

\node[text width=3cm] at (\x+\xx/2-\xx+0.55,\cubey/2+0.12,\cubez/2-\cubez) {\scriptsize $112 \! \times \! 112 \! \times \! 128$};

\pgfmathsetmacro{\cubey}{\cubey/\divyz}
\pgfmathsetmacro{\cubez}{\cubez/\divyz}
\pgfmathsetmacro{\x}{\x + \cubex + \offset}
\pgfmathsetmacro{\xx}{\xx/\divyz}
\draw[maxline,fill=maxfill] (\x+\xx/2,\cubey/2,\cubez/2) -- ++(-\cubex,0,0) -- ++(0,-\cubey,0) -- ++(\cubex,0,0) -- cycle;
\draw[maxline,fill=maxfill] (\x+\xx/2,\cubey/2,\cubez/2) -- ++(0-\xx,0,-\cubez) -- ++(0,-\cubey,0) -- ++(0+\xx,0,\cubez) -- cycle;
\draw[maxline,fill=maxfill] (\x+\xx/2,\cubey/2,\cubez/2) -- ++(-\cubex,0,0) -- ++(0-\xx,0,-\cubez) -- ++(\cubex,0,0) -- cycle;

\pgfmathsetmacro{\cubex}{\cubex*\multx}
\pgfmathsetmacro{\x}{\x + \cubex + \offset}
\draw[convline,fill=convfill] (\x+\xx/2,\cubey/2,\cubez/2) -- ++(-\cubex,0,0) -- ++(0,-\cubey,0) -- ++(\cubex,0,0) -- cycle;
\draw[convline,fill=convfill] (\x+\xx/2,\cubey/2,\cubez/2) -- ++(0-\xx,0,-\cubez) -- ++(0,-\cubey,0) -- ++(0+\xx,0,\cubez) -- cycle;
\draw[convline,fill=convfill] (\x+\xx/2,\cubey/2,\cubez/2) -- ++(-\cubex,0,0) -- ++(0-\xx,0,-\cubez) -- ++(\cubex,0,0) -- cycle;

\pgfmathsetmacro{\x}{\x + \cubex + \offset}
\draw[convline,fill=convfill] (\x+\xx/2,\cubey/2,\cubez/2) -- ++(-\cubex,0,0) -- ++(0,-\cubey,0) -- ++(\cubex,0,0) -- cycle;
\draw[convline,fill=convfill] (\x+\xx/2,\cubey/2,\cubez/2) -- ++(0-\xx,0,-\cubez) -- ++(0,-\cubey,0) -- ++(0+\xx,0,\cubez) -- cycle;
\draw[convline,fill=convfill] (\x+\xx/2,\cubey/2,\cubez/2) -- ++(-\cubex,0,0) -- ++(0-\xx,0,-\cubez) -- ++(\cubex,0,0) -- cycle;

\pgfmathsetmacro{\x}{\x + \cubex + \offset}
\draw[convline,fill=convfill] (\x+\xx/2,\cubey/2,\cubez/2) -- ++(-\cubex,0,0) -- ++(0,-\cubey,0) -- ++(\cubex,0,0) -- cycle;
\draw[convline,fill=convfill] (\x+\xx/2,\cubey/2,\cubez/2) -- ++(0-\xx,0,-\cubez) -- ++(0,-\cubey,0) -- ++(0+\xx,0,\cubez) -- cycle;
\draw[convline,fill=convfill] (\x+\xx/2,\cubey/2,\cubez/2) -- ++(-\cubex,0,0) -- ++(0-\xx,0,-\cubez) -- ++(\cubex,0,0) -- cycle;

\node[text width=3cm] at (\x+\xx/2-\xx+0.23,\cubey/2+0.12,\cubez/2-\cubez) {\scriptsize $56 \! \times \! 56 \! \times \! 256$};

\pgfmathsetmacro{\cubey}{\cubey/2}
\pgfmathsetmacro{\cubez}{\cubez/2}
\pgfmathsetmacro{\x}{\x + \cubex + \offset}
\pgfmathsetmacro{\xx}{\xx/\divyz}
\draw[maxline,fill=maxfill] (\x+\xx/2,\cubey/2,\cubez/2) -- ++(-\cubex,0,0) -- ++(0,-\cubey,0) -- ++(\cubex,0,0) -- cycle;
\draw[maxline,fill=maxfill] (\x+\xx/2,\cubey/2,\cubez/2) -- ++(0-\xx,0,-\cubez) -- ++(0,-\cubey,0) -- ++(0+\xx,0,\cubez) -- cycle;
\draw[maxline,fill=maxfill] (\x+\xx/2,\cubey/2,\cubez/2) -- ++(-\cubex,0,0) -- ++(0-\xx,0,-\cubez) -- ++(\cubex,0,0) -- cycle;

\pgfmathsetmacro{\cubex}{\cubex*\multx}
\pgfmathsetmacro{\x}{\x + \cubex + \offset}
\draw[convline,fill=convfill] (\x+\xx/2,\cubey/2,\cubez/2) -- ++(-\cubex,0,0) -- ++(0,-\cubey,0) -- ++(\cubex,0,0) -- cycle;
\draw[convline,fill=convfill] (\x+\xx/2,\cubey/2,\cubez/2) -- ++(0-\xx,0,-\cubez) -- ++(0,-\cubey,0) -- ++(0+\xx,0,\cubez) -- cycle;
\draw[convline,fill=convfill] (\x+\xx/2,\cubey/2,\cubez/2) -- ++(-\cubex,0,0) -- ++(0-\xx,0,-\cubez) -- ++(\cubex,0,0) -- cycle;

\pgfmathsetmacro{\x}{\x + \cubex + \offset}
\draw[convline,fill=convfill] (\x+\xx/2,\cubey/2,\cubez/2) -- ++(-\cubex,0,0) -- ++(0,-\cubey,0) -- ++(\cubex,0,0) -- cycle;
\draw[convline,fill=convfill] (\x+\xx/2,\cubey/2,\cubez/2) -- ++(0-\xx,0,-\cubez) -- ++(0,-\cubey,0) -- ++(0+\xx,0,\cubez) -- cycle;
\draw[convline,fill=convfill] (\x+\xx/2,\cubey/2,\cubez/2) -- ++(-\cubex,0,0) -- ++(0-\xx,0,-\cubez) -- ++(\cubex,0,0) -- cycle;

\pgfmathsetmacro{\x}{\x + \cubex + \offset}
\draw[convline,fill=convfill] (\x+\xx/2,\cubey/2,\cubez/2) -- ++(-\cubex,0,0) -- ++(0,-\cubey,0) -- ++(\cubex,0,0) -- cycle;
\draw[convline,fill=convfill] (\x+\xx/2,\cubey/2,\cubez/2) -- ++(0-\xx,0,-\cubez) -- ++(0,-\cubey,0) -- ++(0+\xx,0,\cubez) -- cycle;
\draw[convline,fill=convfill] (\x+\xx/2,\cubey/2,\cubez/2) -- ++(-\cubex,0,0) -- ++(0-\xx,0,-\cubez) -- ++(\cubex,0,0) -- cycle;

\node[text width=3cm] at (\x+\xx/2-\xx+0.05,\cubey/2+0.12,\cubez/2-\cubez) {\scriptsize $28 \! \times \! 28 \! \times \! 512$};

\pgfmathsetmacro{\cubey}{\cubey/2}
\pgfmathsetmacro{\cubez}{\cubez/2}
\pgfmathsetmacro{\x}{\x + \cubex + \offset}
\pgfmathsetmacro{\xx}{\xx/\divyz}
\draw[maxline,fill=maxfill] (\x+\xx/2,\cubey/2,\cubez/2) -- ++(-\cubex,0,0) -- ++(0,-\cubey,0) -- ++(\cubex,0,0) -- cycle;
\draw[maxline,fill=maxfill] (\x+\xx/2,\cubey/2,\cubez/2) -- ++(0-\xx,0,-\cubez) -- ++(0,-\cubey,0) -- ++(0+\xx,0,\cubez) -- cycle;
\draw[maxline,fill=maxfill] (\x+\xx/2,\cubey/2,\cubez/2) -- ++(-\cubex,0,0) -- ++(0-\xx,0,-\cubez) -- ++(\cubex,0,0) -- cycle;

\pgfmathsetmacro{\x}{\x + \cubex + \offset}
\draw[convline,fill=convfill] (\x+\xx/2,\cubey/2,\cubez/2) -- ++(-\cubex,0,0) -- ++(0,-\cubey,0) -- ++(\cubex,0,0) -- cycle;
\draw[convline,fill=convfill] (\x+\xx/2,\cubey/2,\cubez/2) -- ++(0-\xx,0,-\cubez) -- ++(0,-\cubey,0) -- ++(0+\xx,0,\cubez) -- cycle;
\draw[convline,fill=convfill] (\x+\xx/2,\cubey/2,\cubez/2) -- ++(-\cubex,0,0) -- ++(0-\xx,0,-\cubez) -- ++(\cubex,0,0) -- cycle;

\pgfmathsetmacro{\x}{\x + \cubex + \offset}
\draw[convline,fill=convfill] (\x+\xx/2,\cubey/2,\cubez/2) -- ++(-\cubex,0,0) -- ++(0,-\cubey,0) -- ++(\cubex,0,0) -- cycle;
\draw[convline,fill=convfill] (\x+\xx/2,\cubey/2,\cubez/2) -- ++(0-\xx,0,-\cubez) -- ++(0,-\cubey,0) -- ++(0+\xx,0,\cubez) -- cycle;
\draw[convline,fill=convfill] (\x+\xx/2,\cubey/2,\cubez/2) -- ++(-\cubex,0,0) -- ++(0-\xx,0,-\cubez) -- ++(\cubex,0,0) -- cycle;

\pgfmathsetmacro{\x}{\x + \cubex + \offset}
\draw[convline,fill=convfill] (\x+\xx/2,\cubey/2,\cubez/2) -- ++(-\cubex,0,0) -- ++(0,-\cubey,0) -- ++(\cubex,0,0) -- cycle;
\draw[convline,fill=convfill] (\x+\xx/2,\cubey/2,\cubez/2) -- ++(0-\xx,0,-\cubez) -- ++(0,-\cubey,0) -- ++(0+\xx,0,\cubez) -- cycle;
\draw[convline,fill=convfill] (\x+\xx/2,\cubey/2,\cubez/2) -- ++(-\cubex,0,0) -- ++(0-\xx,0,-\cubez) -- ++(\cubex,0,0) -- cycle;

\node[text width=3cm] at (\x+\xx/2-\xx-0.05,\cubey/2+0.12,\cubez/2-\cubez) {\scriptsize $14 \! \times \! 14 \! \times \! 512$};

\pgfmathsetmacro{\cubey}{\cubey/2}
\pgfmathsetmacro{\cubez}{\cubez/2}
\pgfmathsetmacro{\x}{\x + \cubex + \offset}
\pgfmathsetmacro{\xx}{\xx/\divyz}
\draw[maxline,fill=maxfill] (\x+\xx/2,\cubey/2,\cubez/2) -- ++(-\cubex,0,0) -- ++(0,-\cubey,0) -- ++(\cubex,0,0) -- cycle;
\draw[maxline,fill=maxfill] (\x+\xx/2,\cubey/2,\cubez/2) -- ++(0-\xx,0,-\cubez) -- ++(0,-\cubey,0) -- ++(0+\xx,0,\cubez) -- cycle;
\draw[maxline,fill=maxfill] (\x+\xx/2,\cubey/2,\cubez/2) -- ++(-\cubex,0,0) -- ++(0-\xx,0,-\cubez) -- ++(\cubex,0,0) -- cycle;

\node[text width=3cm] (sizepool5) at (\x+\xx/2-\xx+0.8,\cubey/2+0.5,0) {\scriptsize $7 \! \times \! 7 \! \times \! 512$};

\draw[-stealth] (\x+\xx/2-\xx -\cubex/2,\cubey/2+0.4,0)--(\x+\xx/2-\xx -\cubex/2,\cubey/2 ,0);

\pgfmathsetmacro{\cubex}{\cubex*2.5}
\pgfmathsetmacro{\cubey}{\cubey/2}
\pgfmathsetmacro{\cubez}{\cubez/2}
\pgfmathsetmacro{\x}{\x + \cubex + \offset}
\pgfmathsetmacro{\xx}{\xx/\divyz}
\draw[fcline,fill=fcfill] (\x+\xx/2,\cubey/2,\cubez/2) -- ++(-\cubex,0,0) -- ++(0,-\cubey,0) -- ++(\cubex,0,0) -- cycle;
\draw[fcline,fill=fcfill] (\x+\xx/2,\cubey/2,\cubez/2) -- ++(0-\xx,0,-\cubez) -- ++(0,-\cubey,0) -- ++(0+\xx,0,\cubez) -- cycle;
\draw[fcline,fill=fcfill] (\x+\xx/2,\cubey/2,\cubez/2) -- ++(-\cubex,0,0) -- ++(0-\xx,0,-\cubez) -- ++(\cubex,0,0) -- cycle;

\pgfmathsetmacro{\x}{\x + \cubex + \offset}
\draw[fcline,fill=fcfill] (\x+\xx/2,\cubey/2,\cubez/2) -- ++(-\cubex,0,0) -- ++(0,-\cubey,0) -- ++(\cubex,0,0) -- cycle;
\draw[fcline,fill=fcfill] (\x+\xx/2,\cubey/2,\cubez/2) -- ++(0-\xx,0,-\cubez) -- ++(0,-\cubey,0) -- ++(0+\xx,0,\cubez) -- cycle;
\draw[fcline,fill=fcfill] (\x+\xx/2,\cubey/2,\cubez/2) -- ++(-\cubex,0,0) -- ++(0-\xx,0,-\cubez) -- ++(\cubex,0,0) -- cycle;

\node[text width=3cm] at (\x+\xx/2-\xx-0.1,\cubey/2+0.12,\cubez/2-\cubez) {\scriptsize $1 \! \times \! 1 \! \times \! 4096$};

\pgfmathsetmacro{\cubex}{\cubex/2}
\pgfmathsetmacro{\x}{\x + \cubex + \offset}
\draw[fcline,fill=fcfill] (\x+\xx/2,\cubey/2,\cubez/2) -- ++(-\cubex,0,0) -- ++(0,-\cubey,0) -- ++(\cubex,0,0) -- cycle;
\draw[fcline,fill=fcfill] (\x+\xx/2,\cubey/2,\cubez/2) -- ++(0-\xx,0,-\cubez) -- ++(0,-\cubey,0) -- ++(0+\xx,0,\cubez) -- cycle;
\draw[fcline,fill=fcfill] (\x+\xx/2,\cubey/2,\cubez/2) -- ++(-\cubex,0,0) -- ++(0-\xx,0,-\cubez) -- ++(\cubex,0,0) -- cycle;

\pgfmathsetmacro{\x}{\x + \cubex + \offset}
\draw[softmaxline,fill=softmaxfill] (\x+\xx/2,\cubey/2,\cubez/2) -- ++(-\cubex,0,0) -- ++(0,-\cubey,0) -- ++(\cubex,0,0) -- cycle;
\draw[softmaxline,fill=softmaxfill] (\x+\xx/2,\cubey/2,\cubez/2) -- ++(0-\xx,0,-\cubez) -- ++(0,-\cubey,0) -- ++(0+\xx,0,\cubez) -- cycle;
\draw[softmaxline,fill=softmaxfill] (\x+\xx/2,\cubey/2,\cubez/2) -- ++(-\cubex,0,0) -- ++(0-\xx,0,-\cubez) -- ++(\cubex,0,0) -- cycle;

\node[text width=3cm] at (\x+\xx/2-\xx+0.5,\cubey/2+0.12,\cubez/2-\cubez) {\scriptsize $1 \! \times \! 1 \! \times \! 1000$};

\pgfmathsetmacro{\y}{-1}
\pgfmathsetmacro{\x}{\x}
\pgfmathsetmacro{\cubex}{\cubex*6}
\draw[] (\x+\xx/2,\y+\cubey/2,\cubez/2) -- ++(-\cubex,0,0) -- ++(0,-\cubey,0) -- ++(\cubex,0,0) -- cycle;
\draw[] (\x+\xx/2,\y+\cubey/2,\cubez/2) -- ++(0-\xx,0,-\cubez) -- ++(0,-\cubey,0) -- ++(0+\xx,0,\cubez) -- cycle;
\draw[] (\x+\xx/2,\y+\cubey/2,\cubez/2) -- ++(-\cubex,0,0) -- ++(0-\xx,0,-\cubez) -- ++(\cubex,0,0) -- cycle;

\node[text width=3cm] at (\x+1.7,\y+0.1) {\scriptsize Pre-trained};
\node[text width=3cm] at (\x+1.7,\y-0.15) {\scriptsize Vectors};

\pgfmathsetmacro{\cubex}{\cubex/6}

\pgfmathsetmacro{\x}{\x/2}
\pgfmathsetmacro{\y}{2.5}
\pgfmathsetmacro{\cubex}{\cubex/1.5}
\pgfmathsetmacro{\cubey}{\cubey*3}
\pgfmathsetmacro{\cubez}{\cubez*3}

\draw[convline,fill=convfill] (\x+\xx/2,\y+\cubey/2,\cubez/2) -- ++(-\cubex,0,0) -- ++(0,-\cubey,0) -- ++(\cubex,0,0) -- cycle;
\draw[convline,fill=convfill] (\x+\xx/2,\y+\cubey/2,\cubez/2) -- ++(0-\xx,0,-\cubez) -- ++(0,-\cubey,0) -- ++(0+\xx,0,\cubez) -- cycle;
\draw[convline,fill=convfill] (\x+\xx/2,\y+\cubey/2,\cubez/2) -- ++(-\cubex,0,0) -- ++(0-\xx,0,-\cubez) -- ++(\cubex,0,0) -- cycle;

\node[text width=3cm] at (\x+1.7,\y+0.05,0) {\scriptsize convolution+ReLU};

\pgfmathsetmacro{\y}{\y-0.35}
\draw[maxline,fill=maxfill] (\x+\xx/2,\y+\cubey/2,\cubez/2) -- ++(-\cubex,0,0) -- ++(0,-\cubey,0) -- ++(\cubex,0,0) -- cycle;
\draw[maxline,fill=maxfill] (\x+\xx/2,\y+\cubey/2,\cubez/2) -- ++(0-\xx,0,-\cubez) -- ++(0,-\cubey,0) -- ++(0+\xx,0,\cubez) -- cycle;
\draw[maxline,fill=maxfill] (\x+\xx/2,\y+\cubey/2,\cubez/2) -- ++(-\cubex,0,0) -- ++(0-\xx,0,-\cubez) -- ++(\cubex,0,0) -- cycle;

\node[text width=3cm] at (\x+1.7,\y+0.05,0) {\scriptsize max pooling};

\pgfmathsetmacro{\y}{\y-0.35}
\draw[fcline,fill=fcfill] (\x+\xx/2,\y+\cubey/2,\cubez/2) -- ++(-\cubex,0,0) -- ++(0,-\cubey,0) -- ++(\cubex,0,0) -- cycle;
\draw[fcline,fill=fcfill] (\x+\xx/2,\y+\cubey/2,\cubez/2) -- ++(0-\xx,0,-\cubez) -- ++(0,-\cubey,0) -- ++(0+\xx,0,\cubez) -- cycle;
\draw[fcline,fill=fcfill] (\x+\xx/2,\y+\cubey/2,\cubez/2) -- ++(-\cubex,0,0) -- ++(0-\xx,0,-\cubez) -- ++(\cubex,0,0) -- cycle;

\node[text width=5cm] at (\x+2.7,\y+0.05,0) {\scriptsize fully connected+ReLU (fc$_6$-V, fc$_7$-V, fc$_8$-V)};

\pgfmathsetmacro{\y}{\y-0.35}
\draw[softmaxline,fill=softmaxfill] (\x+\xx/2,\y+\cubey/2,\cubez/2) -- ++(-\cubex,0,0) -- ++(0,-\cubey,0) -- ++(\cubex,0,0) -- cycle;
\draw[softmaxline,fill=softmaxfill] (\x+\xx/2,\y+\cubey/2,\cubez/2) -- ++(0-\xx,0,-\cubez) -- ++(0,-\cubey,0) -- ++(0+\xx,0,\cubez) -- cycle;
\draw[softmaxline,fill=softmaxfill] (\x+\xx/2,\y+\cubey/2,\cubez/2) -- ++(-\cubex,0,0) -- ++(0-\xx,0,-\cubez) -- ++(\cubex,0,0) -- cycle;

\node[text width=3cm] at (\x+1.7,\y+0.05,0) {\scriptsize softmax (prob-V)};

\pgfmathsetmacro{\x}{\x/2}
\pgfmathsetmacro{\y}{-0.1}




\draw[-stealth,,line width=.4mm] (\x+4.8,\y) to (\x+4.8,\y-0.5);
\draw[-stealth,draw=gray] (\x+3.9,\y) to (\x+3.9,\y-0.5);
\draw[-stealth,draw=gray] (\x+5.6,\y) to (\x+5.6,\y-0.5);
\draw[-stealth,draw=gray] (\x+6.1,\y) to (\x+6.1,\y-0.5);
\node at (\x+2.5,\y-0.9) {\scriptsize SVM};
\node at (\x+2.5,\y-2.1) {\includegraphics[scale=0.13]{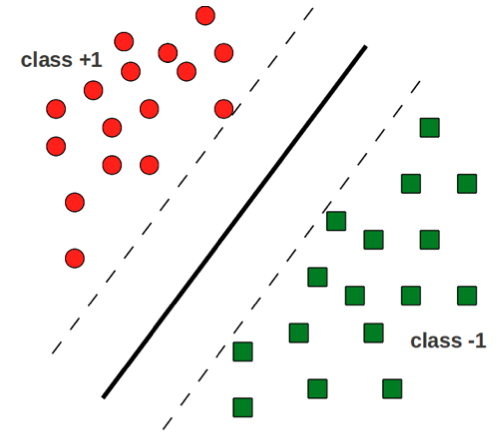}};

\draw[-stealth, line width=.4mm] (\x+4.8,\y-1.0) to (\x+4.8,\y-1.5) to (\x+4, \y-1.5);
\draw[-stealth, line width=.4mm] (\x+4,\y-2.2) to (\x+4.8,\y-2.2);

\node at (\x+5.6,\y-2.2) {\scriptsize {\color{red}Private} / {\color{ao}Public}};

\draw[draw=gray, dash dot] (\x+1.3,\y-0.65) -- (\x+1.3,\y-3.5) -- (\x+8,\y-3.5)-- (\x+8,\y-0.65) -- (\x+1.3,\y-0.65);

\node at (\x+4.8,\y-3.2) {\small {Privacy Prediction}};

\end{tikzpicture}
\caption{\label{fig:pretrainfeat}Image encoding using pre-trained CNN: (1) We employ a CNN (e.g. VGG-16) pre-trained on the ImageNet object dataset. (2) We derive high-level features from the image's visual content using fully connected layers (fc$_6$-V, fc$_7$-V, and fc$_8$-V) and probability layer (softmax) of the pre-trained network.}
\end{figure*}

\subsection{Features Derived Through Pre-Trained CNNs} 
We describe a diverse set of features derived from CNN architectures pre-trained on the ILSVRC-2012 object classification subset of the ImageNet dataset that contains $1000$ object categories and 1.2 million images \cite{ILSVRC15}. We consider powerful features obtained from various fully-connected layers of a CNN that are generated by the previous convolutional layers, and use them to learn a decision function whose sign represents the class ({\em private} or {\em public}) assigned to an input image ${\bf x}$. The activations of the fully connected layers capture the complete object contained in the region of interest. Hence,  we use the activations of the fully-connected layers of a CNN as a feature vector. For image encoding, we also use the probability (prob) layer obtained by applying the softmax function to the output of the (last) fully-connected layer. We extract features from the four pre-trained CNNs as follows.

The {\bf \em AlexNet} architecture implements an eight-layer network; the first five layers of AlexNet are convolutional, and the remaining three layers are fully-connected. 
We extract features from the three fully-connected layers, which are referred as fc$_6$-A, fc$_7$-A, and fc$_8$-A, and from the output layer denoted as ``prob-A.'' The dimensions of fc$_6$-A, fc$_7$-A, fc$_8$-A, and prob-A are $4096$, $4096$, $1000$, and $1000$, respectively. 

The {\bf \em GoogLeNet} architecture implements a $22$ layer deep network with Inception architecture. The  architecture is a combination of all layers with their output filter bank concatenated so as to form input for the next stage. 
We extract features from the last two layers named as ``loss$_3$-G/classifier'' (InnerProduct layer) and the output layer denoted as ``prob-G.'' The dimension of loss$_3$-G and prob-G is $1000$. 

The {\bf \em VGG-16} architecture implements a $16$ layer deep network; a stack of convolutional layers with a very small receptive filed: $3 \times 3$ followed by fully-connected layers. The architecture contains $13$ convolutional layers and $3$ fully-connected layers. The number of channels of the convolutional layers starts from $64$ in the first layer and then increases by a factor of $2$ after each max-pooling layers until it reaches $512$. We refer to features extracted from the fully-connected layers as fc$_6$-V, fc$_7$-V, fc$_8$-V, and the output layer as ``prob-V.'' The dimensions of fc$_6$-V, fc$_7$-V, fc$_8$-V, and prob-V are $4096$, $4096$, $1000$, and $1000$, respectively. 

The {\bf \em ResNet} (or Residual network) alleviates the vanishing gradient problem by introducing short paths to carry gradient throughout the extent of very deep networks and allows the construction of deeper architectures. A residual unit with an identity mapping is defined as:
\[X^{l+1} = X^{l} + \mathcal{F}(X^{l})\]
where $X^{l}$ is the input and $X^{l+1}$ is the output of the residual unit; $\mathcal{F}$ is a residual function, e.g., a stack of two $3 \times 3$ convolution layers in \cite{DBLP:conf/cvpr/HeZRS16}. The main idea of the residual learning is to learn the additive residual function $\mathcal{F}$ with respect to $X^{l}$ \cite{He2016IdentityMI}. Intuitively, ResNets can be explained by considering residual functions as paths through which information can propagate easily. This interprets as ResNets learn more complex feature representations which are combined with the shallower descriptions obtained from previous layers. 
We refer to features extracted from the fully-connected layer as fc-R and the output layer as ``prob-R.'' The dimension of fc-R and prob-R is $1000$.

The feature extraction using the pre-trained network for an input image from our dataset is shown in Figure \ref{fig:pretrainfeat}. In the figure, we show VGG-16 as the pre-trained network for illustrating the feature extraction. 


\begin{figure*}[t]
\centering
\begin{tikzpicture}

\pgfmathsetmacro{\cubex}{0.1}
\pgfmathsetmacro{\cubey}{4}
\pgfmathsetmacro{\cubez}{5}

\pgfmathsetmacro{\multx}{1.5}
\pgfmathsetmacro{\divyz}{2}

\definecolor{convfill}{RGB}{255,255,255}
\definecolor{convline}{RGB}{0,0,0}

\definecolor{maxline}{RGB}{255,0,0}
\definecolor{maxfill}{RGB}{255,255,255}

\definecolor{fcfill}{RGB}{255,255,255}
\definecolor{fcline}{RGB}{6,185,251}

\definecolor{softmaxfill}{RGB}{255,255,255}
\definecolor{softmaxline}{RGB}{191,115,16}

\pgfmathsetmacro{\x}{0}
\pgfmathsetmacro{\offset}{0.08}

\pgfmathsetmacro{\xx}{0.5}

\node[inner sep=0pt,cm={0.37 ,0.5 ,0 ,1  ,(0 cm,0 cm)}] (image) at (-1.6,0)
    {\includegraphics[width=4 cm]{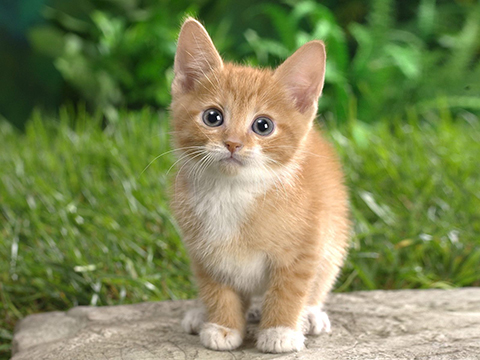}};
    
\node at (-1.8,3.5) {\footnotesize{Object Dataset (ImageNet)}};
\node at (-1.8,3.75) {\footnotesize{Image}};

\node at (4.5,3.5) {\small{VGG-16 CNN pre-trained on ImageNet}};
  
\node[text width=3cm] at (\x+\xx/2-\xx-1.2,\cubey/2+0.12,\cubez/2-\cubez) {\scriptsize $224 \! \times \! 224 \! \times \! 3$};

\draw[convline,fill=convfill] (\x+\xx/2,\cubey/2,\cubez/2) -- ++(-\cubex,0,0) -- ++(0,-\cubey,0) -- ++(\cubex,0,0) -- cycle;
\draw[convline,fill=convfill] (\x+\xx/2,\cubey/2,\cubez/2) -- ++(0-\xx,0,-\cubez) -- ++(0,-\cubey,0) -- ++(0+\xx,0,\cubez) -- cycle;
\draw[convline,fill=convfill] (\x+\xx/2,\cubey/2,\cubez/2) -- ++(-\cubex,0,0) -- ++(0-\xx,0,-\cubez) -- ++(\cubex,0,0) -- cycle;

\pgfmathsetmacro{\x}{\x + \cubex + \offset}
\draw[convline,fill=convfill] (\x+\xx/2,\cubey/2,\cubez/2) -- ++(-\cubex,0,0) -- ++(0,-\cubey,0) -- ++(\cubex,0,0) -- cycle;
\draw[convline,fill=convfill] (\x+\xx/2,\cubey/2,\cubez/2) -- ++(0-\xx,0,-\cubez) -- ++(0,-\cubey,0) -- ++(0+\xx,0,\cubez) -- cycle;
\draw[convline,fill=convfill] (\x+\xx/2,\cubey/2,\cubez/2) -- ++(-\cubex,0,0) -- ++(0-\xx,0,-\cubez) -- ++(\cubex,0,0) -- cycle;

\node[text width=3cm] at (\x+\xx/2-\xx+0.7,\cubey/2+0.12,\cubez/2-\cubez) {\scriptsize $224 \! \times \! 224 \! \times \! 64$};

\pgfmathsetmacro{\cubey}{\cubey/\divyz}
\pgfmathsetmacro{\cubez}{\cubez/\divyz}
\pgfmathsetmacro{\x}{\x + \cubex + \offset}
\pgfmathsetmacro{\xx}{\xx/\divyz}
\draw[maxline,fill=maxfill] (\x+\xx/2,\cubey/2,\cubez/2) -- ++(-\cubex,0,0) -- ++(0,-\cubey,0) -- ++(\cubex,0,0) -- cycle;
\draw[maxline,fill=maxfill] (\x+\xx/2,\cubey/2,\cubez/2) -- ++(0-\xx,0,-\cubez) -- ++(0,-\cubey,0) -- ++(0+\xx,0,\cubez) -- cycle;
\draw[maxline,fill=maxfill] (\x+\xx/2,\cubey/2,\cubez/2) -- ++(-\cubex,0,0) -- ++(0-\xx,0,-\cubez) -- ++(\cubex,0,0) -- cycle;

\pgfmathsetmacro{\cubex}{\cubex*\multx}
\pgfmathsetmacro{\x}{\x + \cubex + \offset}
\draw[convline,fill=convfill] (\x+\xx/2,\cubey/2,\cubez/2) -- ++(-\cubex,0,0) -- ++(0,-\cubey,0) -- ++(\cubex,0,0) -- cycle;
\draw[convline,fill=convfill] (\x+\xx/2,\cubey/2,\cubez/2) -- ++(0-\xx,0,-\cubez) -- ++(0,-\cubey,0) -- ++(0+\xx,0,\cubez) -- cycle;
\draw[convline,fill=convfill] (\x+\xx/2,\cubey/2,\cubez/2) -- ++(-\cubex,0,0) -- ++(0-\xx,0,-\cubez) -- ++(\cubex,0,0) -- cycle;

\pgfmathsetmacro{\x}{\x + \cubex + \offset}
\draw[convline,fill=convfill] (\x+\xx/2,\cubey/2,\cubez/2) -- ++(-\cubex,0,0) -- ++(0,-\cubey,0) -- ++(\cubex,0,0) -- cycle;
\draw[convline,fill=convfill] (\x+\xx/2,\cubey/2,\cubez/2) -- ++(0-\xx,0,-\cubez) -- ++(0,-\cubey,0) -- ++(0+\xx,0,\cubez) -- cycle;
\draw[convline,fill=convfill] (\x+\xx/2,\cubey/2,\cubez/2) -- ++(-\cubex,0,0) -- ++(0-\xx,0,-\cubez) -- ++(\cubex,0,0) -- cycle;

\node[text width=3cm] at (\x+\xx/2-\xx+0.55,\cubey/2+0.12,\cubez/2-\cubez) {\scriptsize $112 \! \times \! 112 \! \times \! 128$};

\pgfmathsetmacro{\cubey}{\cubey/\divyz}
\pgfmathsetmacro{\cubez}{\cubez/\divyz}
\pgfmathsetmacro{\x}{\x + \cubex + \offset}
\pgfmathsetmacro{\xx}{\xx/\divyz}
\draw[maxline,fill=maxfill] (\x+\xx/2,\cubey/2,\cubez/2) -- ++(-\cubex,0,0) -- ++(0,-\cubey,0) -- ++(\cubex,0,0) -- cycle;
\draw[maxline,fill=maxfill] (\x+\xx/2,\cubey/2,\cubez/2) -- ++(0-\xx,0,-\cubez) -- ++(0,-\cubey,0) -- ++(0+\xx,0,\cubez) -- cycle;
\draw[maxline,fill=maxfill] (\x+\xx/2,\cubey/2,\cubez/2) -- ++(-\cubex,0,0) -- ++(0-\xx,0,-\cubez) -- ++(\cubex,0,0) -- cycle;

\pgfmathsetmacro{\cubex}{\cubex*\multx}
\pgfmathsetmacro{\x}{\x + \cubex + \offset}
\draw[convline,fill=convfill] (\x+\xx/2,\cubey/2,\cubez/2) -- ++(-\cubex,0,0) -- ++(0,-\cubey,0) -- ++(\cubex,0,0) -- cycle;
\draw[convline,fill=convfill] (\x+\xx/2,\cubey/2,\cubez/2) -- ++(0-\xx,0,-\cubez) -- ++(0,-\cubey,0) -- ++(0+\xx,0,\cubez) -- cycle;
\draw[convline,fill=convfill] (\x+\xx/2,\cubey/2,\cubez/2) -- ++(-\cubex,0,0) -- ++(0-\xx,0,-\cubez) -- ++(\cubex,0,0) -- cycle;

\pgfmathsetmacro{\x}{\x + \cubex + \offset}
\draw[convline,fill=convfill] (\x+\xx/2,\cubey/2,\cubez/2) -- ++(-\cubex,0,0) -- ++(0,-\cubey,0) -- ++(\cubex,0,0) -- cycle;
\draw[convline,fill=convfill] (\x+\xx/2,\cubey/2,\cubez/2) -- ++(0-\xx,0,-\cubez) -- ++(0,-\cubey,0) -- ++(0+\xx,0,\cubez) -- cycle;
\draw[convline,fill=convfill] (\x+\xx/2,\cubey/2,\cubez/2) -- ++(-\cubex,0,0) -- ++(0-\xx,0,-\cubez) -- ++(\cubex,0,0) -- cycle;

\pgfmathsetmacro{\x}{\x + \cubex + \offset}
\draw[convline,fill=convfill] (\x+\xx/2,\cubey/2,\cubez/2) -- ++(-\cubex,0,0) -- ++(0,-\cubey,0) -- ++(\cubex,0,0) -- cycle;
\draw[convline,fill=convfill] (\x+\xx/2,\cubey/2,\cubez/2) -- ++(0-\xx,0,-\cubez) -- ++(0,-\cubey,0) -- ++(0+\xx,0,\cubez) -- cycle;
\draw[convline,fill=convfill] (\x+\xx/2,\cubey/2,\cubez/2) -- ++(-\cubex,0,0) -- ++(0-\xx,0,-\cubez) -- ++(\cubex,0,0) -- cycle;

\node[text width=3cm] at (\x+\xx/2-\xx+0.23,\cubey/2+0.12,\cubez/2-\cubez) {\scriptsize $56 \! \times \! 56 \! \times \! 256$};

\pgfmathsetmacro{\cubey}{\cubey/2}
\pgfmathsetmacro{\cubez}{\cubez/2}
\pgfmathsetmacro{\x}{\x + \cubex + \offset}
\pgfmathsetmacro{\xx}{\xx/\divyz}
\draw[maxline,fill=maxfill] (\x+\xx/2,\cubey/2,\cubez/2) -- ++(-\cubex,0,0) -- ++(0,-\cubey,0) -- ++(\cubex,0,0) -- cycle;
\draw[maxline,fill=maxfill] (\x+\xx/2,\cubey/2,\cubez/2) -- ++(0-\xx,0,-\cubez) -- ++(0,-\cubey,0) -- ++(0+\xx,0,\cubez) -- cycle;
\draw[maxline,fill=maxfill] (\x+\xx/2,\cubey/2,\cubez/2) -- ++(-\cubex,0,0) -- ++(0-\xx,0,-\cubez) -- ++(\cubex,0,0) -- cycle;

\pgfmathsetmacro{\cubex}{\cubex*\multx}
\pgfmathsetmacro{\x}{\x + \cubex + \offset}
\draw[convline,fill=convfill] (\x+\xx/2,\cubey/2,\cubez/2) -- ++(-\cubex,0,0) -- ++(0,-\cubey,0) -- ++(\cubex,0,0) -- cycle;
\draw[convline,fill=convfill] (\x+\xx/2,\cubey/2,\cubez/2) -- ++(0-\xx,0,-\cubez) -- ++(0,-\cubey,0) -- ++(0+\xx,0,\cubez) -- cycle;
\draw[convline,fill=convfill] (\x+\xx/2,\cubey/2,\cubez/2) -- ++(-\cubex,0,0) -- ++(0-\xx,0,-\cubez) -- ++(\cubex,0,0) -- cycle;

\pgfmathsetmacro{\x}{\x + \cubex + \offset}
\draw[convline,fill=convfill] (\x+\xx/2,\cubey/2,\cubez/2) -- ++(-\cubex,0,0) -- ++(0,-\cubey,0) -- ++(\cubex,0,0) -- cycle;
\draw[convline,fill=convfill] (\x+\xx/2,\cubey/2,\cubez/2) -- ++(0-\xx,0,-\cubez) -- ++(0,-\cubey,0) -- ++(0+\xx,0,\cubez) -- cycle;
\draw[convline,fill=convfill] (\x+\xx/2,\cubey/2,\cubez/2) -- ++(-\cubex,0,0) -- ++(0-\xx,0,-\cubez) -- ++(\cubex,0,0) -- cycle;

\pgfmathsetmacro{\x}{\x + \cubex + \offset}
\draw[convline,fill=convfill] (\x+\xx/2,\cubey/2,\cubez/2) -- ++(-\cubex,0,0) -- ++(0,-\cubey,0) -- ++(\cubex,0,0) -- cycle;
\draw[convline,fill=convfill] (\x+\xx/2,\cubey/2,\cubez/2) -- ++(0-\xx,0,-\cubez) -- ++(0,-\cubey,0) -- ++(0+\xx,0,\cubez) -- cycle;
\draw[convline,fill=convfill] (\x+\xx/2,\cubey/2,\cubez/2) -- ++(-\cubex,0,0) -- ++(0-\xx,0,-\cubez) -- ++(\cubex,0,0) -- cycle;

\node[text width=3cm] at (\x+\xx/2-\xx+0.05,\cubey/2+0.12,\cubez/2-\cubez) {\scriptsize $28 \! \times \! 28 \! \times \! 512$};

\pgfmathsetmacro{\cubey}{\cubey/2}
\pgfmathsetmacro{\cubez}{\cubez/2}
\pgfmathsetmacro{\x}{\x + \cubex + \offset}
\pgfmathsetmacro{\xx}{\xx/\divyz}
\draw[maxline,fill=maxfill] (\x+\xx/2,\cubey/2,\cubez/2) -- ++(-\cubex,0,0) -- ++(0,-\cubey,0) -- ++(\cubex,0,0) -- cycle;
\draw[maxline,fill=maxfill] (\x+\xx/2,\cubey/2,\cubez/2) -- ++(0-\xx,0,-\cubez) -- ++(0,-\cubey,0) -- ++(0+\xx,0,\cubez) -- cycle;
\draw[maxline,fill=maxfill] (\x+\xx/2,\cubey/2,\cubez/2) -- ++(-\cubex,0,0) -- ++(0-\xx,0,-\cubez) -- ++(\cubex,0,0) -- cycle;

\pgfmathsetmacro{\x}{\x + \cubex + \offset}
\draw[convline,fill=convfill] (\x+\xx/2,\cubey/2,\cubez/2) -- ++(-\cubex,0,0) -- ++(0,-\cubey,0) -- ++(\cubex,0,0) -- cycle;
\draw[convline,fill=convfill] (\x+\xx/2,\cubey/2,\cubez/2) -- ++(0-\xx,0,-\cubez) -- ++(0,-\cubey,0) -- ++(0+\xx,0,\cubez) -- cycle;
\draw[convline,fill=convfill] (\x+\xx/2,\cubey/2,\cubez/2) -- ++(-\cubex,0,0) -- ++(0-\xx,0,-\cubez) -- ++(\cubex,0,0) -- cycle;

\pgfmathsetmacro{\x}{\x + \cubex + \offset}
\draw[convline,fill=convfill] (\x+\xx/2,\cubey/2,\cubez/2) -- ++(-\cubex,0,0) -- ++(0,-\cubey,0) -- ++(\cubex,0,0) -- cycle;
\draw[convline,fill=convfill] (\x+\xx/2,\cubey/2,\cubez/2) -- ++(0-\xx,0,-\cubez) -- ++(0,-\cubey,0) -- ++(0+\xx,0,\cubez) -- cycle;
\draw[convline,fill=convfill] (\x+\xx/2,\cubey/2,\cubez/2) -- ++(-\cubex,0,0) -- ++(0-\xx,0,-\cubez) -- ++(\cubex,0,0) -- cycle;

\pgfmathsetmacro{\x}{\x + \cubex + \offset}
\draw[convline,fill=convfill] (\x+\xx/2,\cubey/2,\cubez/2) -- ++(-\cubex,0,0) -- ++(0,-\cubey,0) -- ++(\cubex,0,0) -- cycle;
\draw[convline,fill=convfill] (\x+\xx/2,\cubey/2,\cubez/2) -- ++(0-\xx,0,-\cubez) -- ++(0,-\cubey,0) -- ++(0+\xx,0,\cubez) -- cycle;
\draw[convline,fill=convfill] (\x+\xx/2,\cubey/2,\cubez/2) -- ++(-\cubex,0,0) -- ++(0-\xx,0,-\cubez) -- ++(\cubex,0,0) -- cycle;

\node[text width=3cm] at (\x+\xx/2-\xx-0.05,\cubey/2+0.12,\cubez/2-\cubez) {\scriptsize $14 \! \times \! 14 \! \times \! 512$};

\pgfmathsetmacro{\cubey}{\cubey/2}
\pgfmathsetmacro{\cubez}{\cubez/2}
\pgfmathsetmacro{\x}{\x + \cubex + \offset}
\pgfmathsetmacro{\xx}{\xx/\divyz}
\draw[maxline,fill=maxfill] (\x+\xx/2,\cubey/2,\cubez/2) -- ++(-\cubex,0,0) -- ++(0,-\cubey,0) -- ++(\cubex,0,0) -- cycle;
\draw[maxline,fill=maxfill] (\x+\xx/2,\cubey/2,\cubez/2) -- ++(0-\xx,0,-\cubez) -- ++(0,-\cubey,0) -- ++(0+\xx,0,\cubez) -- cycle;
\draw[maxline,fill=maxfill] (\x+\xx/2,\cubey/2,\cubez/2) -- ++(-\cubex,0,0) -- ++(0-\xx,0,-\cubez) -- ++(\cubex,0,0) -- cycle;

\node[text width=3cm] (sizepool5) at (\x+\xx/2-\xx+0.8,\cubey/2+0.5,0) {\scriptsize $7 \! \times \! 7 \! \times \! 512$};

\draw[-stealth] (\x+\xx/2-\xx -\cubex/2,\cubey/2+0.4,0)--(\x+\xx/2-\xx -\cubex/2,\cubey/2 ,0);

\pgfmathsetmacro{\cubex}{\cubex*2.5}
\pgfmathsetmacro{\cubey}{\cubey/2}
\pgfmathsetmacro{\cubez}{\cubez/2}
\pgfmathsetmacro{\x}{\x + \cubex + \offset}
\pgfmathsetmacro{\xx}{\xx/\divyz}
\draw[fcline,fill=fcfill] (\x+\xx/2,\cubey/2,\cubez/2) -- ++(-\cubex,0,0) -- ++(0,-\cubey,0) -- ++(\cubex,0,0) -- cycle;
\draw[fcline,fill=fcfill] (\x+\xx/2,\cubey/2,\cubez/2) -- ++(0-\xx,0,-\cubez) -- ++(0,-\cubey,0) -- ++(0+\xx,0,\cubez) -- cycle;
\draw[fcline,fill=fcfill] (\x+\xx/2,\cubey/2,\cubez/2) -- ++(-\cubex,0,0) -- ++(0-\xx,0,-\cubez) -- ++(\cubex,0,0) -- cycle;

\pgfmathsetmacro{\x}{\x + \cubex + \offset}
\draw[fcline,fill=fcfill] (\x+\xx/2,\cubey/2,\cubez/2) -- ++(-\cubex,0,0) -- ++(0,-\cubey,0) -- ++(\cubex,0,0) -- cycle;
\draw[fcline,fill=fcfill] (\x+\xx/2,\cubey/2,\cubez/2) -- ++(0-\xx,0,-\cubez) -- ++(0,-\cubey,0) -- ++(0+\xx,0,\cubez) -- cycle;
\draw[fcline,fill=fcfill] (\x+\xx/2,\cubey/2,\cubez/2) -- ++(-\cubex,0,0) -- ++(0-\xx,0,-\cubez) -- ++(\cubex,0,0) -- cycle;

\node[text width=3cm] at (\x+\xx/2-\xx-0.1,\cubey/2+0.12,\cubez/2-\cubez) {\scriptsize $1 \! \times \! 1 \! \times \! 4096$};

\pgfmathsetmacro{\cubex}{\cubex/2}
\pgfmathsetmacro{\x}{\x + \cubex + \offset}
\draw[fcline,fill=fcfill] (\x+\xx/2,\cubey/2,\cubez/2) -- ++(-\cubex,0,0) -- ++(0,-\cubey,0) -- ++(\cubex,0,0) -- cycle;
\draw[fcline,fill=fcfill] (\x+\xx/2,\cubey/2,\cubez/2) -- ++(0-\xx,0,-\cubez) -- ++(0,-\cubey,0) -- ++(0+\xx,0,\cubez) -- cycle;
\draw[fcline,fill=fcfill] (\x+\xx/2,\cubey/2,\cubez/2) -- ++(-\cubex,0,0) -- ++(0-\xx,0,-\cubez) -- ++(\cubex,0,0) -- cycle;

\pgfmathsetmacro{\x}{\x + \cubex + \offset}
\draw[softmaxline,fill=softmaxfill] (\x+\xx/2,\cubey/2,\cubez/2) -- ++(-\cubex,0,0) -- ++(0,-\cubey,0) -- ++(\cubex,0,0) -- cycle;
\draw[softmaxline,fill=softmaxfill] (\x+\xx/2,\cubey/2,\cubez/2) -- ++(0-\xx,0,-\cubez) -- ++(0,-\cubey,0) -- ++(0+\xx,0,\cubez) -- cycle;
\draw[softmaxline,fill=softmaxfill] (\x+\xx/2,\cubey/2,\cubez/2) -- ++(-\cubex,0,0) -- ++(0-\xx,0,-\cubez) -- ++(\cubex,0,0) -- cycle;

\node[text width=3cm] at (\x+\xx/2-\xx+0.5,\cubey/2+0.12,\cubez/2-\cubez) {\scriptsize $1 \! \times \! 1 \! \times \! 1000$};

\pgfmathsetmacro{\x}{\x/2}
\pgfmathsetmacro{\y}{2.5}
\pgfmathsetmacro{\cubex}{\cubex/1.5}
\pgfmathsetmacro{\cubey}{\cubey*3}
\pgfmathsetmacro{\cubez}{\cubez*3}

\draw[convline,fill=convfill] (\x+\xx/2,\y+\cubey/2,\cubez/2) -- ++(-\cubex,0,0) -- ++(0,-\cubey,0) -- ++(\cubex,0,0) -- cycle;
\draw[convline,fill=convfill] (\x+\xx/2,\y+\cubey/2,\cubez/2) -- ++(0-\xx,0,-\cubez) -- ++(0,-\cubey,0) -- ++(0+\xx,0,\cubez) -- cycle;
\draw[convline,fill=convfill] (\x+\xx/2,\y+\cubey/2,\cubez/2) -- ++(-\cubex,0,0) -- ++(0-\xx,0,-\cubez) -- ++(\cubex,0,0) -- cycle;

\node[text width=3cm] at (\x+1.7,\y+0.05,0) {\scriptsize convolution+ReLU};

\pgfmathsetmacro{\y}{\y-0.35}
\draw[maxline,fill=maxfill] (\x+\xx/2,\y+\cubey/2,\cubez/2) -- ++(-\cubex,0,0) -- ++(0,-\cubey,0) -- ++(\cubex,0,0) -- cycle;
\draw[maxline,fill=maxfill] (\x+\xx/2,\y+\cubey/2,\cubez/2) -- ++(0-\xx,0,-\cubez) -- ++(0,-\cubey,0) -- ++(0+\xx,0,\cubez) -- cycle;
\draw[maxline,fill=maxfill] (\x+\xx/2,\y+\cubey/2,\cubez/2) -- ++(-\cubex,0,0) -- ++(0-\xx,0,-\cubez) -- ++(\cubex,0,0) -- cycle;

\node[text width=3cm] at (\x+1.7,\y+0.05,0) {\scriptsize max pooling};

\pgfmathsetmacro{\y}{\y-0.35}
\draw[fcline,fill=fcfill] (\x+\xx/2,\y+\cubey/2,\cubez/2) -- ++(-\cubex,0,0) -- ++(0,-\cubey,0) -- ++(\cubex,0,0) -- cycle;
\draw[fcline,fill=fcfill] (\x+\xx/2,\y+\cubey/2,\cubez/2) -- ++(0-\xx,0,-\cubez) -- ++(0,-\cubey,0) -- ++(0+\xx,0,\cubez) -- cycle;
\draw[fcline,fill=fcfill] (\x+\xx/2,\y+\cubey/2,\cubez/2) -- ++(-\cubex,0,0) -- ++(0-\xx,0,-\cubez) -- ++(\cubex,0,0) -- cycle;

\node[text width=3cm] at (\x+1.7,\y+0.05,0) {\scriptsize fully connected+ReLU};

\pgfmathsetmacro{\y}{\y-0.35}
\draw[softmaxline,fill=softmaxfill] (\x+\xx/2,\y+\cubey/2,\cubez/2) -- ++(-\cubex,0,0) -- ++(0,-\cubey,0) -- ++(\cubex,0,0) -- cycle;
\draw[softmaxline,fill=softmaxfill] (\x+\xx/2,\y+\cubey/2,\cubez/2) -- ++(0-\xx,0,-\cubez) -- ++(0,-\cubey,0) -- ++(0+\xx,0,\cubez) -- cycle;
\draw[softmaxline,fill=softmaxfill] (\x+\xx/2,\y+\cubey/2,\cubez/2) -- ++(-\cubex,0,0) -- ++(0-\xx,0,-\cubez) -- ++(\cubex,0,0) -- cycle;

\node[text width=4cm] at (\x+2.2,\y+0.05,0) {\scriptsize softmax (1000 object classes)};

\draw[draw=gray, dash dot] (\x - 6.5,\y-4.7) -- (\x + 4.5,\y-4.7);

\draw[-stealth,line width=0.4mm] (\x,\y-4.8) to (\x,\y-5.7);
\node at (\x+2,\y-5.2) {Transfer Learning};

\end{tikzpicture} 

\begin{tikzpicture}

\pgfmathsetmacro{\cubex}{0.1}
\pgfmathsetmacro{\cubey}{4}
\pgfmathsetmacro{\cubez}{5}

\pgfmathsetmacro{\multx}{1.5}
\pgfmathsetmacro{\divyz}{2}

\definecolor{convfill}{RGB}{255,255,255}
\definecolor{convline}{RGB}{0,0,0}

\definecolor{maxline}{RGB}{255,0,0}
\definecolor{maxfill}{RGB}{255,255,255}

\definecolor{fcfill}{RGB}{255,255,255}
\definecolor{fcline}{RGB}{6,185,251}

\definecolor{mfcfill}{RGB}{0,0,255}
\definecolor{mfcline}{RGB}{0,0,255}

\definecolor{softmaxfill}{RGB}{255,255,255}
\definecolor{softmaxline}{RGB}{255,255,0}

\pgfmathsetmacro{\x}{0}
\pgfmathsetmacro{\offset}{0.08}

\pgfmathsetmacro{\xx}{0.5}

\node[inner sep=0pt,cm={0.37 ,0.5 ,0 ,1  ,(0 cm,0 cm)}] (image) at (-1.6,0)
    {\includegraphics[width=4 cm]{2370477712.png}};
    
\node at (-1.8,3.5) {\footnotesize{Privacy Dataset (PicAlert)}};
\node at (-1.8,3.75) {\footnotesize{Image}};

\node at (4.5,3.5) {\small{VGG-16 CNN fine-tuned on PicAlert}};
  
\node[text width=3cm] at (\x+\xx/2-\xx-1.2,\cubey/2+0.12,\cubez/2-\cubez) {\scriptsize $224 \! \times \! 224 \! \times \! 3$};

\draw[convline,fill=convfill] (\x+\xx/2,\cubey/2,\cubez/2) -- ++(-\cubex,0,0) -- ++(0,-\cubey,0) -- ++(\cubex,0,0) -- cycle;
\draw[convline,fill=convfill] (\x+\xx/2,\cubey/2,\cubez/2) -- ++(0-\xx,0,-\cubez) -- ++(0,-\cubey,0) -- ++(0+\xx,0,\cubez) -- cycle;
\draw[convline,fill=convfill] (\x+\xx/2,\cubey/2,\cubez/2) -- ++(-\cubex,0,0) -- ++(0-\xx,0,-\cubez) -- ++(\cubex,0,0) -- cycle;

\pgfmathsetmacro{\x}{\x + \cubex + \offset}
\draw[convline,fill=convfill] (\x+\xx/2,\cubey/2,\cubez/2) -- ++(-\cubex,0,0) -- ++(0,-\cubey,0) -- ++(\cubex,0,0) -- cycle;
\draw[convline,fill=convfill] (\x+\xx/2,\cubey/2,\cubez/2) -- ++(0-\xx,0,-\cubez) -- ++(0,-\cubey,0) -- ++(0+\xx,0,\cubez) -- cycle;
\draw[convline,fill=convfill] (\x+\xx/2,\cubey/2,\cubez/2) -- ++(-\cubex,0,0) -- ++(0-\xx,0,-\cubez) -- ++(\cubex,0,0) -- cycle;

\node[text width=3cm] at (\x+\xx/2-\xx+0.7,\cubey/2+0.12,\cubez/2-\cubez) {\scriptsize $224 \! \times \! 224 \! \times \! 64$};

\pgfmathsetmacro{\cubey}{\cubey/\divyz}
\pgfmathsetmacro{\cubez}{\cubez/\divyz}
\pgfmathsetmacro{\x}{\x + \cubex + \offset}
\pgfmathsetmacro{\xx}{\xx/\divyz}
\draw[maxline,fill=maxfill] (\x+\xx/2,\cubey/2,\cubez/2) -- ++(-\cubex,0,0) -- ++(0,-\cubey,0) -- ++(\cubex,0,0) -- cycle;
\draw[maxline,fill=maxfill] (\x+\xx/2,\cubey/2,\cubez/2) -- ++(0-\xx,0,-\cubez) -- ++(0,-\cubey,0) -- ++(0+\xx,0,\cubez) -- cycle;
\draw[maxline,fill=maxfill] (\x+\xx/2,\cubey/2,\cubez/2) -- ++(-\cubex,0,0) -- ++(0-\xx,0,-\cubez) -- ++(\cubex,0,0) -- cycle;

\pgfmathsetmacro{\cubex}{\cubex*\multx}
\pgfmathsetmacro{\x}{\x + \cubex + \offset}
\draw[convline,fill=convfill] (\x+\xx/2,\cubey/2,\cubez/2) -- ++(-\cubex,0,0) -- ++(0,-\cubey,0) -- ++(\cubex,0,0) -- cycle;
\draw[convline,fill=convfill] (\x+\xx/2,\cubey/2,\cubez/2) -- ++(0-\xx,0,-\cubez) -- ++(0,-\cubey,0) -- ++(0+\xx,0,\cubez) -- cycle;
\draw[convline,fill=convfill] (\x+\xx/2,\cubey/2,\cubez/2) -- ++(-\cubex,0,0) -- ++(0-\xx,0,-\cubez) -- ++(\cubex,0,0) -- cycle;

\pgfmathsetmacro{\x}{\x + \cubex + \offset}
\draw[convline,fill=convfill] (\x+\xx/2,\cubey/2,\cubez/2) -- ++(-\cubex,0,0) -- ++(0,-\cubey,0) -- ++(\cubex,0,0) -- cycle;
\draw[convline,fill=convfill] (\x+\xx/2,\cubey/2,\cubez/2) -- ++(0-\xx,0,-\cubez) -- ++(0,-\cubey,0) -- ++(0+\xx,0,\cubez) -- cycle;
\draw[convline,fill=convfill] (\x+\xx/2,\cubey/2,\cubez/2) -- ++(-\cubex,0,0) -- ++(0-\xx,0,-\cubez) -- ++(\cubex,0,0) -- cycle;

\node[text width=3cm] at (\x+\xx/2-\xx+0.55,\cubey/2+0.12,\cubez/2-\cubez) {\scriptsize $112 \! \times \! 112 \! \times \! 128$};

\pgfmathsetmacro{\cubey}{\cubey/\divyz}
\pgfmathsetmacro{\cubez}{\cubez/\divyz}
\pgfmathsetmacro{\x}{\x + \cubex + \offset}
\pgfmathsetmacro{\xx}{\xx/\divyz}
\draw[maxline,fill=maxfill] (\x+\xx/2,\cubey/2,\cubez/2) -- ++(-\cubex,0,0) -- ++(0,-\cubey,0) -- ++(\cubex,0,0) -- cycle;
\draw[maxline,fill=maxfill] (\x+\xx/2,\cubey/2,\cubez/2) -- ++(0-\xx,0,-\cubez) -- ++(0,-\cubey,0) -- ++(0+\xx,0,\cubez) -- cycle;
\draw[maxline,fill=maxfill] (\x+\xx/2,\cubey/2,\cubez/2) -- ++(-\cubex,0,0) -- ++(0-\xx,0,-\cubez) -- ++(\cubex,0,0) -- cycle;

\pgfmathsetmacro{\cubex}{\cubex*\multx}
\pgfmathsetmacro{\x}{\x + \cubex + \offset}
\draw[convline,fill=convfill] (\x+\xx/2,\cubey/2,\cubez/2) -- ++(-\cubex,0,0) -- ++(0,-\cubey,0) -- ++(\cubex,0,0) -- cycle;
\draw[convline,fill=convfill] (\x+\xx/2,\cubey/2,\cubez/2) -- ++(0-\xx,0,-\cubez) -- ++(0,-\cubey,0) -- ++(0+\xx,0,\cubez) -- cycle;
\draw[convline,fill=convfill] (\x+\xx/2,\cubey/2,\cubez/2) -- ++(-\cubex,0,0) -- ++(0-\xx,0,-\cubez) -- ++(\cubex,0,0) -- cycle;

\pgfmathsetmacro{\x}{\x + \cubex + \offset}
\draw[convline,fill=convfill] (\x+\xx/2,\cubey/2,\cubez/2) -- ++(-\cubex,0,0) -- ++(0,-\cubey,0) -- ++(\cubex,0,0) -- cycle;
\draw[convline,fill=convfill] (\x+\xx/2,\cubey/2,\cubez/2) -- ++(0-\xx,0,-\cubez) -- ++(0,-\cubey,0) -- ++(0+\xx,0,\cubez) -- cycle;
\draw[convline,fill=convfill] (\x+\xx/2,\cubey/2,\cubez/2) -- ++(-\cubex,0,0) -- ++(0-\xx,0,-\cubez) -- ++(\cubex,0,0) -- cycle;

\pgfmathsetmacro{\x}{\x + \cubex + \offset}
\draw[convline,fill=convfill] (\x+\xx/2,\cubey/2,\cubez/2) -- ++(-\cubex,0,0) -- ++(0,-\cubey,0) -- ++(\cubex,0,0) -- cycle;
\draw[convline,fill=convfill] (\x+\xx/2,\cubey/2,\cubez/2) -- ++(0-\xx,0,-\cubez) -- ++(0,-\cubey,0) -- ++(0+\xx,0,\cubez) -- cycle;
\draw[convline,fill=convfill] (\x+\xx/2,\cubey/2,\cubez/2) -- ++(-\cubex,0,0) -- ++(0-\xx,0,-\cubez) -- ++(\cubex,0,0) -- cycle;

\node[text width=3cm] at (\x+\xx/2-\xx+0.23,\cubey/2+0.12,\cubez/2-\cubez) {\scriptsize $56 \! \times \! 56 \! \times \! 256$};

\pgfmathsetmacro{\cubey}{\cubey/2}
\pgfmathsetmacro{\cubez}{\cubez/2}
\pgfmathsetmacro{\x}{\x + \cubex + \offset}
\pgfmathsetmacro{\xx}{\xx/\divyz}
\draw[maxline,fill=maxfill] (\x+\xx/2,\cubey/2,\cubez/2) -- ++(-\cubex,0,0) -- ++(0,-\cubey,0) -- ++(\cubex,0,0) -- cycle;
\draw[maxline,fill=maxfill] (\x+\xx/2,\cubey/2,\cubez/2) -- ++(0-\xx,0,-\cubez) -- ++(0,-\cubey,0) -- ++(0+\xx,0,\cubez) -- cycle;
\draw[maxline,fill=maxfill] (\x+\xx/2,\cubey/2,\cubez/2) -- ++(-\cubex,0,0) -- ++(0-\xx,0,-\cubez) -- ++(\cubex,0,0) -- cycle;

\pgfmathsetmacro{\cubex}{\cubex*\multx}
\pgfmathsetmacro{\x}{\x + \cubex + \offset}
\draw[convline,fill=convfill] (\x+\xx/2,\cubey/2,\cubez/2) -- ++(-\cubex,0,0) -- ++(0,-\cubey,0) -- ++(\cubex,0,0) -- cycle;
\draw[convline,fill=convfill] (\x+\xx/2,\cubey/2,\cubez/2) -- ++(0-\xx,0,-\cubez) -- ++(0,-\cubey,0) -- ++(0+\xx,0,\cubez) -- cycle;
\draw[convline,fill=convfill] (\x+\xx/2,\cubey/2,\cubez/2) -- ++(-\cubex,0,0) -- ++(0-\xx,0,-\cubez) -- ++(\cubex,0,0) -- cycle;

\pgfmathsetmacro{\x}{\x + \cubex + \offset}
\draw[convline,fill=convfill] (\x+\xx/2,\cubey/2,\cubez/2) -- ++(-\cubex,0,0) -- ++(0,-\cubey,0) -- ++(\cubex,0,0) -- cycle;
\draw[convline,fill=convfill] (\x+\xx/2,\cubey/2,\cubez/2) -- ++(0-\xx,0,-\cubez) -- ++(0,-\cubey,0) -- ++(0+\xx,0,\cubez) -- cycle;
\draw[convline,fill=convfill] (\x+\xx/2,\cubey/2,\cubez/2) -- ++(-\cubex,0,0) -- ++(0-\xx,0,-\cubez) -- ++(\cubex,0,0) -- cycle;

\pgfmathsetmacro{\x}{\x + \cubex + \offset}
\draw[convline,fill=convfill] (\x+\xx/2,\cubey/2,\cubez/2) -- ++(-\cubex,0,0) -- ++(0,-\cubey,0) -- ++(\cubex,0,0) -- cycle;
\draw[convline,fill=convfill] (\x+\xx/2,\cubey/2,\cubez/2) -- ++(0-\xx,0,-\cubez) -- ++(0,-\cubey,0) -- ++(0+\xx,0,\cubez) -- cycle;
\draw[convline,fill=convfill] (\x+\xx/2,\cubey/2,\cubez/2) -- ++(-\cubex,0,0) -- ++(0-\xx,0,-\cubez) -- ++(\cubex,0,0) -- cycle;

\node[text width=3cm] at (\x+\xx/2-\xx+0.05,\cubey/2+0.12,\cubez/2-\cubez) {\scriptsize $28 \! \times \! 28 \! \times \! 512$};

\pgfmathsetmacro{\cubey}{\cubey/2}
\pgfmathsetmacro{\cubez}{\cubez/2}
\pgfmathsetmacro{\x}{\x + \cubex + \offset}
\pgfmathsetmacro{\xx}{\xx/\divyz}
\draw[maxline,fill=maxfill] (\x+\xx/2,\cubey/2,\cubez/2) -- ++(-\cubex,0,0) -- ++(0,-\cubey,0) -- ++(\cubex,0,0) -- cycle;
\draw[maxline,fill=maxfill] (\x+\xx/2,\cubey/2,\cubez/2) -- ++(0-\xx,0,-\cubez) -- ++(0,-\cubey,0) -- ++(0+\xx,0,\cubez) -- cycle;
\draw[maxline,fill=maxfill] (\x+\xx/2,\cubey/2,\cubez/2) -- ++(-\cubex,0,0) -- ++(0-\xx,0,-\cubez) -- ++(\cubex,0,0) -- cycle;

\pgfmathsetmacro{\x}{\x + \cubex + \offset}
\draw[convline,fill=convfill] (\x+\xx/2,\cubey/2,\cubez/2) -- ++(-\cubex,0,0) -- ++(0,-\cubey,0) -- ++(\cubex,0,0) -- cycle;
\draw[convline,fill=convfill] (\x+\xx/2,\cubey/2,\cubez/2) -- ++(0-\xx,0,-\cubez) -- ++(0,-\cubey,0) -- ++(0+\xx,0,\cubez) -- cycle;
\draw[convline,fill=convfill] (\x+\xx/2,\cubey/2,\cubez/2) -- ++(-\cubex,0,0) -- ++(0-\xx,0,-\cubez) -- ++(\cubex,0,0) -- cycle;

\pgfmathsetmacro{\x}{\x + \cubex + \offset}
\draw[convline,fill=convfill] (\x+\xx/2,\cubey/2,\cubez/2) -- ++(-\cubex,0,0) -- ++(0,-\cubey,0) -- ++(\cubex,0,0) -- cycle;
\draw[convline,fill=convfill] (\x+\xx/2,\cubey/2,\cubez/2) -- ++(0-\xx,0,-\cubez) -- ++(0,-\cubey,0) -- ++(0+\xx,0,\cubez) -- cycle;
\draw[convline,fill=convfill] (\x+\xx/2,\cubey/2,\cubez/2) -- ++(-\cubex,0,0) -- ++(0-\xx,0,-\cubez) -- ++(\cubex,0,0) -- cycle;

\pgfmathsetmacro{\x}{\x + \cubex + \offset}
\draw[convline,fill=convfill] (\x+\xx/2,\cubey/2,\cubez/2) -- ++(-\cubex,0,0) -- ++(0,-\cubey,0) -- ++(\cubex,0,0) -- cycle;
\draw[convline,fill=convfill] (\x+\xx/2,\cubey/2,\cubez/2) -- ++(0-\xx,0,-\cubez) -- ++(0,-\cubey,0) -- ++(0+\xx,0,\cubez) -- cycle;
\draw[convline,fill=convfill] (\x+\xx/2,\cubey/2,\cubez/2) -- ++(-\cubex,0,0) -- ++(0-\xx,0,-\cubez) -- ++(\cubex,0,0) -- cycle;

\node[text width=3cm] at (\x+\xx/2-\xx-0.05,\cubey/2+0.12,\cubez/2-\cubez) {\scriptsize $14 \! \times \! 14 \! \times \! 512$};

\pgfmathsetmacro{\cubey}{\cubey/2}
\pgfmathsetmacro{\cubez}{\cubez/2}
\pgfmathsetmacro{\x}{\x + \cubex + \offset}
\pgfmathsetmacro{\xx}{\xx/\divyz}
\draw[maxline,fill=maxfill] (\x+\xx/2,\cubey/2,\cubez/2) -- ++(-\cubex,0,0) -- ++(0,-\cubey,0) -- ++(\cubex,0,0) -- cycle;
\draw[maxline,fill=maxfill] (\x+\xx/2,\cubey/2,\cubez/2) -- ++(0-\xx,0,-\cubez) -- ++(0,-\cubey,0) -- ++(0+\xx,0,\cubez) -- cycle;
\draw[maxline,fill=maxfill] (\x+\xx/2,\cubey/2,\cubez/2) -- ++(-\cubex,0,0) -- ++(0-\xx,0,-\cubez) -- ++(\cubex,0,0) -- cycle;

\node[text width=3cm] (sizepool5) at (\x+\xx/2-\xx+0.8,\cubey/2+0.5,0) {\scriptsize $7 \! \times \! 7 \! \times \! 512$};

\draw[-stealth] (\x+\xx/2-\xx -\cubex/2,\cubey/2+0.4,0)--(\x+\xx/2-\xx -\cubex/2,\cubey/2 ,0);

\pgfmathsetmacro{\cubex}{\cubex*2.5}
\pgfmathsetmacro{\cubey}{\cubey/2}
\pgfmathsetmacro{\cubez}{\cubez/2}
\pgfmathsetmacro{\x}{\x + \cubex + \offset}
\pgfmathsetmacro{\xx}{\xx/\divyz}
\draw[fcline,fill=fcfill] (\x+\xx/2,\cubey/2,\cubez/2) -- ++(-\cubex,0,0) -- ++(0,-\cubey,0) -- ++(\cubex,0,0) -- cycle;
\draw[fcline,fill=fcfill] (\x+\xx/2,\cubey/2,\cubez/2) -- ++(0-\xx,0,-\cubez) -- ++(0,-\cubey,0) -- ++(0+\xx,0,\cubez) -- cycle;
\draw[fcline,fill=fcfill] (\x+\xx/2,\cubey/2,\cubez/2) -- ++(-\cubex,0,0) -- ++(0-\xx,0,-\cubez) -- ++(\cubex,0,0) -- cycle;

\pgfmathsetmacro{\x}{\x + \cubex + \offset}
\draw[fcline,fill=fcfill] (\x+\xx/2,\cubey/2,\cubez/2) -- ++(-\cubex,0,0) -- ++(0,-\cubey,0) -- ++(\cubex,0,0) -- cycle;
\draw[fcline,fill=fcfill] (\x+\xx/2,\cubey/2,\cubez/2) -- ++(0-\xx,0,-\cubez) -- ++(0,-\cubey,0) -- ++(0+\xx,0,\cubez) -- cycle;
\draw[fcline,fill=fcfill] (\x+\xx/2,\cubey/2,\cubez/2) -- ++(-\cubex,0,0) -- ++(0-\xx,0,-\cubez) -- ++(\cubex,0,0) -- cycle;

\node[text width=3cm] at (\x+\xx/2-\xx-0.1,\cubey/2+0.12,\cubez/2-\cubez) {\scriptsize $1 \! \times \! 1 \! \times \! 4096$};

\pgfmathsetmacro{\cubex}{\cubex/3}
\pgfmathsetmacro{\x}{\x + \cubex + \offset}
\draw[mfcline,fill=fcfill] (\x+\xx/2,\cubey/2,\cubez/2) -- ++(-\cubex,0,0) -- ++(0,-\cubey,0) -- ++(\cubex,0,0) -- cycle;
\draw[mfcline,fill=fcfill] (\x+\xx/2,\cubey/2,\cubez/2) -- ++(0-\xx,0,-\cubez) -- ++(0,-\cubey,0) -- ++(0+\xx,0,\cubez) -- cycle;
\draw[mfcline,fill=fcfill] (\x+\xx/2,\cubey/2,\cubez/2) -- ++(-\cubex,0,0) -- ++(0-\xx,0,-\cubez) -- ++(\cubex,0,0) -- cycle;

\pgfmathsetmacro{\x}{\x + \cubex + \offset}
\draw[softmaxline,fill=softmaxfill] (\x+\xx/2,\cubey/2,\cubez/2) -- ++(-\cubex,0,0) -- ++(0,-\cubey,0) -- ++(\cubex,0,0) -- cycle;
\draw[softmaxline,fill=softmaxfill] (\x+\xx/2,\cubey/2,\cubez/2) -- ++(0-\xx,0,-\cubez) -- ++(0,-\cubey,0) -- ++(0+\xx,0,\cubez) -- cycle;
\draw[softmaxline,fill=softmaxfill] (\x+\xx/2,\cubey/2,\cubez/2) -- ++(-\cubex,0,0) -- ++(0-\xx,0,-\cubez) -- ++(\cubex,0,0) -- cycle;

\node[text width=3cm] at (\x+\xx/2-\xx+0.75,\cubey/2+0.12,\cubez/2-\cubez) {\scriptsize $1 \! \times \! 1 \! \times \! 2$};

\pgfmathsetmacro{\x}{\x/2}
\pgfmathsetmacro{\y}{2.5}
\pgfmathsetmacro{\cubex}{\cubex/1.5}
\pgfmathsetmacro{\cubey}{\cubey*3}
\pgfmathsetmacro{\cubez}{\cubez*3}

\draw[convline,fill=convfill] (\x+\xx/2,\y+\cubey/2,\cubez/2) -- ++(-\cubex,0,0) -- ++(0,-\cubey,0) -- ++(\cubex,0,0) -- cycle;
\draw[convline,fill=convfill] (\x+\xx/2,\y+\cubey/2,\cubez/2) -- ++(0-\xx,0,-\cubez) -- ++(0,-\cubey,0) -- ++(0+\xx,0,\cubez) -- cycle;
\draw[convline,fill=convfill] (\x+\xx/2,\y+\cubey/2,\cubez/2) -- ++(-\cubex,0,0) -- ++(0-\xx,0,-\cubez) -- ++(\cubex,0,0) -- cycle;

\node[text width=3cm] at (\x+1.7,\y+0.05,0) {\scriptsize convolution+ReLU};

\pgfmathsetmacro{\y}{\y-0.35}
\draw[maxline,fill=maxfill] (\x+\xx/2,\y+\cubey/2,\cubez/2) -- ++(-\cubex,0,0) -- ++(0,-\cubey,0) -- ++(\cubex,0,0) -- cycle;
\draw[maxline,fill=maxfill] (\x+\xx/2,\y+\cubey/2,\cubez/2) -- ++(0-\xx,0,-\cubez) -- ++(0,-\cubey,0) -- ++(0+\xx,0,\cubez) -- cycle;
\draw[maxline,fill=maxfill] (\x+\xx/2,\y+\cubey/2,\cubez/2) -- ++(-\cubex,0,0) -- ++(0-\xx,0,-\cubez) -- ++(\cubex,0,0) -- cycle;

\node[text width=3cm] at (\x+1.7,\y+0.05,0) {\scriptsize max pooling};

\pgfmathsetmacro{\y}{\y-0.35}
\draw[fcline,fill=fcfill] (\x+\xx/2,\y+\cubey/2,\cubez/2) -- ++(-\cubex,0,0) -- ++(0,-\cubey,0) -- ++(\cubex,0,0) -- cycle;
\draw[fcline,fill=fcfill] (\x+\xx/2,\y+\cubey/2,\cubez/2) -- ++(0-\xx,0,-\cubez) -- ++(0,-\cubey,0) -- ++(0+\xx,0,\cubez) -- cycle;
\draw[fcline,fill=fcfill] (\x+\xx/2,\y+\cubey/2,\cubez/2) -- ++(-\cubex,0,0) -- ++(0-\xx,0,-\cubez) -- ++(\cubex,0,0) -- cycle;

\node[text width=3cm] at (\x+1.7,\y+0.05,0) {\scriptsize fully connected+ReLU};

\pgfmathsetmacro{\y}{\y-0.35}
\draw[mfcline,fill=fcfill] (\x+\xx/2,\y+\cubey/2,\cubez/2) -- ++(-\cubex,0,0) -- ++(0,-\cubey,0) -- ++(\cubex,0,0) -- cycle;
\draw[mfcline,fill=fcfill] (\x+\xx/2,\y+\cubey/2,\cubez/2) -- ++(0-\xx,0,-\cubez) -- ++(0,-\cubey,0) -- ++(0+\xx,0,\cubez) -- cycle;
\draw[mfcline,fill=fcfill] (\x+\xx/2,\y+\cubey/2,\cubez/2) -- ++(-\cubex,0,0) -- ++(0-\xx,0,-\cubez) -- ++(\cubex,0,0) -- cycle;

\node[text width=4cm] at (\x+2.15,\y+0.05,0) {\scriptsize modified fully connected layer (fc$_8$-P)};

\pgfmathsetmacro{\y}{\y-0.35}
\draw[softmaxline,fill=softmaxfill] (\x+\xx/2,\y+\cubey/2,\cubez/2) -- ++(-\cubex,0,0) -- ++(0,-\cubey,0) -- ++(\cubex,0,0) -- cycle;
\draw[softmaxline,fill=softmaxfill] (\x+\xx/2,\y+\cubey/2,\cubez/2) -- ++(0-\xx,0,-\cubez) -- ++(0,-\cubey,0) -- ++(0+\xx,0,\cubez) -- cycle;
\draw[softmaxline,fill=softmaxfill] (\x+\xx/2,\y+\cubey/2,\cubez/2) -- ++(-\cubex,0,0) -- ++(0-\xx,0,-\cubez) -- ++(\cubex,0,0) -- cycle;

\node[text width=3cm] at (\x+1.7,\y+0.05,0) {\scriptsize softmax (2 privacy classes)};





\draw[-stealth] (\x+4, \y-1.2) -- (\x+4,\y-1.5);
\node at (\x+4,\y-1.9){\scriptsize {\color{red}Private} / {\color{ao}Public}};

\draw[-stealth] (\x+3.5, \y-0.8) -- (\x+3.5,\y-0.5);
\node at (\x+3.5,\y-0.35){\scriptsize {fc$_8$-P}};

\draw[draw=gray, dash dot] (\x - 6.5,\y+2.9) -- (\x + 4.5,\y+2.9);

\end{tikzpicture}
\caption{\label{fig:finetuned} Image encoding using fine-tuned CNN: (1) We modify the last fully-connected layer of the pre-trained network (top network) by changing the output units from $1000$ (object categories) to $2$ (privacy classes). (2) To train the modified network (bottom network) on privacy dataset, we first adopt weights of all the layers of the pre-trained network as initial weights and then iterate through all the layers using privacy data. (3) To make a prediction for an input image (privacy dataset), we use the probability distribution over $2$ privacy classes (softmax layer, yellow rectangle) for the input image obtained by applying the softmax function over the last modified fully-connected layer (fc$_8$-P, bottom network) of the fine-tuned network.
}
\end{figure*}

\subsection{Fine-tuned CNN} 
For this type of encoding, models trained on a large dataset (e.g., the ImageNet dataset) are fine-tuned using a smaller dataset (e.g., the privacy-labeled dataset). Fine-tuning a network is a procedure based on the concept of transfer learning \cite{donahue13, journals/jmlr/Bengio12}. This strategy fine-tunes the weights of the pre-trained network by continuing the back-propagation on the small dataset, i.e., privacy dataset in our scenario. The features become more dataset-specific after fine-tuning, and hence, are distinct from the features obtained from the pre-trained CNN. We modify the last fully-connected layer 
of all four network architectures, AlexNet, GoogLeNet, VGG-16, and ResNet by changing the output units from $1000$ (object categories) to $2$ (with respect to privacy classes) (e.g., changing fc$_8$ with 1000 output units to fc$_8$-P with 2 output units). We initialize the weights of all the layers of this modified architectures with the weights of the respective layers obtained from the pre-trained networks. We train the network by iterating through all the layers of the networks using the privacy data. We use the softmax function to predict the privacy of an image. Precisely, we use the probability distribution over $2$ privacy classes for the input image obtained by applying the softmax function over the modified last fully-connected layer (e.g. fc$_8$-P in VGG-16) of the fine-tuned networks (See Figure \ref{fig:finetuned}, second network, blue rectangle). 
The conditional probability distribution  over $2$ privacy classes  
can be defined using a softmax function as given below: 
\[P({y=P_r}|{\bf z})=\frac{exp(z_{P_r})}{exp(z_{P_u}) + exp(z_{P_r})}, P({y=P_u}|{\bf z})=\frac{exp(z_{P_u})}{exp(z_{P_u}) + exp(z_{P_r})} \] 
where, in our case, ${\bf z}$ is the output of the modified last fully connected layer (e.g., the fc$_8$-P layer of VGG-16) and $P_r$ and $P_u$ denote {\em private} and {\em public} class,  respectively. The fine-tuning process using VGG-16 is shown in Figure \ref{fig:finetuned}. 

 \begin{figure*}[t]
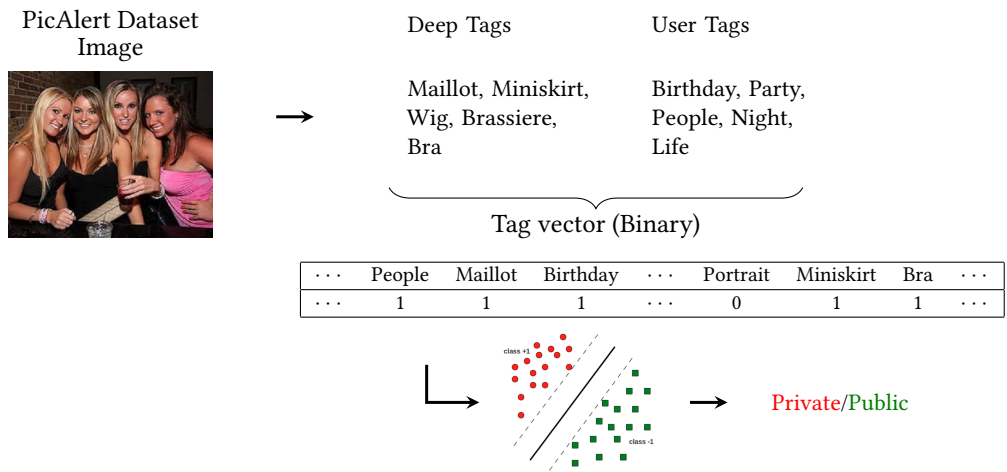

\centering

 \tikzstyle{rectangle}=[draw=gray,thick]
 \tikzstyle{entity}=[fill=gray!25,draw=gray!50,thick,text width=0.54cm, font=\footnotesize, text centered] 
  \tikzstyle{entity2}=[fill=gray!25,draw=gray!50,thick,text width=1cm, font=\footnotesize, text centered] 
    \tikzstyle{entity4}=[fill=gray!25,draw=gray!50,thick,text width=1.45cm, font=\footnotesize, text centered] 
    \tikzstyle{entity5}=[draw=gray!50,thick,text width=2.45cm, font=\footnotesize, text centered] 
    \tikzstyle{entity3}=[draw=black,fill=gray!25,semithick,text width=0.12cm, text height=0.12cm, font=\footnotesize, text centered] 
\tikzstyle{cut}=[circle,draw=black,fill=gray!70,semithick, inner sep=0pt,minimum size=2mm]
\tikzstyle{node}=[circle,draw=black,semithick,inner sep=0pt,minimum size=2mm] 
\tikzstyle{cut}=[circle,draw=black,fill=gray!50,thin, inner sep=0pt,minimum size=4mm]   
\tikzstyle{node3}=[circle,draw=gray!80,thick,inner sep=0pt,minimum size=3mm] 
\tikzstyle{node4}=[circle,draw=black,thick,inner sep=0pt,minimum size=3mm] 
\tikzstyle{node5}=[circle,draw=magenta!80!black,thick,inner sep=0pt,minimum size=3mm] 
\tikzstyle{db}=[cylinder,draw=black,shape border rotate=90, minimum height=60, minimum width=70, outer sep=-0.5\pgflinewidth]

 
\begin{tikzpicture}


\node at (-13.2,2.1) {\large {PicAlert Dataset}};
\node at (-13.2,1.7) {\large {Image}};

\node at (-13.2,0.3) {\includegraphics[scale=0.35]{2370477712.png}};

\node at (-7.0,-3.0) {\includegraphics[scale=0.12]{sl5}};


\newcommand{\Depth}{0.2}
\newcommand{\Height}{3.0}
\newcommand{\Width}{1}
\newcommand{\XO}{-10.5}
\newcommand{\YO}{1.3}

\node[text width=2.5cm] at (-8,2) {Deep Tags};
\node[text width=2.5cm] at (-8,0.8) {Maillot, Miniskirt, Wig, Brassiere, Bra};

\node[text width=2cm] at (-5,2) {User Tags};
\node[text width=2cm] at (-5,0.8) {Birthday, Party, People, Night, Life};

\draw[-stealth,line width=.4mm] (-11.0,0.8) to (-10.5,0.8);

\node (tab1) at (-6.0,-1.5) {%
	\begin{small}

  \begin{tabular}{|ccccccccc|}
  \hline
  $\cdots$ & People & Maillot & Birthday & $\cdots$ & Portrait & Miniskirt & Bra & $\cdots$ \\
  \hline 
  $\cdots$ & 1 & 1 & 1 & $\cdots$ & 0 & 1 & 1 & $\cdots$\\
  \hline
  
  \end{tabular}
	\end{small}  
};

\draw [decorate,decoration={brace,amplitude=8pt,mirror,raise=4pt},yshift=0pt]
(-9.5,0) -- (-4,0) node [black,midway,yshift=-0.65cm] {\large Tag vector (Binary)};

\draw[-stealth, line width=.4mm] (-9,-2.5) to (-9,-3.0) to (-8.3, -3.0);
\draw[-stealth, line width=.4mm] (-5.5,-3.0) to (-5.0,-3.0);

\node at (-3.5,-3.0) {{\color{red} Private}/{\color{ao} Public}};

\end{tikzpicture} 
\caption{\label{fig:tagfeat} Image encoding using tag features: We encode the combination of user tags and deep tags using binary vector representation, showing presence and absence of tags from tag vocabulary $V$. We set $1$ if a tag is present in the tag set or $0$ otherwise. We refer this model as Bag-of-Tags (BoT) model. 
}
\end{figure*}

\begin{figure*}[t]
\centering
\includegraphics[scale=0.6]{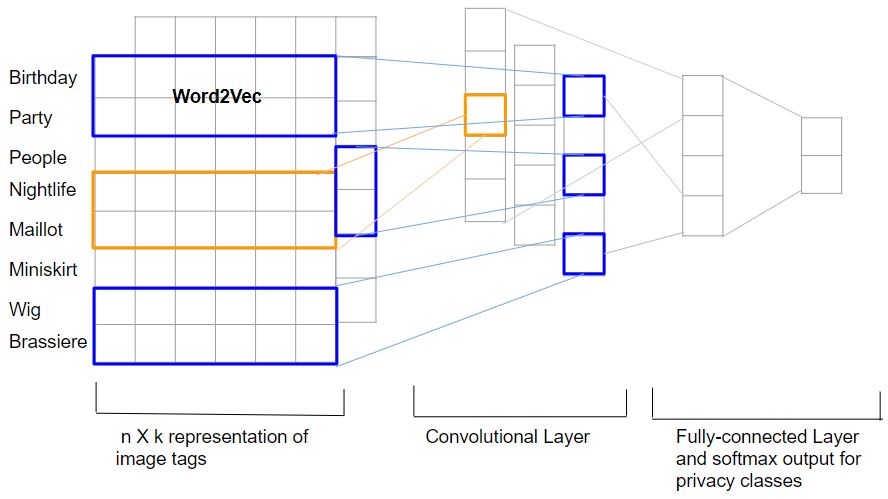}
\caption{\label{fig:cnntags} Tag CNN architecture to classify an image as public or private using image tags.} 
\end{figure*}

\subsection{Image Tags (Bag-of-Tags model)}\label{deeptags}


Prior works on privacy prediction \cite{Squicciarini2014,Zerr:2012,DBLP:conf/aaai/TongeC16,tagtoprotect,DBLP:journals/corr/TongeC15} found that the tags associated with images are indicative of their sensitive content. Tags are also crucial for image-related applications such as indexing, sharing,  searching, content detection and social discovery \cite{citeulike:3503530,DBLP:journals/josis/HollensteinP10,Tang:2009:ISC:1631272.1631305,Gao:2011:TSI:2072298.2072054}.
Since not all images on social networking sites have user tags 
or the set of user tags is very sparse  \cite{mum12}, we use an automatic technique to annotate images with tags based on their visual content as described in our previous work \cite{DBLP:conf/aaai/TongeC16,DBLP:journals/corr/TongeC15}.
Precisely, we predict top $k$ object categories from the probability distribution extracted from a pre-trained CNN. 
These top $k$ categories are images' deep tags, used to describe an image. For example, 
we obtain deep tags such as ``Maillot,'' ``Wig,'' ``Brassiere,'' ``Bra,'' ``Miniskirt'' for the picture in Figure \ref{fig:tagfeat} (note that only top 5 deep tags are shown in the figure). 
Note that the deep tags give some description about the image, but still some relevant tags such as ``people'' and ``women'' are not included since the $1000$ object categories of the ImageNet dataset 
do not contain these tags. Images on social networking sites also give additional information about them through the tags assigned by the user. We call these tags ``User Tags.'' Examples of user tags for the image in Figure \ref{fig:tagfeat} are: ``Birthday Party,'' ``Night Life,'' ``People,'' etc. For user tags, we remove special characters and numbers from the user tags, as they do not provide any information with respect to privacy. 


We combine deep tags and user tags and generate a binary vector representation for the tag set of an image, illustrating presence or absence of tags from tag vocabulary $V$. Particularly, we create a vector of size $|V|$, wherein, for all tags in the tag set, we set $1$ on the position of the tag in the vocabulary ($V$) and
$0$ otherwise. We refer to this model as a Bag-of-Tags (BoT) model and show it's pictorial representation in Figure \ref{fig:tagfeat}.
 
\subsection{Tag CNN}
CNN based models have achieved exceptional results for various NLP tasks such as semantic parsing \cite{yih2014semantic}, search query retrieval, sentence modeling \cite{DBLP:journals/corr/KalchbrennerGB14}, sentence classification \cite{DBLP:conf/emnlp/Kim14}, and other traditional NLP tasks \cite{Collobert:2011:NLP:1953048.2078186}. \citet{DBLP:conf/emnlp/Kim14} developed a CNN architecture for sentence level classification task. A sentence contains keywords in the form of objects, subjects, and verbs that help in the classification task. Image tags are nothing but keywords that are used to describe an image. Thus, 
for privacy prediction, we employ a CNN architecture that has proven adequate for sentence classification \cite{DBLP:conf/emnlp/Kim14}. 

The CNN architecture by \citet{DBLP:conf/emnlp/Kim14} shown in Figure \ref{fig:cnntags} is a slight variant of the CNN architecture of \citet{Collobert:2011:NLP:1953048.2078186}. This architecture contains one layer of convolution on top of word vectors obtained from an unsupervised neural language model. The first layer embeds words (tags in our case) into the word vectors. The word vectors are first initialized with the word vectors that were trained on 100 billion words of Google News, given by \citet{DBLP:conf/icml/LeM14}. Words that are not present in the set of pre-trained words are initialized randomly. These word vectors are then fine-tuned on the tags from the privacy dataset. The next layer performs convolutions on the embedded word vectors using multiple filter sizes of $3$, $4$ and $5$, where we use $128$ filters from each size and produce a tag feature representation. A max-pooling operation \cite{Collobert:2011:NLP:1953048.2078186} over a feature map is applied to take the maximum value of the features to capture the most important feature of each feature map. These features are passed to a fully connected softmax layer to obtain the probability distribution over privacy labels. An illustration of the Tag CNN model can be seen in Figure \ref{fig:cnntags}.



\begin{figure*}[t]
\centering
\subfigure[AlexNet]{\includegraphics[scale=0.21]{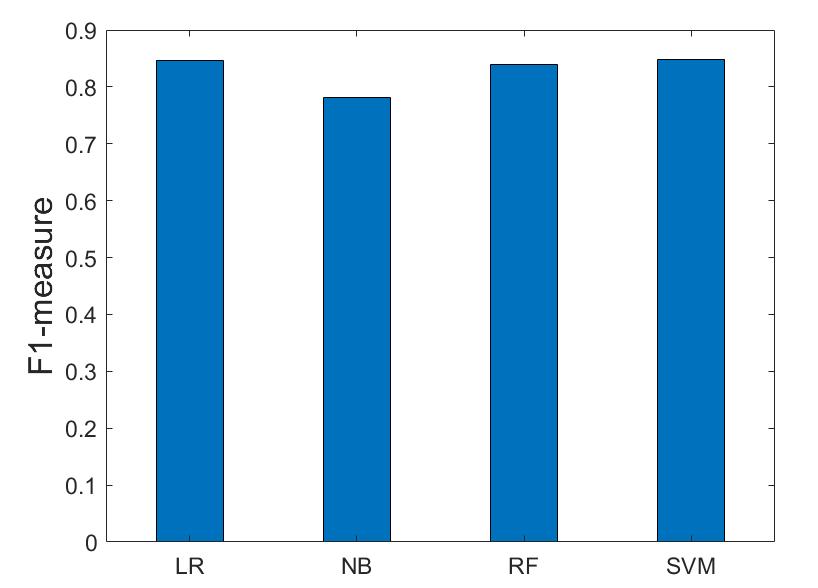}}
\subfigure[GoogLeNet]{\includegraphics[scale=0.21]{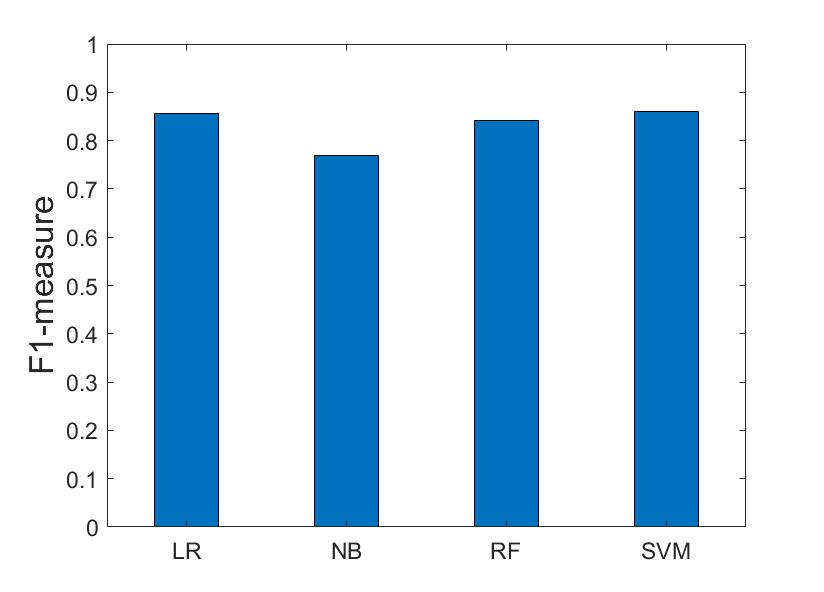}}
\subfigure[VGG]{\includegraphics[scale=0.21]{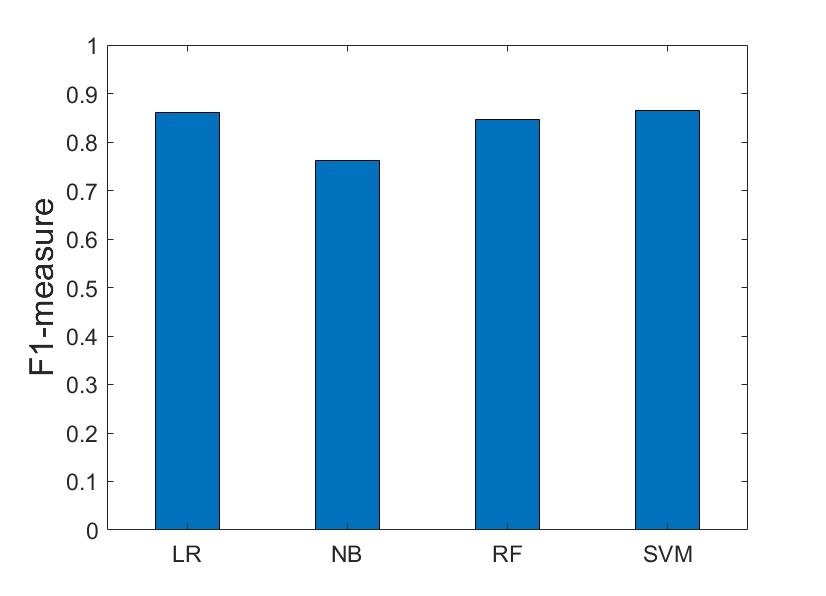}}
\subfigure[ResNet]{\includegraphics[scale=0.21]{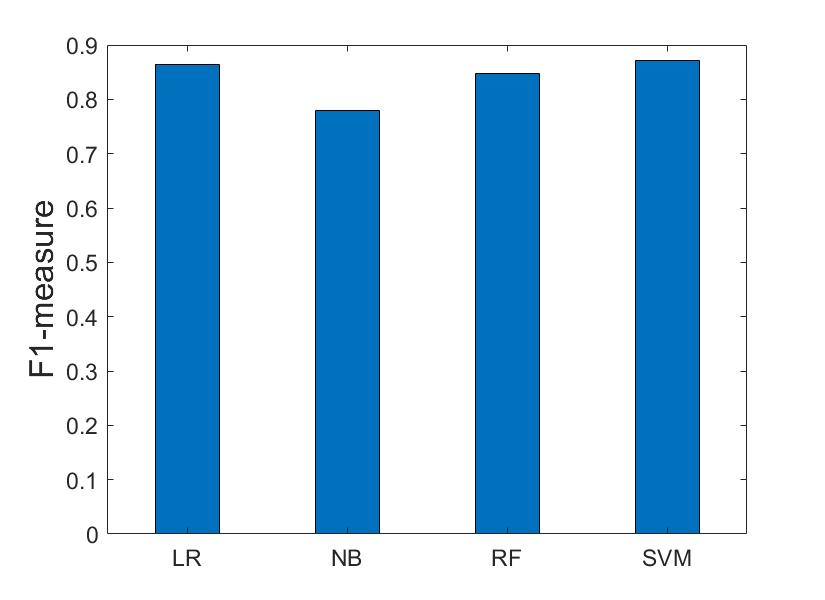}}
\caption{\label{fig:cls} Performance of various classifiers (LR, NB, RF, SVM) using the features derived from all four architectures AlexNet, GoogLeNet, VGG, and ResNet.}
\end{figure*}

\section{Dataset}\label{sec:data}

We evaluated our approach on a subset of $32,000$ Flickr images sampled from the PicAlert dataset, made available by \citet{Zerr:2012,conf/cikm/ZerrSH12}. PicAlert consists of Flickr images on various subjects, which are manually labeled as {\em public} or {\em private} by external viewers. The dataset contains  photos uploaded on Flickr during the period from January to April 2010. The data have been labeled by six teams providing a total of 81  users of ages between 10 and 59 years. One of the teams included graduate computer science students working together at a research center, whereas other teams contained users of social platforms. Users were instructed to consider that their camera has taken these pictures and to mark them as ``private,'' ``public,'' or ``undecidable.'' The guideline to select the label is given as private images belong to the private sphere (like self-portraits, family, friends, someone's home) or contain information that one would not share with everyone else (such as private documents). The remaining images are labeled as public. In case no decision could be made, the image was marked as undecidable. Each image was shown to at least two different users. In the event of disagreement, the photos were presented to additional users. We only consider images that are labeled as public or private.

For all experiments, our $32,000$ images dataset is split into train and test sets of $27,000$ and $5,000$ images, respectively. Each experiment is repeated five times with a different train/test split (obtained using five different random seeds), and the final results are averaged across the five runs. 
The public and private images are in the ratio of 3:1 in both train and test sets.

\section{Experiments, Results and Observations} \label{sec:experiments}

In this section, we perform a broad spectrum of experiments to evaluate features extracted from various deep architectures in order to understand which architecture can capture the complex privacy characteristics and help to distinguish between privacy classes. We first choose the machine learning classifier between generative models, ensemble methods, and discriminative algorithms for privacy prediction. Then, we use the chosen classifier to examine the visual features extracted from all four deep architectures AlexNet, GoogLeNet, VGG-16, and ResNet pre-trained on object data. We further investigate these architectures by fine-tuning them on the privacy data. Next, we compare the performance of models trained on the highest performing
features with that of the state-of-the-art models and baseline approaches for privacy prediction. Additionally, we show the performance of the deep tags obtained through all four pre-trained networks and also study the combination of deep tags and user tags in details for privacy prediction. We show the tag performance in two settings: (1) Bag-of-Tags models and (2) Tag CNN.
We analyze the most promising features derived from both visual and tag encodings for privacy classification. We also provide a detailed analysis of the most informative tags for privacy prediction. Finally, we show the performance of the models trained on the fusion of visual and most informative tag features.



\subsection{Classification Experiments for Features Derived From Pre-Trained CNNs}

We first determine the classifier that works best with the features derived from the pre-trained CNNs. We study the performance of the features using the following classification algorithms: Naive Bayes (NB), Random Forest (RF), Logistic Regression (LR) and Support Vector Machine (SVM). NB is a generative model, whereas RF is an ensemble method using decision trees, and SVM and LR are discriminative algorithms. 
We evaluate the performance of these classifiers using the features derived from the last fully-connected layer of all the architectures, i.e., fc$_8$-A of AlexNet, loss$_3$-G of GoogLeNet, fc$_8$-V of VGG-16, and fc-R of ResNet. Figure \ref{fig:cls} shows the performance of these classifiers in terms of F1-measure for all four architectures. From the figure, we notice that almost all the classifiers perform similarly except NB which performs worse. For example, for Alexnet, with NB we get an F1-measure of $0.781$, whereas SVM obtains an F1-measure of $0.849$. We can also observe that, generally, SVM and LR perform better than RF. For example, for ResNet, using SVM, we get an F1-measure of $0.872$, whereas with RF we get an F1-measure of $0.848$. SVM and LR perform comparably to each other for almost all the architectures except for ResNet. For ResNet, we obtain F1-measure of $0.872$ and $0.865$ using SVM and LR, respectively. The results of SVM over the LR classifier are statistically significant for p-values $<0.05$. Thus, we chose to use SVM with the features derived from pre-trained CNNs for all of our next experiments.

To evaluate the proposed features, we used the SVM Weka implementation and chose the hyper-parameters that gave the best performance using 10-fold cross-validation on the training set. We experimented with $C=\{0.001, 0.01, 1.0, \cdots, 10.0\}$, kernels: Polynomial and RBF, the $\gamma$ parameter in RBF, and the degree $d$ of a polynomial. Hyper-parameters shown in all subsequent tables follow the format: ``R/P,C,$\gamma$/$d$'' where ``R'' denotes ``RBF'' and ``P'' denotes ``Polynomial'' kernel.

\begin{table*}[t]
\renewcommand{\arraystretch}{1.35}
\centering
\begin{small}
\begin{tabular}{|l|l|c|c|c|c|c|c|c|c|c|c|}
\hline
&&& \multicolumn{3}{|c|}{Overall} & \multicolumn{3}{|c|}{Private} & \multicolumn{3}{|c|}{Public}\\
\hline
{Features} & {H-Param} & {Acc \%} & {F1} & {Prec} & {Re} & {F1} & {Prec} & {Re} & {F1} & {Prec} & {Re}  \\
\hline
\hline
\multicolumn{12}{|l|}{AlexNet}\\
\hline
\hline
fc$_6$-A & R,$1.0$,$0.05$ & 82.29 &	0.82 & 0.819 & 0.823 & 0.613 & 0.639 & 0.591 & 0.885 & 0.875 & 0.895\\
fc$_7$-A & R,$2.0$,$0.01$ & 82.97 & 0.827 & 0.825 & 0.83 & 0.627 & 0.656 & 0.602 & 0.889 & 0.878 & 0.901 \\
fc$_8$-A & R,$1.0$,$0.05$ & {\em  \color{burntorange} 85.51} & {\em  \color{burntorange} 0.849} & {\em  \color{burntorange} 0.849} & {\em  \color{burntorange} 0.855} & {\em  \color{burntorange} 0.661} & {\em  \color{burntorange} 0.746} & {\em  \color{burntorange} 0.595} & {\em  \color{burntorange} 0.908} & {\em  \color{burntorange} 0.881} & {\em  \color{burntorange} 0.936} \\
prob-A & R,$5.0$,$1.0$ & 82.76 & 0.815 & 0.816 & 0.828 & 0.568 & 0.704 & 0.477 & 0.892 & 0.851 & 0.937\\
\hline
\hline
\multicolumn{12}{|l|}{GoogLeNet}\\
\hline
\hline
loss$_3$-G & P,$0.001$,$2.0$ & {\em \color{burntorange}  86.42} & {\em \color{burntorange}  0.861} & {\em \color{burntorange} 0.86} & {\em \color{burntorange}  0.864} & {\em \color{burntorange}  0.695} & {\em \color{burntorange}  0.746} & {\em \color{burntorange}  0.652} & {\em \color{burntorange}  0.913} & {\em \color{burntorange}  0.895} & {\em \color{burntorange}  0.93}\\
prob-G & R,$50.0$,$0.05$ & 82.66 & 0.815 & 0.816 & 0.827 & 0.573 & 0.694 & 0.488 & 0.891 & 0.853 & 0.933\\
\hline
\hline
\multicolumn{12}{|l|}{VGG-16}\\
\hline
\hline
fc$_6$-V & R,$1.0$,$0.01$ & 83.85 & 0.837 & 0.836 & 0.839 & 0.652 & 0.67 & 0.636 & 0.895 & 0.888 & 0.902\\
fc$_7$-V & R,$2.0$,$0.01$ & 84.43 & 0.843 & 0.842 & 0.844 & 0.663 & 0.684 & 0.644 & 0.899 & 0.891 & 0.907\\
fc$_8$-V & R,$2.0$,$0.05$ & {\em  \color{burntorange} 86.72} & {\em  \color{burntorange} 0.864} & {\em  \color{burntorange} 0.863} & {\em  \color{burntorange} 0.867} & {\em  \color{burntorange} 0.7} & {\em  \color{burntorange} 0.758} & {\em  \color{burntorange} 0.65} & {\em  \color{burntorange} 0.915} & {\em  \color{burntorange} 0.895} & {\em  \color{burntorange} 0.935}\\

prob-V & R,$2.0$,$0.05$ & 81.72 & 0.801 & 0.804 & 0.817 & 0.528 & 0.687 & 0.429 & 0.887 & 0.84 & 0.939\\
\hline
\hline
\multicolumn{12}{|l|}{ResNet}\\
\hline
\hline
fc-R & R,$1.0$,$0.05$ & {\color{blue}{\bf 87.58}} & {\color{blue}{\bf 0.872}} & {\color{blue}{\bf 0.872}} & {\color{blue}{\bf 0.876}} & {\color{blue}{\bf 0.717}} & {\color{blue}{\bf 0.783}} & {\color{blue}{\bf 0.662}} & {\color{blue}{\bf 0.92}} & {\color{blue}{\bf 0.899}} & {\color{blue}{\bf 0.943}}\\
prob-R & R,$2.0$,$0.05$ & 80.6 & 0.784 & 0.789 & 0.806 & 0.473 & 0.67 & 0.366 & 0.881 & 0.826 & 0.943 \\
\hline
\end{tabular}
\caption{\label{table:table1pretr} Comparison of SVMs trained on features extracted from pre-trained architectures AlexNet, GoogLeNet, VGG-16 and ResNet. The best performance is shown in bold and blue color. The best performance for each network is shown in italics and orange color.} 
\end{small}

\end{table*}

\subsection{The Impact of the CNN Architecture on the Privacy Prediction}

In this experiment, we aim to determine which architecture performs best for privacy prediction by investigating the performance of privacy prediction models based on visual semantic features extracted from all four architectures, AlexNet, GoogLeNet, VGG-16, and ResNet pre-trained on object data of ImageNet. 
We extract deep visual features: (1) fc$_6$-A, fc$_7$-A, fc$_8$-A and ``prob-A'' from AlexNet, (2) loss$_3$-G and ``prob-G'' from GoogLeNet, (3) fc$_6$-V, fc$_7$-V, fc$_8$-V and ``prob-V'' from VGG-16, and (4) fc-R and ``prob-R'' from ResNet.
For AlexNet and GoogLeNet, we used the pre-trained networks that come with the CAFFE open-source framework for CNNs \cite{Jia:2014:CCA:2647868.2654889}. For VGG-16, we used an improved version of pre-trained models presented by the VGG-16 team in the ILSVRC-2014 competition \cite{DBLP:journals/corr/SimonyanZ14a}. For ResNet, we use the ResNet pre-trained models of 101 layers given by \citet{DBLP:conf/cvpr/HeZRS16}.

\begin{figure*}[t]
\centering
\begin{minipage}{.44\textwidth}
  \centering
\includegraphics[scale=0.3]{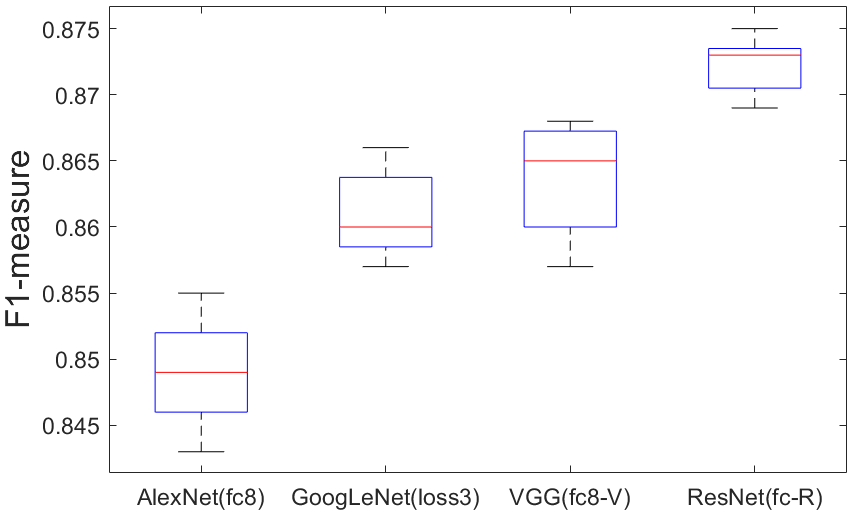}
\caption{\label{fig:pre-trainedboxplot} Box plot of F1-measure (overall) obtained for the best-performing features derived from each CNN over five splits.} 
\end{minipage}%
\hspace{10mm}
\begin{minipage}{.48\textwidth}
  \centering
\includegraphics[scale=0.45]{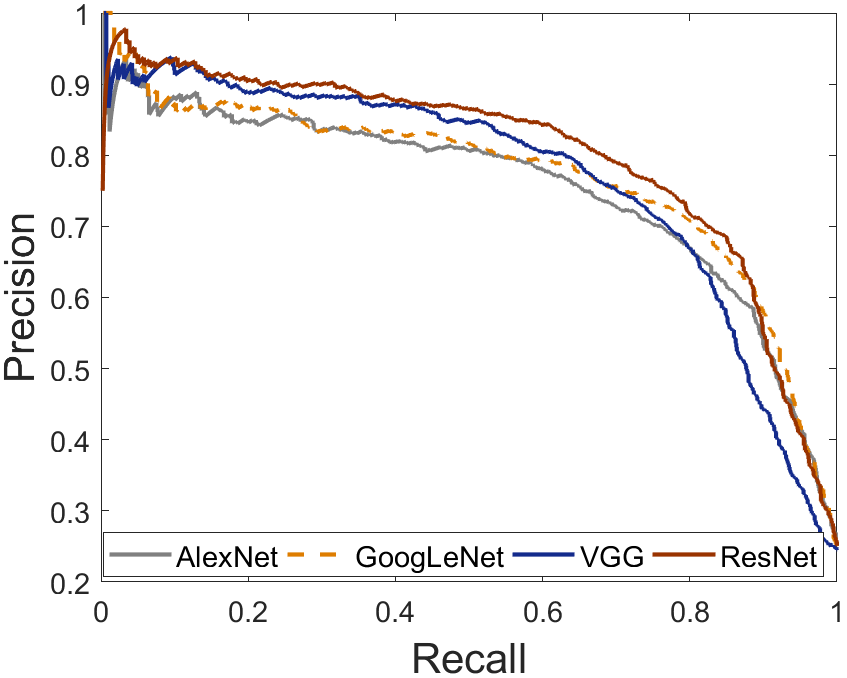}
\caption{\label{fig:pre-trainedprcurves} Precision-recall curves for the private class obtained using features extracted from all four architectures AlexNet (fc$_8$), GoogLeNet (loss$_3$), VGG-16 (fc$_8$-V) and ResNet (fc-R).} 
\end{minipage}
\end{figure*}

Table \ref{table:table1pretr} shows the performance (Accuracy, F1-measure, Precision, Recall) of SVMs trained on the features extracted from all four pre-trained networks. 
From the table, we can observe that the models trained on the features extracted from ResNet consistently yield the best performance. For example, ResNet achieves an F1-measure of $0.872$ as compared with $0.849$, $0.861$, $0.864$ achieved by AlexNet, GoogLeNet, and VGG-16, respectively.
These results suggest that 
the deep Residual Networks have more representational abilities compared to the other networks, and are more effective for predicting appropriate privacy classes of images. Additionally, ResNets are substantially deeper than their ``plain'' counterparts, which allows extracting various image-specific features that are beneficial for learning images' privacy characteristics better. 
Since privacy involves understanding the complicated relationship between the objects present in images, the features derived from ResNet prove to be more adequate than the features obtained by simply stacking convolutional layers. 
In Table \ref{table:table1pretr}, we also show the class-specific privacy prediction performance in order to identify which features characterize the private class effectively as sharing private images on the Web with everyone is not desirable. 
Interestingly, we found that the model trained on features obtained from ResNet provides improved F1-measure, precision, and recall for the private class. Precisely, F1-measure for the private class improves from $0.661$ (for AlexNet) to $0.717$ (for ResNet), yielding an improvement of $6\%$. Similarly, for precision and recall, we obtain an increase of $4\%$ and $7\%$, respectively, using ResNet features over the AlexNet features. 

From Table \ref{table:table1pretr}, we also notice that the overall best performance (shown in orange and blue color) obtained for each network is higher than $\approx 85\%$ in terms of all compared measures (overall - Accuracy, F1-measure, precision and recall). 
Note that a naive baseline which classifies every image as ``public'' obtains an accuracy of $75\%$. Additionally, analyzing the results obtained by the VGG-16 features, we notice that as we ascend the fully-connected layers of the VGG-16 network from fc$_6$-V to fc$_8$-V, the F1-measure improves from $0.837$ to $0.864$ (see Table \ref{table:table1pretr}). Similarly, for AlexNet, the F1-measure improves from $0.82$ (for fc$_6$-A) to $0.849$ (for fc$_8$-A). This shows that the high-level object interpretations obtained through the last fully-connected layer helped to derive better privacy characteristics. Moreover, it is worth noting that ``prob'' features perform worse than the features extracted from the fully-connected layers (on all architectures). For example, prob-G obtains an F1-measure of $0.815$, whereas loss$_3$-G achieves an F1-measure of $0.861$. One possible explanation could be that squashing the values at the previous layer (e.g., loss$_3$-G in GoogleNet) through the softmax function, which yields the ``prob'' layer, produces a non-linearity that is less useful for SVM compared to the untransformed values.  We also experimented with a combination of features, e.g., fc$_7$-A concatenated with fc$_8$-A, but we did not obtain a significant improvement over the individual (fc$_7$-A and fc$_8$-A) features.

We also analyze the performance by showing the box plots of F1-measure in Figure \ref{fig:pre-trainedboxplot}, obtained for the most promising features of all the architectures over the five random splits of the dataset. The figure indicates that the model trained on ResNet features is statistically significantly better than the models that are trained on the features derived from the other architectures. We further compare features derived through all the architectures using precision-recall curves given in Figure \ref{fig:pre-trainedprcurves}. The curves show again that features derived from ResNet perform better than the features obtained from the other architectures, for a recall ranging from $0.5$ to $0.8$. For example, for a recall of $0.7$, we achieve a precision of $0.75$, $0.8$, $0.8$ and $0.85$ for AlexNet, GoogLeNet, VGG-16, and ResNet, respectively. 

\begin{table*}[]
\renewcommand{\arraystretch}{1.3}
\centering
\begin{small}
\begin{adjustbox}{max width=15.7cm}
\begin{tabular}{|p{1.3cm}|l|c|c|c|c|c|c|c|c|c|c|}
\hline
&&& \multicolumn{3}{|c|}{Overall} & \multicolumn{3}{|c|}{Private} & \multicolumn{3}{|c|}{Public}\\
\hline
{Features} & {H-Param} & {Acc \%} & {F1} & {Prec} & {Re} & {F1} & {Prec} & {Re} & {F1} & {Prec} & {Re}  \\
\hline
\hline
\multicolumn{12}{|l|}{Fine-tuned AlexNet}\\
ft-A & fc & 85.01 & 0.846 & 0.845 & 0.851 & 0.657 & 0.723 & 0.606 & 0.904 & 0.883 & 0.926\\
ft-A & fc-all & 85.14 &  0.849 & 0.847 & 0.852 & 0.669 & 0.713 & 0.632 & 0.904 & 0.889 & 0.92\\
ft-A & all & 85.07 & 0.848 & 0.847 & 0.851 & 0.67 & 0.707 & {\em  \color{burntorange} 0.638} & 0.904 & 0.89 & 0.917\\
\multicolumn{12}{|l|}{Pre-trained AlexNet}\\
fc$_8$-A & R,$1$,$0.05$ & 85.51 & 0.849 & 0.849 & 0.855 &  0.661 & {\em  \color{burntorange} 0.746} & 0.595 & 0.908 & 0.881 & 0.936 \\
\hline 
\hline
\multicolumn{12}{|l|}{Fine-tuned GoogLeNet}\\
ft-G & fc & 86.27 & 0.86 & 0.859 & 0.863 & 0.694 & 0.74 & 0.653 & 0.911 & 0.895 & 0.928\\
ft-G & all & 86.77 & 0.867 & 0.867 & 0.868 & {\em  \color{burntorange} 0.717} & 0.732 & {\em  \color{burntorange} 0.705} & 0.914 & 0.909 & 0.919\\
\multicolumn{12}{|l|}{Pre-trained GoogLeNet}\\
loss$_3$-G & P,$0.001$,$2$ & 86.42 & 0.861 & 0.86 & 0.864 &  0.695 & 0.746 & 0.652 & 0.913 & 0.895 & 0.930\\
\hline
\hline
\multicolumn{12}{|l|}{Fine-tuned VGG-16}\\
ft-V & fc & 86.74 & 0.864 & 0.865 & 0.869 & 0.695 & {\em  \color{burntorange} 0.782} & 0.631 & 0.916 & 0.891 & {\em  \color{burntorange} 0.944}\\
ft-V & fc-all & 86.92 & 0.869 & 0.87 & 0.869 & {\color{blue}{\bf 0.722}} & 0.73 & {\color{blue}{\bf 0.717}} & 0.914 & {\color{blue}{\bf 0.912}} & 0.917\\
ft-V & all & 86.76 & 0.867 & 0.867 & 0.868 & 0.718 & 0.729 & 0.709 & 0.913 & 0.91 & 0.917\\
\multicolumn{12}{|l|}{Pre-trained VGG-16}\\
fc$_8$-V & R,$2$,$0.05$ & 86.72 & 0.864 & 0.863 & 0.867 & 0.700 & 0.758 & 0.65 & 0.915 & 0.895 & 0.935\\
\hline
\hline
\multicolumn{12}{|l|}{Fine-tuned ResNet}\\
ft-R & fc & 87.23 & 0.87 & 0.869 & 0.873 & 0.717 & 0.759 & {\em  \color{burntorange} 0.68} & 0.918 & 0.903 & 0.932\\
ft-R & all & 86.19 & 0.856 & 0.856 & 0.863 & 0.672 & 0.776 & 0.594 & 0.913 & 0.881 & {\color{blue}{\bf 0.946}}\\
\multicolumn{12}{|l|}{Pre-trained ResNet}\\
fc-R & R,$1$,$0.05$ & {\color{blue}{\bf 87.58}} & {\color{blue}{\bf 0.872}} & {\color{blue}{\bf 0.872}} & {\color{blue}{\bf 0.876}} & 0.717 & {\color{blue}{\bf 0.783}} & 0.662 & {\color{blue}{\bf 0.92}} &  0.899 & 0.943\\
\hline
\end{tabular}
\end{adjustbox}
\end{small}
\caption{Fine-tuned networks vs. Pre-trained networks.  The best performance is shown in bold and blue color. The performance measures that achieve a better performance after fine-tuning a CNN over pre-trained features are shown in italics and orange color.}
\label{table:finetune}
\end{table*}

\subsection{Fine-Tuned Networks vs. Pre-Trained Networks}
%
%

Previous works showed that the features transferred from the network pre-trained on the object dataset to the privacy data achieved a good performance \cite{Tran:2016:PFD:3015812.3016006}. Moreover, many other works used ``transfer learning'' to get more dataset specific features \cite{donahue13, journals/jmlr/Bengio12}. Thus, we determine the performance of fine-tuned networks on the privacy dataset. We compare fine-tuned networks of all four architectures with the deep features obtained from pre-trained networks. We refer the fine-tuned networks of AlexNet, GoogLeNet, VGG-16, and ResNet as ``ft-A,'' ``ft-G,'' ``ft-V,'' and ``ft-R'' respectively. For fine-tuning, we used the same CNN architectures pre-trained on the object dataset, and employed in previous experiments. To fine-tune the networks, we experiment with the three types of settings: (1) fine-tune the last fully-connected layer (that has two output units corresponding to $2$ privacy classes) with higher learning rates as compared to the learning rates of the rest of the layers of the networks ($0.001$ vs. $0.0001$), referred as ``fc.'' (2) fine-tune all the fully-connected layers of the networks with higher learning rates and convolutional layers are learned with smaller learning rates. We refer to this setting as ``fc-all.'' (3) fine-tune all layers with the same learning rates and denoted as ``all.'' Note that since ResNet and GoogLeNet have only one fully-connected layer, we report the performance obtained only using ``fc,'' and ``all'' settings. The very low learning rate avoids substantial learning of the pre-trained layers. In other words, due to a very low learning rate ($0.0001$), pre-trained layers learn very slowly as compared to the layers that have a higher learning rate ($0.001$) to learn the required weights for privacy data.

\begin{table*}[t]
\renewcommand{\arraystretch}{1.4}
\centering
\begin{small}
\begin{tabular}{|p{1.0cm}|l|c|c|c|c|c|c|c|c|c|c|}
\hline
&&& \multicolumn{3}{|c|}{Overall} & \multicolumn{3}{|c|}{Private} & \multicolumn{3}{|c|}{Public}\\
\hline
{Features} & {H-Param} & {Acc \%} & {F1} & {Prec} & {Re} & {F1} & {Prec} & {Re} & {F1} & {Prec} & {Re}  \\
\hline
\hline
\multicolumn{12}{|l|}{Highest performing CNN architecture}\\
\hline
\hline
fc-R & R,$1.0$,$0.05$ & {\color{blue}{\bf 87.58}} & {\color{blue}{\bf 0.872}} & {\color{blue}{\bf 0.872}} & {\color{blue}{\bf 0.876}} & {\color{blue}{\bf 0.717}} & {\color{blue}{\bf 0.783}} & {\color{blue}{\bf 0.662}} & {\color{blue}{\bf 0.92}} &  0.899 & {\color{blue}{\bf 0.943}}\\
\hline
\hline
\multicolumn{12}{|l|}{\#1 PCNH framework 
}\\
\hline
\hline
PCNH & $-$ & 83.13 & 0.824 & 0.823 & 0.831 & 0.624 & 0.704 & 0.561 & 0.891 & 0.863 & 0.921\\
\hline
\hline
\multicolumn{12}{|l|}{\#2 AlexNet Deep Features 
}\\
\hline
\hline
fc$_8$-A & R,$1.0$,$0.05$ & 85.51 & 0.849 & 0.849 & 0.855 &  0.661 & 0.746 & 0.595 & 0.908 & 0.881 & 0.936 \\
\hline
\hline
\multicolumn{12}{|l|}{\#3 SIFT \& GIST models 
}\\
\hline
\hline
SIFT & P,$1.0$,$2.0$ & 77.31 & 0.674	 & 0.598 & 0.773 & 0.002 & 0.058 & 0.001 & 0.87 & 0.772 & 0.995\\
GIST & R,$0.001$,$0.5$ & 77.33 & 0.674	 & 0.598 & 0.773 & 0.002 & 0.058 & 0.001 & 0.87 & 0.772 & 0.995\\
SIFT \& GIST & R,$0.05$,$0.5$ & 72.67 & 0.704	 & 0.691 & 0.727 & 0.27 & 0.343 & 0.223 & 0.832 & 0.793 & 0.874\\
\hline
\hline
\multicolumn{12}{|l|}{\#4 Rule-based models}\\
\hline
\hline
Rule-1 & $-$ & 77.35 & 0.683 & 0.694 & 0.672 & 0.509 & 0.47 & 0.556 & 0.853 & 0.875 & 0.832\\
Rule-2 & $-$ & 77.93 & 0.673 & 0.704 & 0.644 & 0.458 & 0.373 & 0.593 & 0.897 & {\color{blue}{\bf 0.914}} & 0.88\\
\hline
\end{tabular}
\end{small}
\caption{Highest performing visual features (fc-R) vs. Prior works.} 
\label{table:compvisualf}
\end{table*}

Table \ref{table:finetune} shows the performance comparison of the models obtained by fine-tuning architectures on privacy data and the models trained on the features derived from the pre-trained networks.  We notice that we get mostly similar results when we fine-tune pre-trained models on our privacy dataset as compared to the models trained on the features derived from the pre-trained architectures. However, we get improved recall for the private class when we fine-tune the networks on the privacy dataset. For example, the fine-tuned VGG-16 network gets an improvement of $6.7\%$ in the recall for the private class (see ft-V, fc-all setting vs. fc$_8$-V) over the models trained on the features extracted from the pre-trained VGG-16. The performance measures that achieve a better performance after fine-tuning a CNN over pre-trained features are shown in italics and orange color for each network.  We notice that the fine-tuned VGG gives the best performance for the F1-measure and recall of the private class (shown in bold and blue color). However, the models trained on the features derived from the pre-trained ResNet yield the best overall performance (shown in bold and blue color). Thus, we compare the models trained on fc-R features with prior privacy prediction approaches in the next subsection.

\subsection{ResNet Features-Based Models vs. Prior Works}


We compare the performance of the state-of-the-art works on privacy prediction, as detailed below, with the models trained using ResNet features, i.e., fc-R.

{\bf 1. PCNH privacy framework} \cite{Tran:2016:PFD:3015812.3016006}:
This framework combines features obtained from two architectures: one that extracts convolutional features (size = $24$, referred as Convolutional CNN), and another that extracts object features (size = $24$, referred as Object CNN). The Convolutional CNN contains two convolutional layers and three fully-connected layers of size $512$, $512$, $24$, respectively. On the other hand, the object CNN is an extension of AlexNet architecture that appends three fully-connected layers of size $512$, $512$, and $24$, at the end of the last fully-connected layer of AlexNet and form a deep network of $11$ layers. The two CNNs are connected at the output layer. The PCNH framework is first trained on the ImageNet dataset and then fine-tuned on a small privacy dataset.

{\bf 2. AlexNet features} \cite{DBLP:conf/aaai/TongeC16,tongemsm18,DBLP:journals/corr/TongeC15}: We consider the model trained on the features extracted from the last fully-connected layer of AlexNet, i.e., fc$_8$-A as another baseline, since in our previous works we achieved a good performance using these features for privacy prediction.  

{\bf 3. SIFT \& GIST} \cite{Zerr:2012,Squicciarini2014,Squicciarini:2017:TAO:3062397.2983644}:
We also consider classifiers trained on the best performing features between SIFT, GIST, and their combination as our baselines. Our choice of these features is motivated by their good performance over other visual features such as colors, patterns, and edge directions in prior works \cite{Squicciarini2014,Zerr:2012}. For SIFT, we construct a vocabulary of 128 visual words for our experiments. We tried different numbers of visual words such as 500, 1000, etc., but we did not get a significant improvement over the 128 visual words. 
For a given image, GIST is computed by first convolving the image with 32 Gabor filters at 4 scale and 8 orientations, which produces 32 feature maps; second, dividing the feature map into a $4\times4$ grid and averaging feature values of each cell; and third, concatenating these 16 averaged values for 32 feature maps, which results in a feature vector of 512 ($16 \times 32$) length. 

{\bf 3. Rule-based classifiers}: We also compare the performance of models trained on ResNet features fc-R with two rule-based classifiers which predict an image as {\em private} if it contains persons. Otherwise, the image is classified as {\em public}. For the first rule-based classifier, we detect front and profile faces by using Viola-Jones algorithm \cite{Viola01robustreal-time}. For the second rule-based classifier, we consider user tags such as ``women,'' ``men,'' ``people.'' 
Recall that these tags are not present in the set of $1,000$ categories of the ILSVRC-2012 subset of the ImageNet dataset, and hence, we restrict to user tags only. 
If an image contains one of these tags or detects a face, we consider it as ``private,'' otherwise ``public.''

\begin{figure*}[htb]
\centering
\subfigure[Precision-Recall curves]{\includegraphics[scale=0.3]{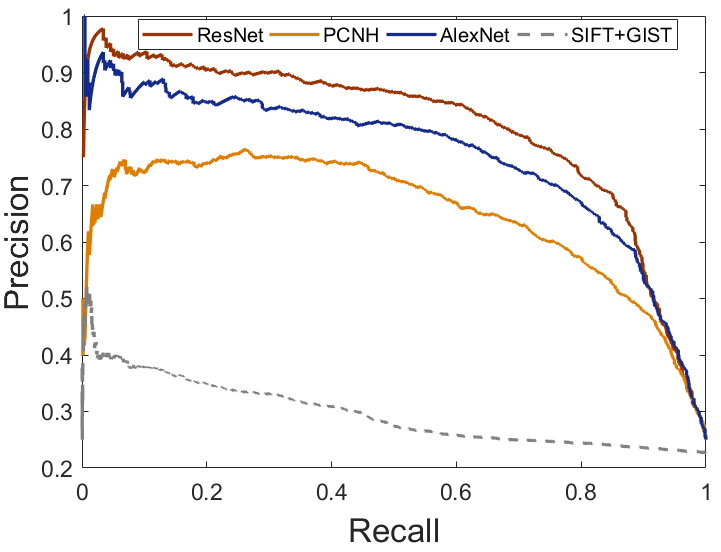}}
\hspace{10mm}
\subfigure[Threshold curves]{\includegraphics[scale=0.3]{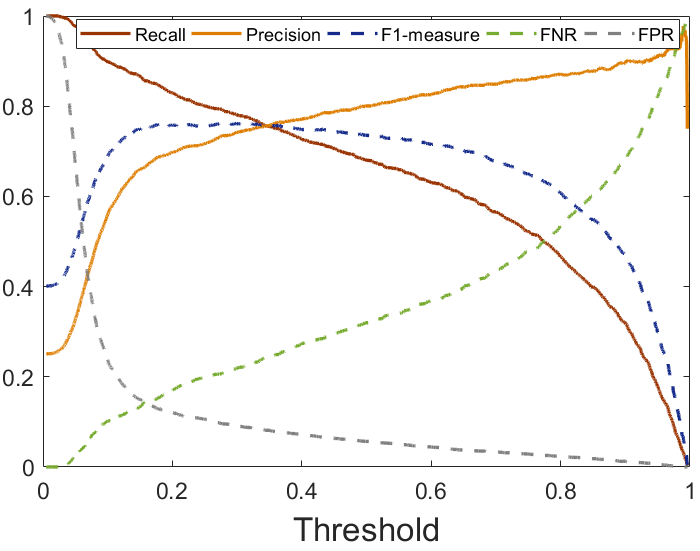}}
\caption{\label{fig:prcurve} Precision-Recall and Threshold curves for the private class obtained using ResNet features (fc-R) and prior works.}
\end{figure*}

Table \ref{table:compvisualf} compares the performance of models trained on fc-R features (the highest performing features obtained from our previous experiments) with the performance obtained by prior works. 
As can be seen from the table, the deep features extracted from the pre-trained ResNet achieve the highest performance, and hence, are able to learn the privacy characteristics better than the prior works with respect to both the classes. 
Precisely, using fc-R features, F1-measure improves from $0.824$ obtained by PCNH framework 
to $0.872$ obtained by fc-R features, providing an overall improvement of $5\%$. Moreover, for the private class, fc-R features yield an improvement of $9.8\%$ in F1-measure over the more sophisticated PCNH framework (from $0.624$, PCNH to $0.717$, fc-R features). 


One possible explanation could be that the object CNN of PCNH framework is formed by appending more fully-connected layers to the AlexNet architecture and the increase in the number of complex non-linear layers (fully-connected layers) introduces more parameters to learn. At the same time, with a relatively small amount of training data (PicAlert vs. ImageNet), the object CNN model can over-fit. 
On the other hand, 
as images' privacy greatly depends on the objects in images, we believe that the low-level  features controlling the distinct attributes of the objects (e.g., edges of swim-suit vs. short pants) obtained through the convolutional layers can better approximate the privacy function compared with adding more non-linear layers (as in PCNH). This is justified by the results, where the network with more convolutional layers, i.e., ResNet achieves a better performance as compared to the network with more fully-connected layers, i.e., PCNH. Additionally, even though PCNH attempted to capture convolutional features using Convolutional CNN, both CNN (convolutional and object) vary in their discriminative power and thus obtaining an optimal unification of convolutional CNN and object CNN is difficult. 
Moreover, PCNH is required to first train on ImageNet and then fine-tune on the PicAlert dataset. Training a deep network such as PCNH two times significantly increases the processing power and time. 
On the other hand, through our experiments, we found that the features derived from the state-of-the-art ResNet model can reduce the overhead of re-training and achieve a better performance for privacy prediction.

As discussed before, the models trained on ResNet features outperform those trained on AlexNet features. Interestingly, the best performing baseline among all corresponds to the SVM trained on the deep features extracted from the AlexNet architecture. For example, the SVM trained on the AlexNet features (fc$_8$-A) yields an F1-measure of $0.849$ as compared with the F1-measure of $0.824$ achieved by the PCNH framework. We hypothesize that this is due to the model complexity and the small size of the privacy dataset used to train the PCNH framework. For example, merging two CNNs (as in PCNH) that vary in depth, width, and optimization algorithm can become very complex and thus the framework potentially has more local minima, that may not yield the best possible results. Additionally, unlike \citet{Tran:2016:PFD:3015812.3016006}, that used $800$ images in their evaluation, we evaluate the models on a large set of images (32000), containing a large variety of image subjects. The features derived from the various layers of the  state-of-the-art AlexNet reduce the overhead of training the complex structure and still achieve a good performance for privacy prediction. 

Another interesting aspect to note is that, although we showed earlier that the fine-tuned network (in this case VGG-$16$) does not show a significant improvement over the ResNet pre-trained features (see Table \ref{table:finetune}), our fine-tuning approach yields better results compared to the PCNH framework. For example, fine-tuned VGG-$16$ (ft-V) achieves an F1-measure of $0.869$ whereas PCNH achieves an F1-measure of $0.824$ (see Tables \ref{table:finetune} and \ref{table:compvisualf}). The possible reasons could be that we use a larger privacy dataset to fine-tune a simpler architecture, unlike PCNH that merges two convolutional neural networks. Additionally, we fine-tune the state-of-the-art VGG-16 model presented by \citet{DBLP:journals/corr/SimonyanZ14a}, contrary to PCNH that required estimating optimal network parameters to train the merged architecture on the ImageNet dataset.

As expected, we can see from Table \ref{table:compvisualf} that the baseline models trained on SIFT/GIST and the rule-based models are the lowest performing models. For example, the fc-R based models achieve improvement in performance as high as $17\%$ over SIFT/GIST models. 
With a paired T-test, the improvements over the prior approaches for F1-measure are statistically significant for p-values $< 0.05$. It is also interesting to note that rules based on facial features exhibit better performance than SIFT and GIST and suggest that feature representing persons are helpful to predict private images. 
However, fc-R features outperform the rule-based models based on facial features by more than 10\% in terms of all measures. 

\begin{table*}[t]
\renewcommand{\arraystretch}{1.1}
\centering
\begin{small}
\begin{tabular}{|p{1.2cm}|l|c|c|c|c|c|c|c|c|c|c|}
\hline
&&& \multicolumn{3}{|c|}{Overall} & \multicolumn{3}{|c|}{Private} & \multicolumn{3}{|c|}{Public}\\
\hline
{Features} & {H-Param} & {Acc \%} & {F1} & {Prec} & {Re} & {F1} & {Prec} & {Re} & {F1} & {Prec} & {Re}  \\
\hline
\hline
\multicolumn{12}{|l|}{Best performing CNN architecture}\\
\hline
\hline
fc-R & R,$1.0$,$0.05$ & {\color{blue}{\bf 87.58}} & {\color{blue}{\bf 0.872}} & {\color{blue}{\bf 0.872}} & {\color{blue}{\bf 0.876}} & {\color{blue}{\bf 0.717}} & {\color{blue}{\bf 0.783}} & 0.662 & {\color{blue}{\bf 0.92}} & 0.899 & {\color{blue}{\bf 0.943}}\\
\hline
\hline
\multicolumn{12}{|l|}{\#1 User Tags (BoT)}\\
\hline
\hline
UT & R,$2.0$,$0.05$ & 78.63 & 0.777 & 0.772 & 0.786 & 0.496 & 0.565 & 0.442 & 0.865 & 0.837 & 0.894\\

\hline
\hline
\multicolumn{12}{|l|}{\#2 Deep Tags (BoT)}\\
\hline
\hline

DT-A & R,$1.0$,$0.1$ & 83.34 & 0.825 & 0.824 & 0.833 & 0.601 & 0.699 & 0.529 & 0.895 & 0.863 & 0.929 \\

DT-G & R,$1.0$,$0.05$ & 83.59 & 0.828 & 0.827 & 0.836 & 0.606 & 0.699 & 0.534 & 0.896 & 0.866 & 0.929 \\

DT-V & P,$1.0$,$1.0$ & 83.42 & 0.826 & 0.825 & 0.834 & 0.607 & 0.698 & 0.537 & 0.895 & 0.865 & 0.927 \\

DT-R & P,$1.0$,$1.0$ & 83.78 & 0.833 & 0.831 & 0.838 & 0.631 & 0.688 & 0.584 & 0.896 & 0.876 & 0.917 \\

\hline
\hline
\multicolumn{12}{|l|}{\#3 User Tags \& Deep Tags }\\
\hline
\hline

UT+DT-R (BoT) & R,$1.0$,$0.05$ & 84.33 & 0.84 & 0.839 & 0.843 & 0.67 & 0.709 & 0.636 & 0.897 & 0.882 & 0.913 \\

Tag CNN & $-$ & 85.13 & 0.855 & 0.855 & 0.854 & 0.706 & 0.700 & {\color{blue}{\bf 0.712}} & 0.901 & {\color{blue}{\bf 0.903}} & 0.898 \\

\hline
\end{tabular}
\end{small}
\caption{Visual features vs. Tag features.} 
\label{table:compvisualftagf}
\end{table*}

We further analyze fc-R features and compare their performance with the prior works through precision-recall curves shown in Figure \ref{fig:prcurve} (a). As can be seen from the figure, the SVM trained on ResNet features achieve a precision of $\approx 0.8$ for recall values up to $0.8$, and after that, the precision drops steadily. 

The performance measures shown in previous experiments are calculated using a classification threshold of $0.5$. In order to see how the performance measures vary for different classification thresholds, we plot the threshold curve and show this in Figure \ref{fig:prcurve} (b).  From the figure, we can see that the precision increases from $\approx 0.68$ to $\approx 0.97$ at a slower rate with the classification threshold. The recall slowly decreases to $0.8$ for a threshold value of $\approx 0.4$, and the F1-measure remains comparatively constant until $\approx 0.75$. At a threshold of $\approx 0.4$, we get equal precision and recall of $\approx 0.78$, which corresponds to the breakeven point. In the figure, we also show the false negative rate and false positive rate, so that 
depending on a user's need (high precision or high recall), the classifier can run at the desired threshold. Also, to reduce the number of content-sensitive images shared with everyone on the Web, lower false negative (FN) rates are desired. From Figure \ref{fig:prcurve} (b), we can see that we achieve lower FN rates up to $\approx 0.4$ for the threshold values up to $0.8$.

\subsection{Best Performing Visual Features vs. Tag Features}\label{sec:visualvstag}


Image tags provide relevant cues for privacy-aware image retrieval  \cite{Zerr:2012} and can become an essential tool for surfacing the hidden content of the deep Web without exposing sensitive details. Additionally, previous works showed that user tags performed better or on par compared with visual features \cite{DBLP:conf/aaai/TongeC16,Squicciarini2014,Zerr:2012,tongemsm18,DBLP:journals/corr/TongeC15}.
For example, in our previous work \cite{DBLP:conf/aaai/TongeC16,tongemsm18,DBLP:journals/corr/TongeC15}, we showed that the combination of user tags and deep tags derived from AlexNet performs comparably to the AlexNet based visual features. Hence, in this experiment, we compare the performance of fc-R features with the tag features. For deep tags, we follow the same approach as in our previous work \cite{DBLP:conf/aaai/TongeC16,tongemsm18,DBLP:journals/corr/TongeC15} and consider the top $k=10$ object labels since $k=10$ worked best. ``DT-A,'' ``DT-G,'' ``DT-V,'' and ``DT-R'' denote deep tags generated by AlexNet, GoogLeNet, VGG-16, and ResNet, respectively. Deep tags are generated using the probability distribution over $1,000$ object categories for the input image obtained by applying the softmax function over the last fully-connected layer of the respective CNN.

\begin{figure}
\centering
\includegraphics[scale=0.35]{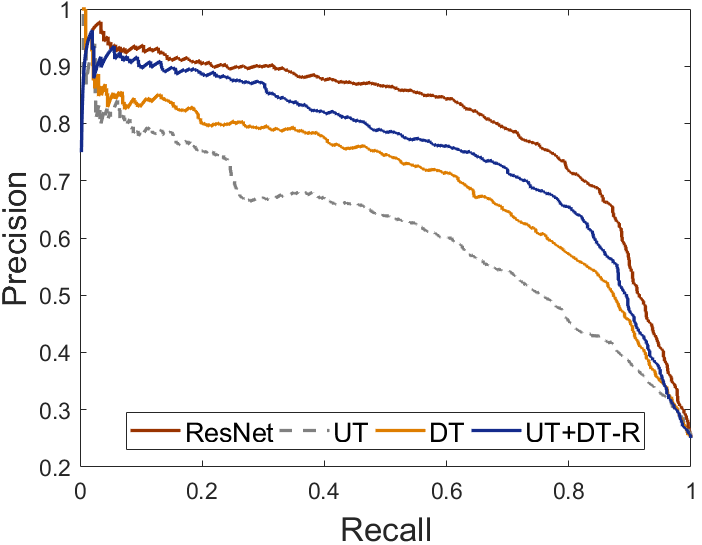}
\caption{\label{fig:prcurvetag} Precision-Recall curves for the private class obtained using visual features (fc-R) and tag features as user tags (UT), deep tags (DT-R), the combination of user tags and deep tags (UT + DT-R).}
\end{figure}

\begin{figure*}[t]
\renewcommand{\arraystretch}{1.1}
\setlength{\tabcolsep}{2pt}
\centering
\begin{adjustbox}{max width=14cm}
\begin{small}
\scalebox{1.0}{
\begin{tabular}{@{}lcccccc@{}}
\hline
\hline
Features &
\subfigure[]{\includegraphics[scale=0.18]{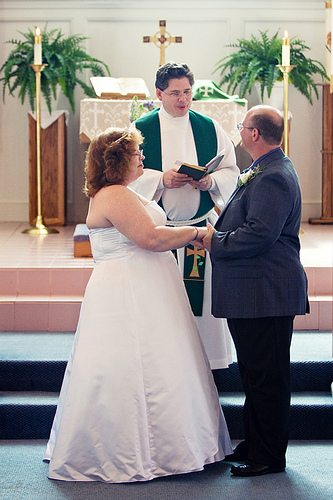}} &
\subfigure[]{\includegraphics[scale=0.25]{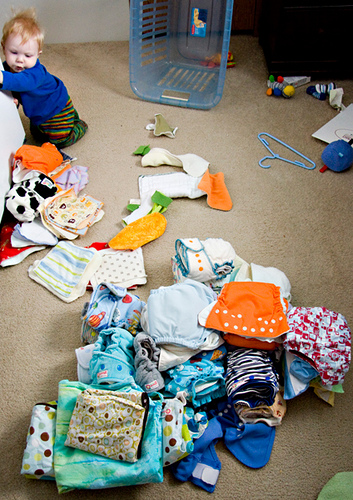}} &
\subfigure[]{\includegraphics[scale=0.2]{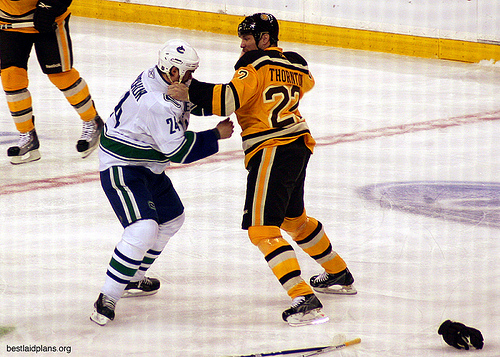}} &
\subfigure[]{\includegraphics[scale=0.2]{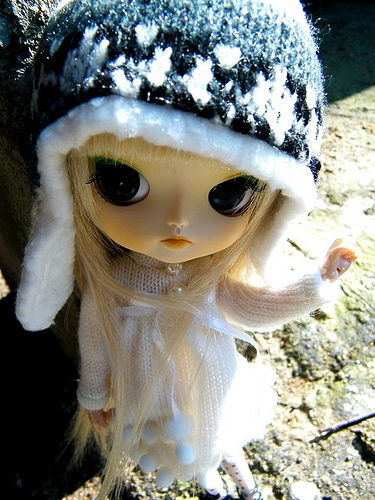}} &
\subfigure[]{\includegraphics[scale=0.2]{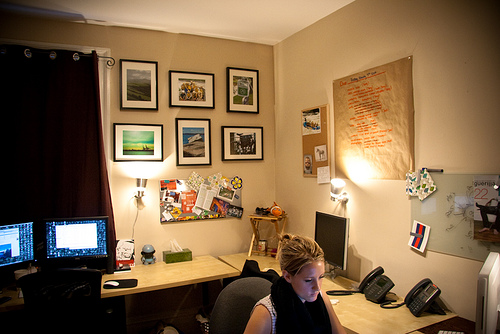}} &
\subfigure[]{\includegraphics[scale=0.2]{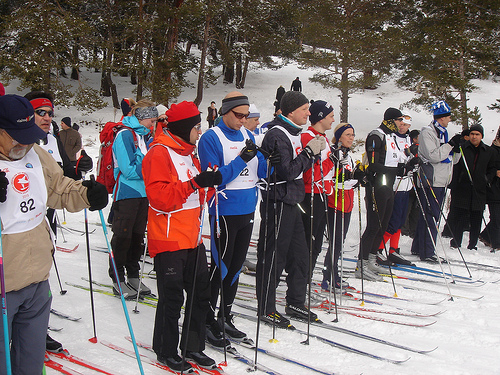}} \\
\hline
Visual & \color{ao}{\em private} & \color{ao}{\em private} & \color{ao}{\em public} & \color{ao}{\em public} & public & private\\
Tags & \color{ao}{\em private} & \color{ao}{\em private} & private & private & \color{ao}{\em private} & \color{ao}{\em public}\\
\hline
\end{tabular}	 
}
\end{small}
\end{adjustbox}
\caption{Privacy predictions obtained by image content encodings.} 
\label{fig:results}
\end{figure*}

\begin{table*}[t]
\centering
\begin{tabular}{|c|c|c|c|c|}
\hline
Rank 1-10 & Rank 11-20 & Rank 21-30 & Rank 31-40 & Rank 41-50 \\
\hline
{\bf people}		& pyjama			& maillot		 & {\bf promontory}	 & jersey \\
wig					& jammies			& {\bf girl} 	 & t-shirt 			 & mole \\
{\bf portrait}		& sweatshirt		& suit of clothes & foreland		  & groin \\
bow-tie				& {\bf outdoor}		& ice lolly 	 & {\bf headland}	 & bulwark \\
neck brace			& {\bf lakeside}	& suit 			 & bandeau			 & seawall \\
{\bf groom}			& {\bf lakeshore}  	& lollipop 		 & miniskirt		 & {\bf seacoast}\\
{\bf bridegroom}	& sun blocker 		& two-piece		 & breakwater		 & {\bf indoor}\\
laboratory coat 	& sunscreen 		& tank suit		 & {\bf vale}		 & stethoscope\\
hair spray 			& sunglasses 		& bikini 		 & hand blower 		 & {\bf valley}\\
shower cap 			& military uniform 	& swimming cap 	 & {\bf jetty}		 & {\bf head} \\

\hline
\end{tabular}
\caption{Top $50$ highly informative tags. We use the combination of deep tags and user tags (DT+UT) to calculate the information gain. User tags are shown in bold.}
\label{table:ig}
\end{table*}

Table \ref{table:compvisualftagf} compares the performance obtained using models trained on fc-R features with the performance of models trained on the tag features. We consider tag features as: (1) user tags (UT); 
(2) deep tags (DT) obtained from all architectures; 
(3) the combination of user tags and best performing deep tag features using Bag-of-Tags (BoT) model; and 
(4) Tag CNN applied to the combination of user and deep tags. 
As can be seen from the table, the visual features extracted from ResNet outperform the user tags and deep tags independently as well as their combination. The models trained on fc-R features achieve an improvement of $2\%$ over the CNN trained on the combination of user tags and deep tags (Tag CNN).  Additionally, the models trained on fc-R features yield an increase of $9.5\%$ in the F1-measure over the user tags alone and an increase of $4\%$ over the best performing deep tags, i.e., DT-R (among the deep tags of the four architectures). 

From Table \ref{table:compvisualftagf}, we also observe that the Tag CNN performs better than the Bag-of-Tags model (DT-R+UT), yielding an improvement of $3.0\%$ in the F1-measure of private class. Additionally, even though the visual features (fc-R) yield overall a better performance than the tag features, for the private class, the F1-measure ($0.717$) of the visual features (fc-R) is comparable to the F1-measure ($0.706$) of the Tag CNN. It is also interesting to note that the Visual CNN (fc-R) achieves an increase of $8\%$ in the precision (private class) over the Tag CNN whereas the Tag CNN obtains an improved recall (private class) of $5\%$ over the Visual CNN. 

In order to see how precision varies for different recall values, we also show the precision-recall curves for the visual and tag features in Figure \ref{fig:prcurvetag}. To avoid clutter we show the precision-recall curves for deep tags derived through ResNet and the combination of user tags and deep tags (DT-R) using BoT model. 
From the curves, we can see that the ResNet visual features perform better than the tag features, for a wide range of recall values from $0.3$ to $0.8$. 

We further analyze both the type of image encodings (visual \& tag) by examining the privacy predictions obtained for anecdotal examples using both the encodings. 


\subsubsection{\bf \em  Anecdotal Examples:} In order to understand the quality of predictions obtained by visual and tag features, we show privacy predictions for some samples obtained by both type of features. 
Figure \ref{fig:results} shows the predictions obtained using SVM models trained on the visual features and those trained on the combination of user tags and deep tags.  Correct predictions are shown in italics and green in color. We can see that  for images (a) and (b), the models trained on image tags (UT+DT) and visual features provide correct predictions. 
The tags such as ``groom,'' ``bride,'' ``wedding,'' ``photography'' describe the picture (a) adequately, and hence, using these tags appropriate predictions are obtained. 
Similarly, visual features identify the required objects, and a relationship between the objects and provide an accurate prediction for these images. Consider now examples (c) and (d). For these images, visual features capture the required objects to make accurate predictions, whereas, image tags such as ``bruins,'' ``fight,'' of image (c) and ``cute,'' ``wool,'' ``bonnet'' of image (d) do not provide adequate information about the picture and hence, yield an incorrect prediction.  
However, tags such as ``hockey,'' ``sports'' for image (c) and ``toy,'' ``doll'' for image (d) would have helped to make an appropriate prediction. 
We also show some examples, (e) and (f), for which visual features fail to predict correct privacy classes. Particularly, for image (f), we notice that visual features capture object information that identifies the image as private. On the other hand, the image tags such as ``festival'' and ``sport'' (describing the scene) provide additional information (over the object information) that helps the tag-based classifier to identify the picture as public. 

Next, we provide the detailed analysis of image tags with respect to privacy.

\begin{figure*}[t]
\centering
\subfigure[Private]{\includegraphics[scale=0.42]{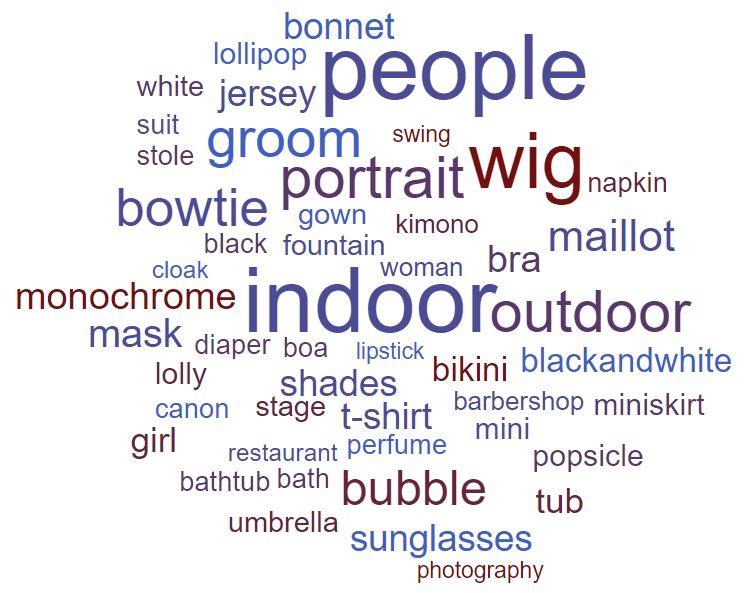}}
\subfigure[Public]{\includegraphics[scale=0.42]{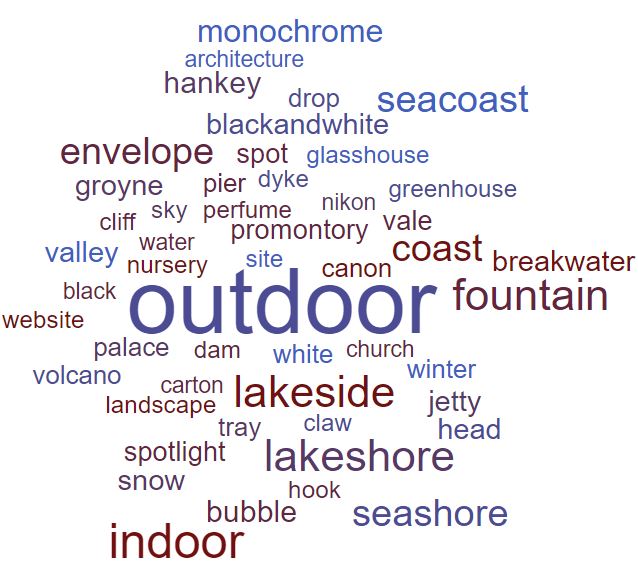}}
\caption{\label{fig:tagcloud} High frequency tag clouds with respect to public and private images.} 
\end{figure*}

\begin{figure*}[t]
\centering
\subfigure[Private]{\includegraphics[scale=0.32]{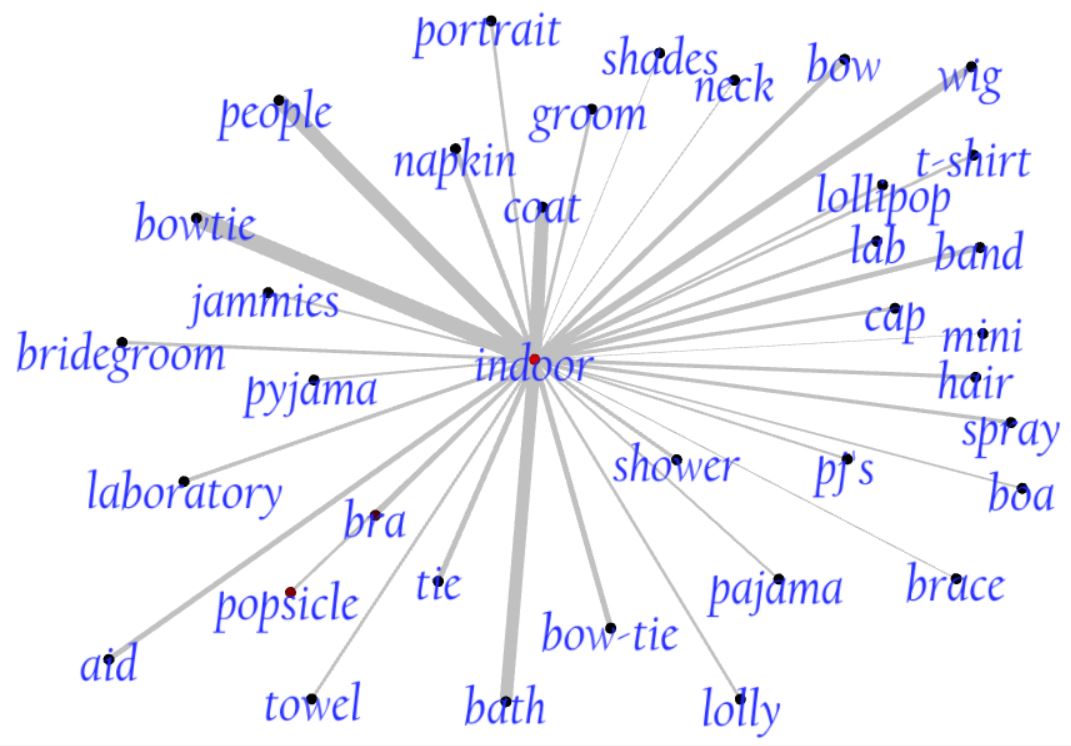}}
\subfigure[Public]{\includegraphics[scale=0.34]{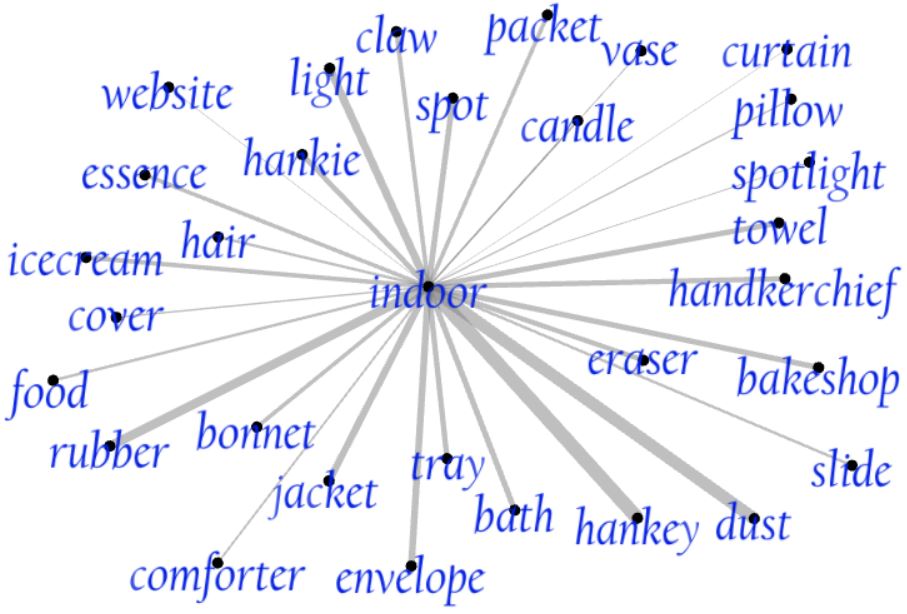}}
\caption{\label{fig:tagasso} Tag association graph.} 
\end{figure*}

\subsubsection{\bf \em Analysis of Image Tags with Respect to Privacy Classes:}

We provide an analysis of the deep tags (capturing the visual content of the image) and user tags to learn their correlation with the private and public classes. First, we rank user tags and deep tags based on their information gain on the train set. Table \ref{table:ig} shows top $50$ tags with high information gain. From the table, we observe that the tags such as ``maillot,'' ``two-piece,'' ``sandbar'' provide high correlation to the privacy classes. We also notice that deep tags (objects) contribute to a significant section of top $50$ highly informative tags.
 Secondly, we rank both the tags (user and deep tags) based on their frequency in public and private classes. We show $50$ most frequent tags for each privacy class using word clouds in Figure \ref{fig:tagcloud}. The tags with larger word size depict a higher frequency of the tag. We notice that tags such as ``indoor,'' ``people,'' ``portrait'' occur more frequently in the private class, whereas tags such as ``outdoor,'' ``lakeside,'' ``fountain,'' occur more frequently in the public class. 

We also observe that some informative tags overlap in both public and private clouds (See Figure \ref{fig:tagcloud}, e.g., ``indoor'').  Thus, we analyze other tags that co-occur with the overlapping tags to further discriminate between their association with the public and private classes. 
To inspect the overlapping tags, we create two graphs with respect to public and private classes. For the public graph, we consider each tag as a node in the graph and draw an edge between the two nodes if both the tags belong to the same public image. Likewise, we construct another graph using private images.  
Figure \ref{fig:tagasso} shows portions of both public and private graphs for ``indoor'' tag. To reduce the complexity of visualization, we only display nodes with stronger edges that have the co-occurrence greater than a certain threshold. Note that stronger edges (edges with higher width) represent the high co-occurrence coefficient between two nodes (in our case, tags). From the graphs, we observe that the overlapping tag ``indoor'' tends to have different highly co-occurring tags for public and private classes. For example, the ``indoor'' tag shows high co-occurrence with tags such as ``people,'' ``bath,'' ``coat,'' ``bowtie,'' ``bra'' (tags describing private class) in the private graph. On the other hand, in the public graph, the tag shows high co-occurrence with ``dust,'' ``light,'' ``hankey,'' ``bakeshop,'' ``rubber,'' and so forth (tags describing public class). Even though some tags in the graph have comparatively low co-occurrence, the tags occurring in the private graph tend to associate with the private class whereas the tags from the public graph are more inclined towards the public class.

\begin{figure*}[t]
\centering
\subfigure[Ratio of private to public images for top $10$ tags]{\includegraphics[scale=0.3]{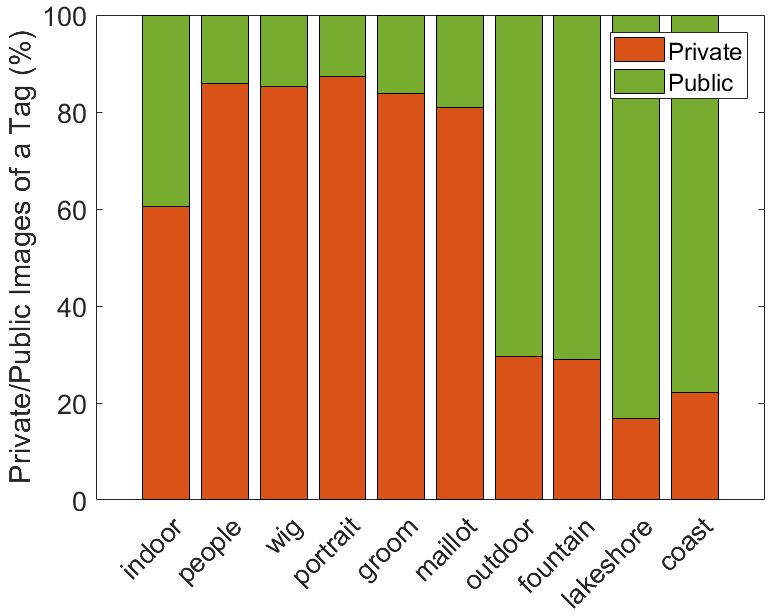}}
\subfigure[Tag frequency of top $1000$ tags]{\includegraphics[scale=0.32]{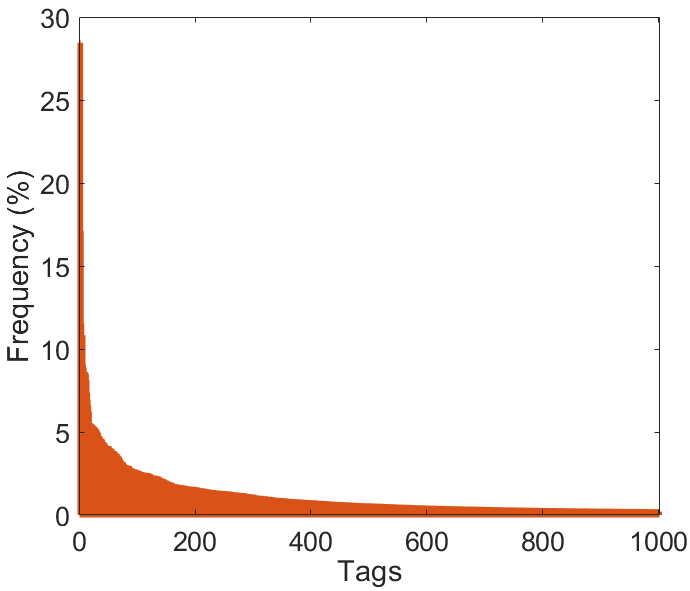}}
\caption{\label{fig:tagfreq} Analysis of top frequently occurring tags.} 
\end{figure*}

\begin{table*}[t]
\centering
\begin{small}
\begin{tabular}{|l|c|c|c|c|c|c|c|c|c|c|}
\hline
& \multicolumn{4}{|c|}{Overall} & \multicolumn{3}{|c|}{Private} & \multicolumn{3}{|c|}{Public} \\
\hline
{Features} & {Acc \%} & {F1} & {Prec} & {Re} & {F1} & {Prec} & {Re} & {F1} & {Prec} & {Re}\\
\hline
fc-R & 87.58 & 0.872 & 0.872 & 0.876 & 0.717 & 0.783 & 0.662 & 0.92 & 0.899 & {\color{blue}{\bf 0.943}} \\
fc-R+UT & {\color{blue}{\bf 88.29}} & {\color{blue}{\bf 0.881}} & {\color{blue}{\bf 0.88}} & {\color{blue}{\bf 0.883}} & {\color{blue}{\bf 0.753}} & {\color{blue}{\bf 0.799}} & {\color{blue}{\bf 0.713}} & {\color{blue}{\bf 0.923}}	& {\color{blue}{\bf 0.907}}	& 0.94
\\
\hline
\end{tabular}
\end{small}
\caption{Results for the combination of visual and tag features.} 
\label{table:visualntag}
\end{table*}

We further analyze the privacy differences of top $10$ private and public image subjects. We consider ``outdoor,'' ``indoor,'' ``fountain,'' ``lakeshore,'' and ``coast'' for the  public class. On the other hand, we consider ``indoor,'' ``people,'' ``wig,'' ``portrait,'' ``outdoor,'' ``groom,'' and ``maillot'' for the private class. Note that since images may have various tags associated with them,  an image can be counted towards more than one tag. 
Given that the dataset contains three times more public images than private images ($3:1$ public to private ratio), we count $3$ for each private image as opposed to the public class where we count $1$ for each public image for a fair comparison. The ratio of private to public content for a specific tag is shown in Figure \ref{fig:tagfreq} (a). For example, out of the total images that possess the ``indoor'' tag, $60\%$ images are of private class. From the figure, we observe that tags except for ``indoor'' show a significant difference in the inclination towards public and private classes. We also plot the frequency of top $1000$ tags normalized by the dataset size in Figure \ref{fig:tagfreq} (b). The plot shows that the top $200$ tags befall in $3\%-30\%$ of the dataset with very few tags occurring in around $20\%$ of the dataset. We also observe that most of the tags lie below $3\%$ of the dataset showing the variation in the images' subjects and complexity of the dataset which justifies the fact that increasing the number of images increases the challenges of the problem statement.

\subsection{Fusion of Visual and Tag Features for Image Privacy Prediction}
Visual encoding and tag encoding capture different aspects of images. 
Thus, we add the top $350$ correlated tags to the visual features fc-R and evaluate their performance for privacy prediction. 
We experiment with the number of top correlated tags $= \{10, 20, \cdots, 50, 100, \cdots, 500, 1000, 5000, 10000\}$. However, we get the best results with the top $350$ correlated tags. Table \ref{table:visualntag} shows the results obtained using SVMs trained on fc-R and the combination of fc-R with the top $350$ correlated user tags (fc-R+tag). The results reveal that adding the highly correlated tags improves the privacy prediction performance. Precisely, we get a significant improvement of $4\%$ on F1-measure of private class over the performance obtained using visual features fc-R. Note that, in our previous works \cite{tongemsm18,DBLP:conf/aaai/TongeC16,DBLP:journals/corr/TongeC15} and Experiment \ref{sec:visualvstag} (where we compare visual and tag features), we described visual content using tags (deep tags) and combined with the user tag to achieve a better performance. However, the combination of user tags and deep tags (combining one type of encoding) yields a lower performance as compared to the combination of user tags and fc-R features (combining two types of encodings). Precisely, the combination of user tags (UT) and fc-R features yields an improvement of $5\%$ in the F1-measure of private class (refer Tables \ref{table:compvisualftagf} and \ref{table:visualntag}) over the combination of user tags and deep tags. 

\section{Conclusion}\label{sec:conclusion}
In this paper, we provide a comprehensive study of the deep features derived from various CNN architectures of increasing depth to discover the best features that can provide an accurate privacy prediction for online images.  Specifically, we explored features obtained from various layers of the pre-trained CNNs such as AlexNet, GoogLeNet, VGG-16, and ResNet and used them with SVM classifiers to predict an image's privacy as {\em private} or {\em public}. We also fine-tuned these architectures on a privacy dataset.
The study reveals that the SVM models trained on features derived from ResNet perform better than the models trained on the features derived from 
AlexNet, GoogLeNet, and VGG-16. We found that the overall performance obtained using models trained on the features derived through pre-trained networks is comparable to the fine-tuned architectures. However, fine-tuned networks provide improved performance for the private class as compared to the models trained on pre-trained features. The results show remarkable improvements in the performance of image privacy prediction as compared to the models trained on CNN-based and traditional baseline features. Additionally, models trained on the deep features outperform rule-based models that classify images as private if they contain people.  We also investigate the combination of user tags and deep tags derived from CNN architectures in two settings: (1) using SVM on the bag-of-tags features; and (2) applying the text CNN over these tags. We thoroughly compare these models with the models trained on the highest performing visual features obtained for privacy prediction. We further provide a detailed analysis of tags that gives insights for the most informative tags for privacy predictions. We finally show that the combination of deep visual features with these informative tags yields improvement in the performance over the individual sets of features (visual and tag). 

The result of our classification task is expected to aid other very practical applications. For example, a law enforcement agent who needs to review digital evidence on a suspected equipment to detect sensitive content in images and videos, e.g., child pornography. The learning models developed here can be used to filter or narrow down the number of images and videos having sensitive or private content before other more sophisticated approaches can be applied to the data. Consider another example, images today are often stored in the cloud (e.g., Dropbox or iCloud) as a form of file backup to prevent their loss from physical damages and they are vulnerable to unwanted exposure when the storage provider is compromised. Our work can alert users before uploading their private (or sensitive) images to the cloud systems to control the amount of personal information (eg. social security number) shared through images. 

In the future, using this study, an architecture can be developed, that will incorporate other contextual information about images such as personal information about the image owner, owner's privacy preferences or the owner social network activities, in addition to the visual content of the image. Another interesting direction is to extend these CNN architectures to describe and localize the sensitive content in private images.

\balance


\section*{Acknowledgments}
This research is supported by the NSF grant \#1421970. The computing for this project was performed on Amazon Web Services.

\bibliographystyle{ACM-Reference-Format}
\bibliography{deepprivate}

\end{document}